\documentclass{article}

\usepackage{arxiv}
\usepackage{natbib}
\setcitestyle{square,sort,comma,numbers}
\usepackage[utf8]{inputenc} 
\usepackage[T1]{fontenc}    
\usepackage{hyperref}       
\usepackage{url}            
\usepackage{booktabs}       
\usepackage{amsfonts}       
\usepackage{nicefrac}       
\usepackage{microtype}      
\usepackage{lipsum}
\usepackage{framed,multirow}
\usepackage{booktabs}
\usepackage{gensymb}
\usepackage{algorithm}
\usepackage{algorithmic}%
\usepackage{bbm}
\usepackage{tabularx}
\usepackage{lineno}
\usepackage{amssymb}
\usepackage{latexsym}
\usepackage{amsmath}
\usepackage{subfigure}
\usepackage{placeins}
\usepackage{hyperref}
\usepackage{float}
\usepackage{pdfcomment}
\usepackage{graphics}
\usepackage{graphicx}
\usepackage{xcolor}
\definecolor{newcolor}{rgb}{.8,.349,.1}
\definecolor{newcolor}{rgb}{.8,.349,.1}
\definecolor{mygreen}{rgb}{0.1, 0.75, 0.15}
\newcommand{\norme}[1]{\left\Vert #1\right\Vert}

\newcommand{\ra}[1]{\renewcommand{\arraystretch}{#1}}
\newcommand{\tabitem}{~~\llap{\textbullet}~~}

\newcommand{\removeGG}[1]{}

\newcommand{\removeRH}[1]{}

\usepackage{tikz}
    \usetikzlibrary{shapes.arrows}
    \usetikzlibrary{decorations.shapes}
    \usetikzlibrary{decorations.pathreplacing}
    \usetikzlibrary{fadings,shapes.arrows,shadows}
\makeatletter
\DeclareRobustCommand\sfrac[1]{\@ifnextchar/{\@sfrac{#1}}%
                                            {\@sfrac{#1}/}}
\def\@sfrac#1/#2{\leavevmode\kern.1em\raise.5ex
         \hbox{$\m@th\mbox{\fontsize\sf@size\z@
                           \selectfont#1}$}\kern-.1em
         /\kern-.15em\lower.25ex
          \hbox{$\m@th\mbox{\fontsize\sf@size\z@
                            \selectfont#2}$}}

\usetikzlibrary{positioning}

\newcommand{\overbar}[1]{\mkern 1.5mu\overline{\mkern-1.5mu#1\mkern-1.5mu}\mkern 1.5mu}
\usepackage{lipsum} 
\newcommand*{\LargerCdot}{\raisebox{-0.25ex}{\scalebox{1.2}{$\cdot$}}}

\title{Towards Human Body-Part Learning for Model-Free Gait Recognition}

\author{
  Imad Rida  \\
  Normandie Univ, UNIROUEN, UNIHAVRE, INSA Rouen, LITIS \\
  76 000, Rouen\\
  France \\
  \texttt{imad.rida@insa-rouen.fr} \\
}

\begin{document}
\maketitle

\begin{abstract}
Gait based biometric aims to discriminate among people by the way or manner they walk. It represents  a biometric at distance which has many advantages over other biometric modalities. State-of-the-art methods require a limited cooperation from the individuals. Consequently, contrary to other modalities, gait is a non-invasive approach. As a behavioral analysis, gait is difficult to circumvent. Moreover, gait can be performed without the subject being aware of it. Consequently, it is more difficult to try to tamper one own biometric signature. In this paper  we review different features and approaches used in gait recognition. A novel method able to learn the discriminative human body-parts to improve the recognition accuracy will be introduced. Extensive experiments will be performed on CASIA gait benchmark database and results will be compared to state-of-the-art methods.
\end{abstract}


\section{Introduction}

Biometrics technologies were primarily used by law enforcement. Nowadays, biometrics are increasingly being used by government agencies and private industries to verify person's identity, secure the nation's borders, and to restrict access to secure sites including buildings and computer networks \cite{fei2017enhanced,rida2018ensemble,rida2018novel}. Biometrics systems recognize a person based on physiological characteristics, such as fingerprints, hand, facial features, iris patterns, or behavioral characteristics that are learned or acquired, such as how a person signs his name, typing rhythm, or even walking pattern \cite{al2018palmprint,rida2018feature,rida2018palmprint,shariatmadari2018off}. 

The problem of resolving the identity of a person can be categorized into two fundamentally distinct problems with inherent complexities: the authentication and recognition (most commonly known as identification). In fact, they do not address the same problem. Authentication, also known as verification, answers to the question " am I who I claim to be"\cite{rida2018palmprint}. The biometric system compares the information registered on the proof identity to the current person features. It corresponds to the concept of one-to-one matching. Identification refers to the question "who am I?". The subject is compared to the subjects already enrolled in the system. It is analogous to the notion of one-to-many matching \cite{micheletto2018multiple,rida2019forensic}. In our paper we are rather interested in the recognition context.

Gait is defined to be the coordinated, cyclic combination of the movements that result in human locomotion. The movements are coordinated in the sense that they must occur with a specific temporal pattern for the gait to occur. The movements in a gait repeat as a walker cycles between steps with alliterating feet. It is both coordinated and cyclic nature of the motion that makes gait a unique phenomenon \citep{boyd2005}.

People are often able to identify a familiar person from distance simply by recognizing the way the person walks. Based on this common experience, and the growing interest of biometrics, researchers exploit the gait characteristics for identification purpose. Initially, the ability of humans to recognize gaits arouses interest of the psychologists \citep{johansson1973, johansson1975} who showed that humans can quickly identify moving patterns corresponding to the human walking.

Gait recognition can be defined as the recognition of some salient property, such as, identity, style of walk, or pathology, based on the coordinated cyclic motions that result in human locomotion. In our chapter we are rather interested in recognizing the identity based on the gait characteristics. A distinction could be made between gait recognition and the so called quasi gait recognition. In the first one, salient property which is in our case the identity can be recognized from the gait characteristics of the walking subject; when in the second one the identity is recognized based on features extracted during walking, however these features do not rely on gait. For example, body dimensions could be measured and used for individuals recognition.

\begin{figure}[!h]
\centering
\subfigure[Subject at instant $t$] {\label{fig:aaa}\includegraphics[width=31mm]{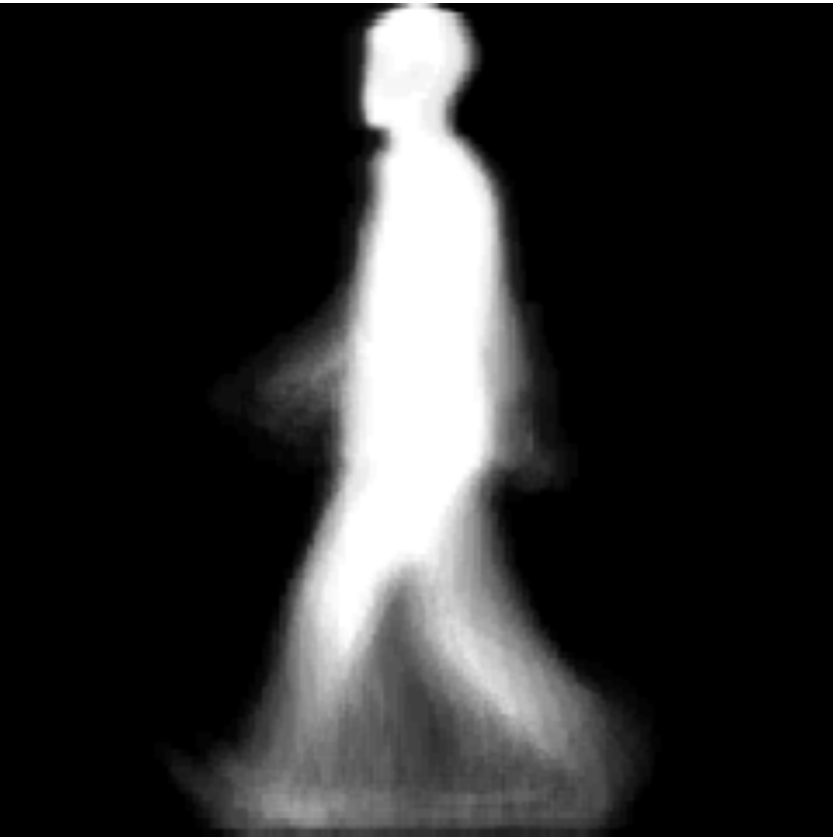}}
\subfigure[Subject at instant $t+1$ ]{\label{fig:bbb}\includegraphics[width=31mm]{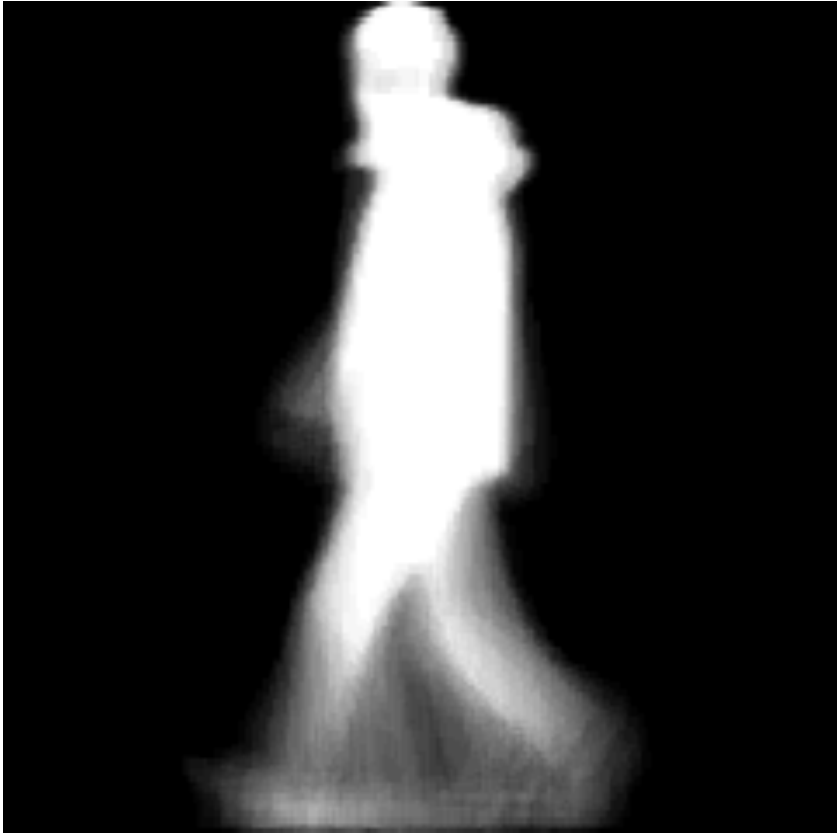}}
\subfigure[Intra-class variations]{\label{fig:ccc}\includegraphics[width=31mm]{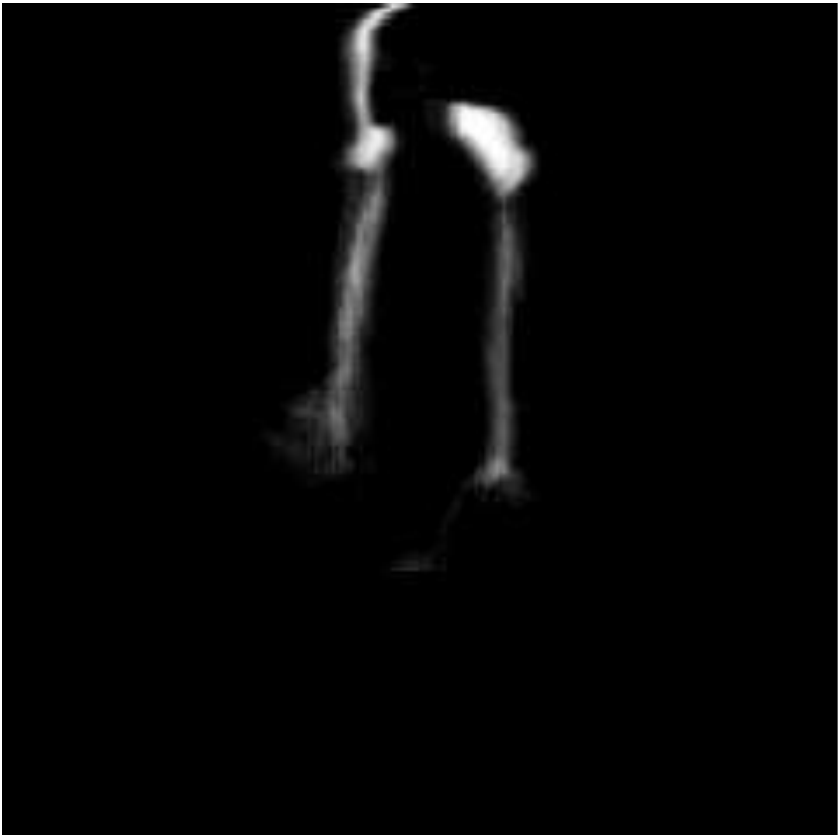}}
        \caption{Example of intra-class variations caused by clothing variations of the same subject recorded at instant $t$ and $t+1$. Image (c) is the difference of (a) and (b).}
\label{fig:intra}
\end{figure}

It has been demonstrated that the gait recognition performance is drastically influenced by different intra-class variations related to the subject itself, such as clothing variation, carrying conditions; or related to the environment such as view angle variations, walking surface, shadows and segmentation errors \citep{matovski2012,yu2006,han2006}. Figure \ref{fig:intra} shows an example of intra-class variations caused by the clothing variations of the same subject recorded at instants $t$ and $t+1$. The researchers in \citep{sarkar2005,yu2006} considered several conditions including carrying conditions, view angle and clothing variations and measure their impact on the recognition accuracy. Due to the influence of the previous intra-class variations caused by these conditions, considerable efforts have been devoted to build robust systems able to deal with individuals under different conditions. In this chapter we introduce a novel method able to select the robust human body-part corresponding to the dynamic part of the body which has been demonstrated to be less influenced by intra-class variations \citep{bashir2010,dupuis2013}.

\section{Gait analysis}
\subsection{Gait cycle}
\begin{figure}[!h]
\centering
\includegraphics [width= 7 cm] {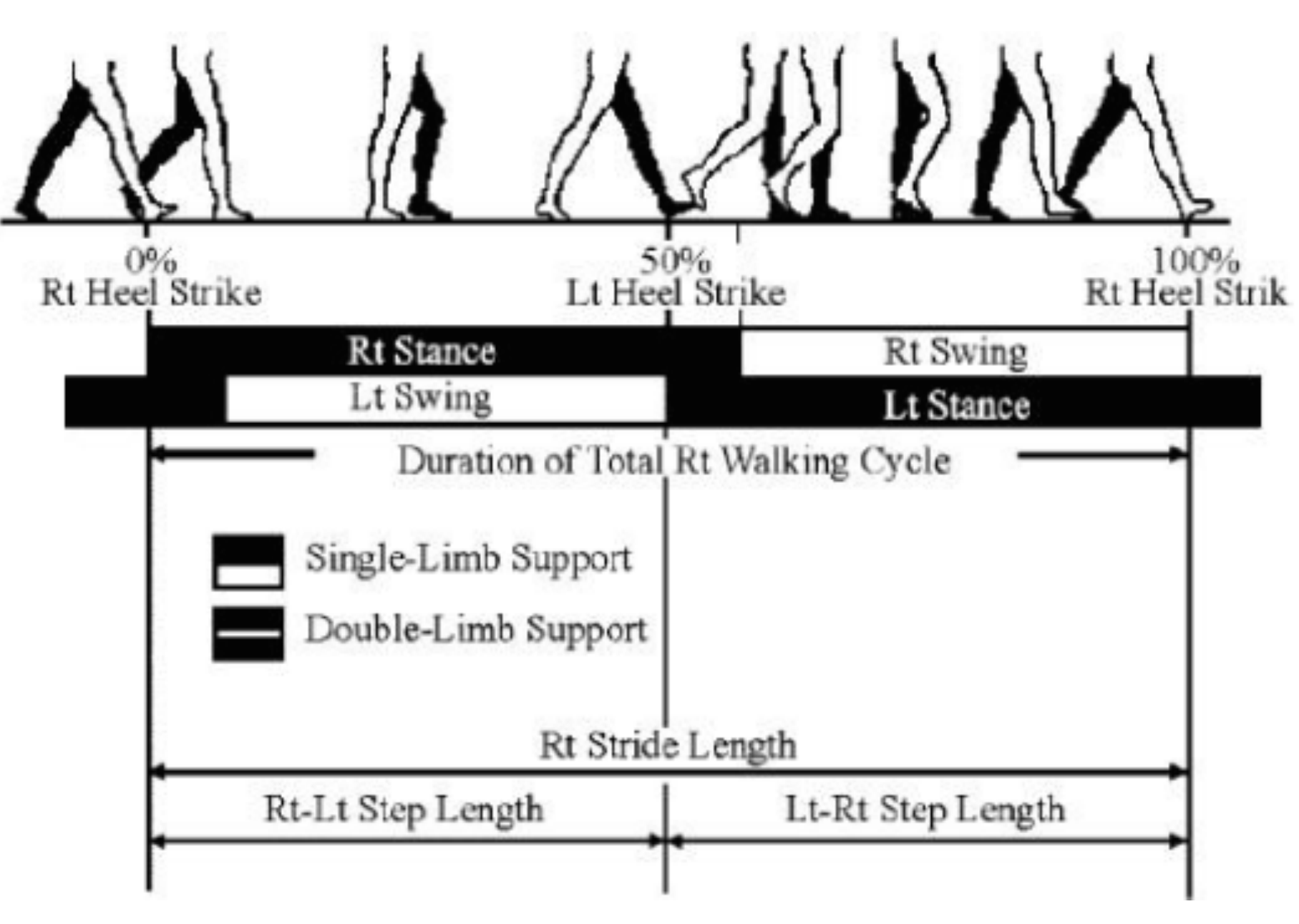}
\caption{Gait cycle of a subject depicting the two phases of the right foot: Right (Rt) stance and Rt swing \citep{cunado2003}.}
\label{fig:stance1}
\end {figure}

The gait cycle is the continuous repetitive pattern of walking or running. It is the time interval between successive instances of initial foot-to-floor contact "heel strike" for the same foot \citep{cunado2003}. A complete gait cycle can be divided into two main phases: stance and swing as is shown in Figure \ref {fig:stance1}, these phases can be even eventually further split up. It has been shown that when a person walks, stance phase accounts 60 \% of the gait cycle, however when a person runs, the main proportion of the gait exists in the swing phase. Moreover, the double support frame does not exist as there is a period in which neither feet touch the ground.

There has been a considerable amount of work regarding the variations in the intrinsic properties of the gait during the human walk such as velocity, motion, body length and width etc. These works mainly extract the features in a time duration corresponding to the human walking cycle. This shows that the detection and estimation of the walking cycle is of extreme importance in recognition. Based on the adopted approach, different information could be extracted.

\subsection{Characteristics of human gait}

It has been demonstrated that the human gait is unique \citep{murray1964, murray1967}. It has also been shown that the information such pelvic and thorax is different from one person to another. This information  could be used for individuals discrimination, however the main issue is that these patterns are not adapted for computer vision based biometric systems since they are hardly measured during the individual walk.  

Since many features established by medical studies appear unsuited to a computer vision-based system, the components for this investigation have been limited to the rotation patterns of the hip and knee. These patterns are possible to be extracted from real images, furthermore it has been shown from medical studies that they possess a high degree of individual consistency and inter-individual variability. These features belong to the so called model-based gait recognition which will be introduced in Section \ref{modelbased}.

Currently, more adapted vision systems features called holistic have been introduced. These features take in consideration all the body motion which contains very discriminative information to differentiate between different individuals. These features belong to the so called model-free approach described in Section  \ref{modelfree}.

\section{Gait recognition approaches}

\subsection{Model-based gait recognition}
\label{modelbased}
In the model-based approach, the features representatives of a gait are derived from a known structure or fitted model. The model mimics the human skeleton. Consequently, model-based approaches are based on prior knowledge \cite{rida2018robust,rida2015unsupervised}. 

\begin{figure}[!htbp]
\centering
\subfigure[\citep{nixon2009model}] {\label{fig:a}\includegraphics[width=50mm]{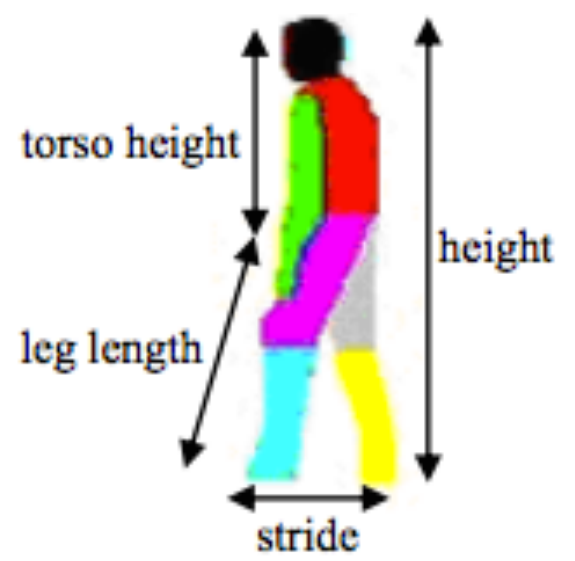}}
\subfigure[\citep{wagg2004}]{\label{fig:d}\includegraphics[width=24mm]{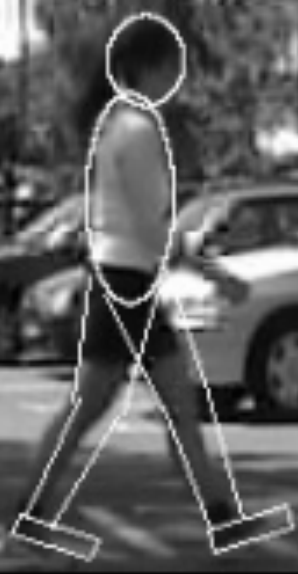}}
\subfigure[\citep{wang2004}]{\label{fig:c}\includegraphics[width=26mm]{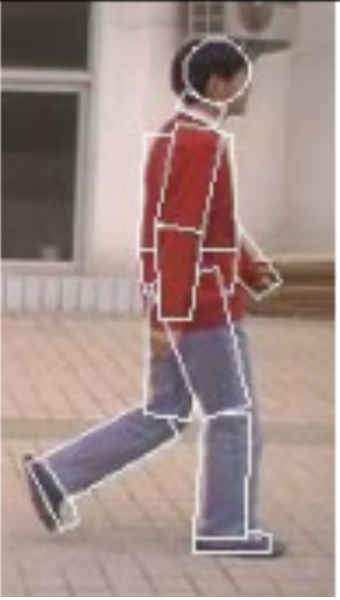}}
        \caption{Example of body models.}
\label{fig:models}
\end{figure}

The model based approaches, often need both a structural and a motion model which attempt to capture both static and dynamic information of the gait. The models could be 2 or 3 dimensional. The structured model describes the body topology, such as stride length, height, hip, torso, knee. This model can be made up of primitive shapes (cylinders, cones, and blobs), stick figures, or arbitrary shapes describing the edge of these body parts. On the other hand, a motion model describes the kinematics or the dynamics of the motion of each body part. Kinematics generally describe how the subject changes position with time without considering the effect of masses and forces, whereas dynamics will take into account the forces that act upon these body masses and hence the resulted motion \citep{benabdelkader2002}. Examples of the models are depicted  in Figure \ref{fig:models}.

The proposed works in model-based approach can be broadly splitted into two types of methods, those based on the estimation of the body parameters (length, width, cadence etc) directly from the raw videos and those trying to fit a model to capture the evolution of these parameters over time. 

In the body parameters estimation approach, \citep{bobick2001} proposed to recover static body and stride parameters of subjects, the comparison metric being based on mutual information. \citep{tanawongsuwan2001} used the trajectories of joint angles from motion, nearest-neighbor has been used for classification.\citep{benabdelkader2002} used stride and cadence of a walking person and a Bayesian approach has been used for classification. \citep{boulgouris2007} separated human body into different components, they adopted a distance metric to describe the resemblance between two silhouettes with respect to a certain body component. \citep{cunado2003} extracted the angular information during the walking process from the upper leg using the Fourier series, then the nearest neighbor technique is applied for classification. \citep{zeng2014} used side silhouette lower limb joint angles to characterize the dynamic gait part. Radial Basis Function (RBF) neural networks through deterministic learning has been used for recognition.

In the fitting model approach, \citep{lee2002} used appearance and dynamic traits of gait by analyzing parameters of fitted ellipses to regions of a subject's silhouette. \citep{wang2004} modeled human body as fourteen rigid parts connected to one another at the joints. Dynamic information as well as static information combined with nearest-neighbor classifier have been used for classification. \citep{zhang2004} introduced a non-rigid 2D body contour by a Bayesian graphical model whose nodes correspond to point positions along the contour.  \citep{zhang2007} suggested  a five-link biped human locomotion model to extract the joint position trajectories. The recognition step is then performed using Hidden Markov Models (HMMs). \citep{lu2007} used a full-body layered deformable model to capture information from the silhouette of the walking subject.  \citep{ariyanto2012} introduced a new 3D model approach using a marionette and mass-spring model. \citep{yoo2008} extracted nine coordinates from the human body contours based on human anatomical knowledge to construct a 2D model; back-propagation neural network algorithm has been used for classification. \citep{tafazzoli2010} used  active contour models and Hough transform to model the movements of the articulated parts of the body. Nearest-neighbor is applied for classification. 

Table \ref{tab:modelbased} summarizes the captured features and the classifiers used in model-based techniques introduced above. Model-based methods seem to be very attractive and promising since they have the ability to deal with the various intra-class variations caused by different conditions such as clothing, carrying, which affects the subjects appearance. However the complexity of the models and the extraction of their components from the video stream is not a trivial task. Consequently, model-based techniques are preferred in practice.

\begin{table}
  \centering
  \caption {Overview of model-based methods (features and classifiers).}
   \noindent
\resizebox{0.8\linewidth}{!}{
  \begin{tabular}{lll}
    \multicolumn{3}{c}{}  \\
        \toprule
         \toprule

    Method & Features & Classification   \\
    \midrule  
     \midrule  

    \tabitem \citep{bobick2001} & length, width, stride & nearest-neighbor \\
    \tabitem \citep{tanawongsuwan2001} & joint-angle trajectories & nearest-neighbor  \\
     \tabitem \citep{benabdelkader2002} & stride, cadence  & Bayesian    \\
      \tabitem \citep{boulgouris2007} & body components &  metric based body parts   \\
       \tabitem \citep{cunado2003} & motion upper leg & nearest-neighbor \\
        \tabitem \citep{zeng2014} & lower limb joint-angles & RBF neutral network \\
        \hline
        \hline
          \tabitem \citep{lee2002} & parameters of fitted ellipse model & support vector machine \\
         \tabitem \citep{wang2004} & rigid model  (joint-angles) & nearest-neighbor \\
        \tabitem \citep{zhang2004} & non-rigid model (deformations) & chain-like model   \\
        \tabitem  \citep{zhang2007} & five-link biped model (joint-trajectories) & hidden Markov models \\
        \tabitem \citep{lu2007} & deformable model (length, width, orientations) & adaboost \\
         \tabitem \citep{ariyanto2012} & 3D model (motion) & nearest-neighbor  \\
        \tabitem  \citep{yoo2008} & 2D model (rhythmic, periodic motion) & neural network \\
        \tabitem \citep{tafazzoli2010} & model based anatomy (leg and arm movement) & nearest-neighbor  \\
    \bottomrule
     \bottomrule

  \end{tabular}}
  \label{tab:modelbased}
\end{table}

 \subsection{Model-free gait recognition}
 \label{modelfree}
In the model-free approach, the gait characteristics are derived from the moving shape of the subject. It actually corresponds to image measurements. In this case, no human model to rebuild the human walking steps is needed. A random example of model-free approach features is the shape variation within a particular region of walking subject.  In the recent past, a lot of features have been introduced in the context of model-free gait recognition. The features can either be solely based on the moving shape  (no prior shape information is explicitly taken in consideration) or also integrate the motion within the feature representation \cite{rida2018robust}. 

\subsubsection{Model-free gait features}

Mainly human gait features are organized as temporal and spatial, however \citep{dupuis2013} proposed an interesting and more general taxonomy organizing gait features in five main categories: contour, optical flow, silhouette, moments and gait energy/entropy/motion history.

\begin{itemize}
\item Contour: the contours have the advantage of being low computational cost, however they  suffer too much from the intra-class variations. An example of gait recognition based contour features is symmetry operators introduced by \citep{hayfron2003} which are able to form a robust feature representation from few training samples.

\begin{figure}[!h]
\centering
\includegraphics [width= 9 cm] {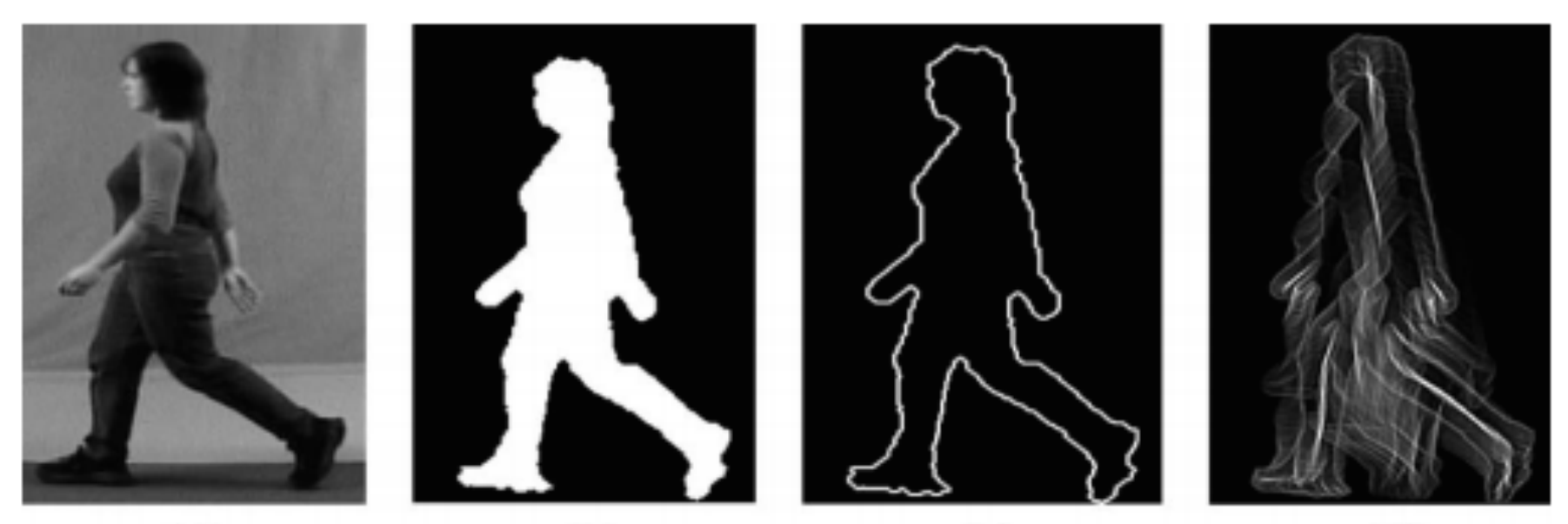}
\caption{Symmetry operator introduced in \citep{hayfron2003}.}
\label{fig:stance}
\end {figure}

\item Optical flow: the dynamic aspect of the human motion is extracted based on the optical flow shown  in Figure \ref{fig:flow}. It represents a robust feature representation against the various intra-class variations because it takes only motion information in consideration. However it needs a lot of computational cost \citep{bashir2009}.

\begin{figure}[!h]
\centering
\includegraphics [width= 9 cm] {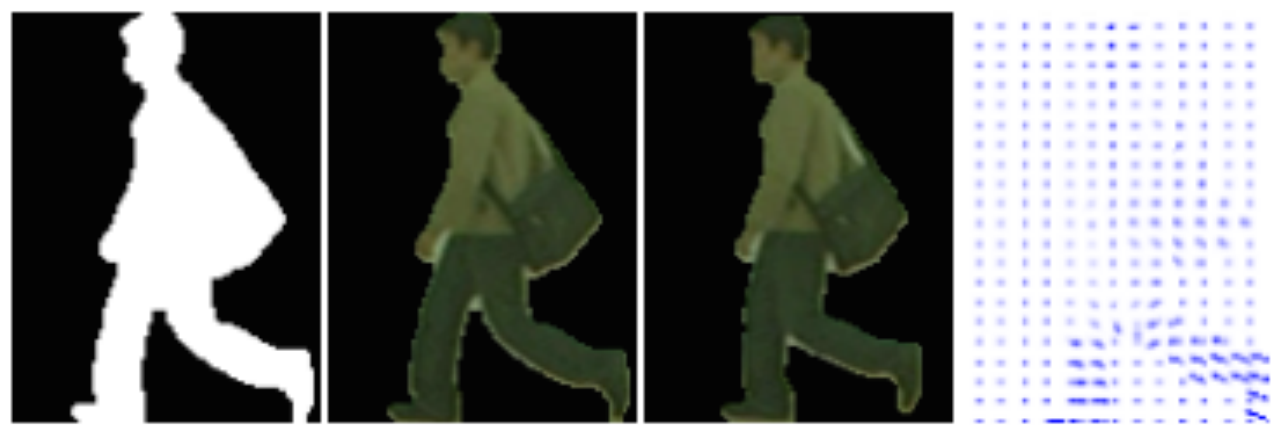}
\caption{Example of the optical flow in \citep{bashir2009}.}
\label{fig:flow}
\end {figure}

\item Silhouette: the whole silhouette is taken in consideration. This can be advantageous because the errors of silhouette segmentation are avoided. An example of gait recognition based on silhouette is the self-similarity introduced by \citep{benabdelkader2001}. It consists on calculating the cross-correlation between each pair of images in a gait sequence (see Figure \ref{fig:ssp}).
\begin{figure}[!h]
\centering
\includegraphics [width= 9 cm] {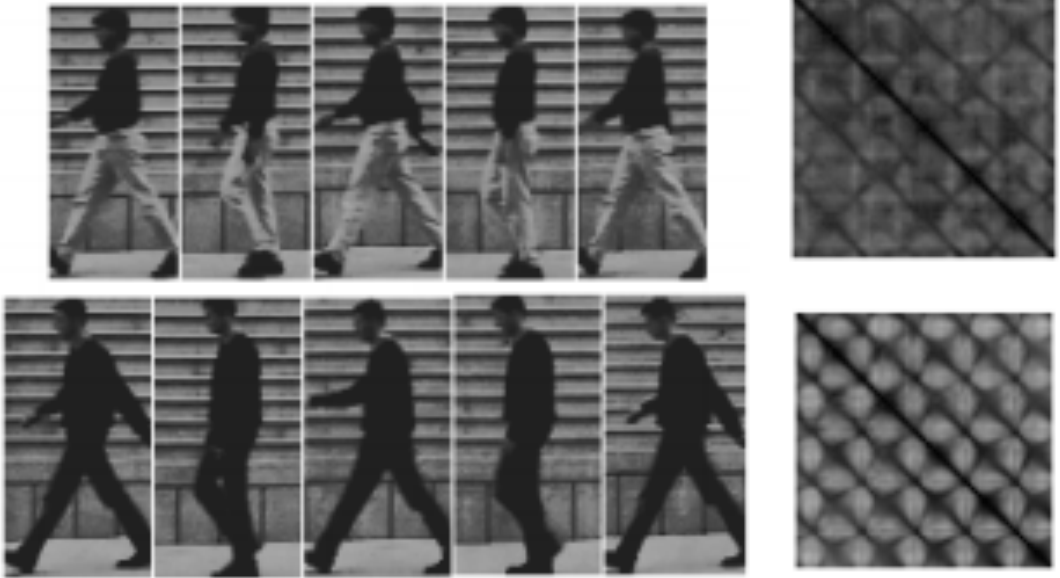}
\caption{Example of self similarity features in  \citep{benabdelkader2001}. The rightmost images represent self similarity representation.}
\label{fig:ssp}
\end {figure}

\item Moments: moments are extracted from the silhouettes based on feature extractors including, Local Binary Patterns (LBP), Histogram of Oriented Gradients (HOG), etc. They are more robust to intra-class variations caused by occlusion and shape variations. A good example describing this category is features extracted from the silhouette based on Gabor filters \citep {tao2007}. Figure \ref{fig:gabor} shows the sum of Gabor filter responses over directions, scales and both scales and directions.

\begin{figure}[!h]
\centering
\includegraphics [width= 9 cm] {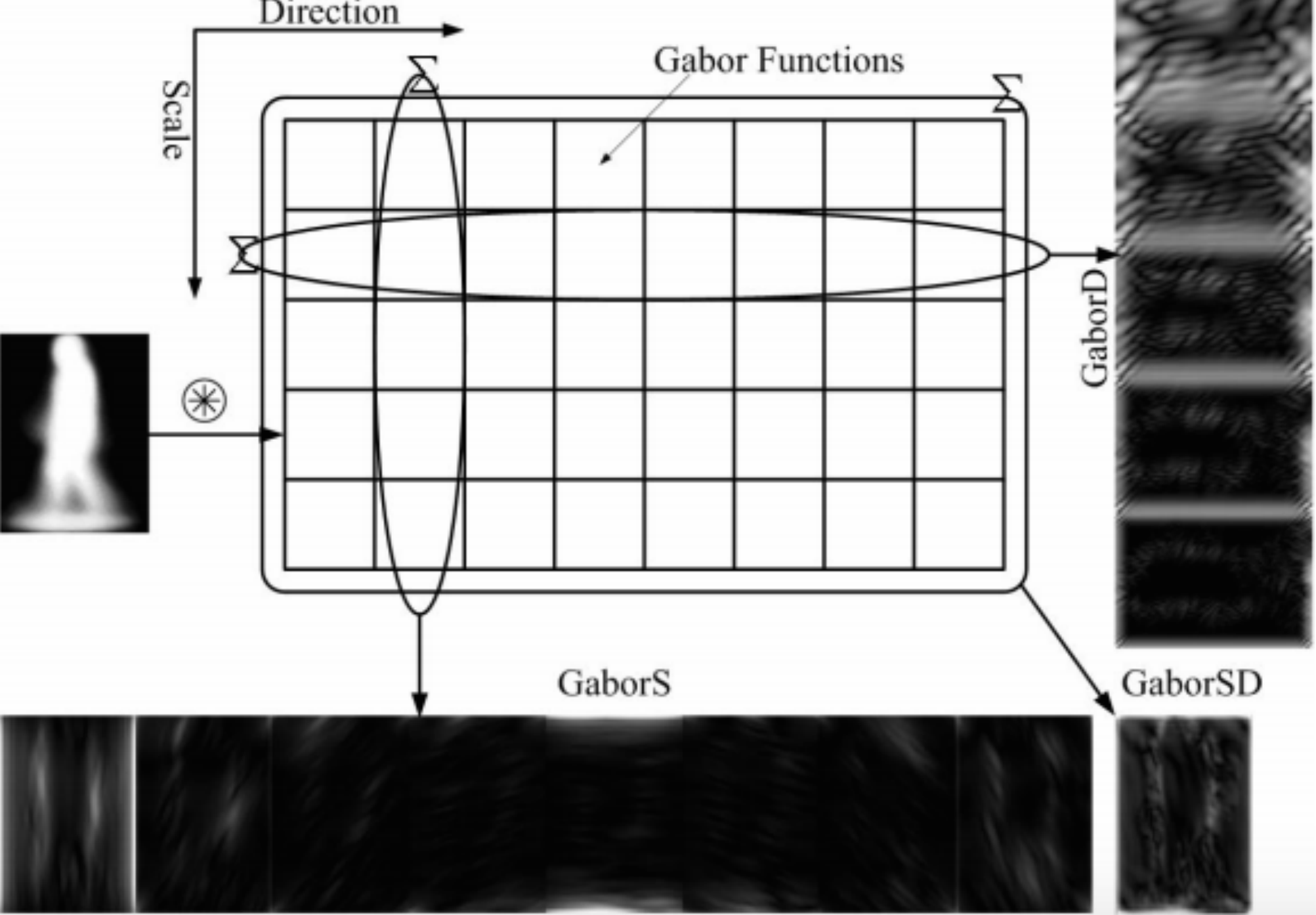}
\caption{Example of extracted features using Gabor filters in  \citep {tao2007}. GaborD, GaborS, GaborSD represent the sum over directions, scales and both directions and scales of Gabor functions respectively.}
\label{fig:gabor}
\end {figure}

\item Energy/entropy/motion history: these features attempt to capture both spatial and temporal information of the gait using a single robust signature. The average image which represents a gait cycle is a good example which describes this family  \citep{liu2004}. Figure \ref{fig:average} shows the average image for several subjects obtained by averaging the segmented silhouettes of walking subjects during an entire cycle.

\begin{figure}[!h]
\centering
\includegraphics [width= 9 cm] {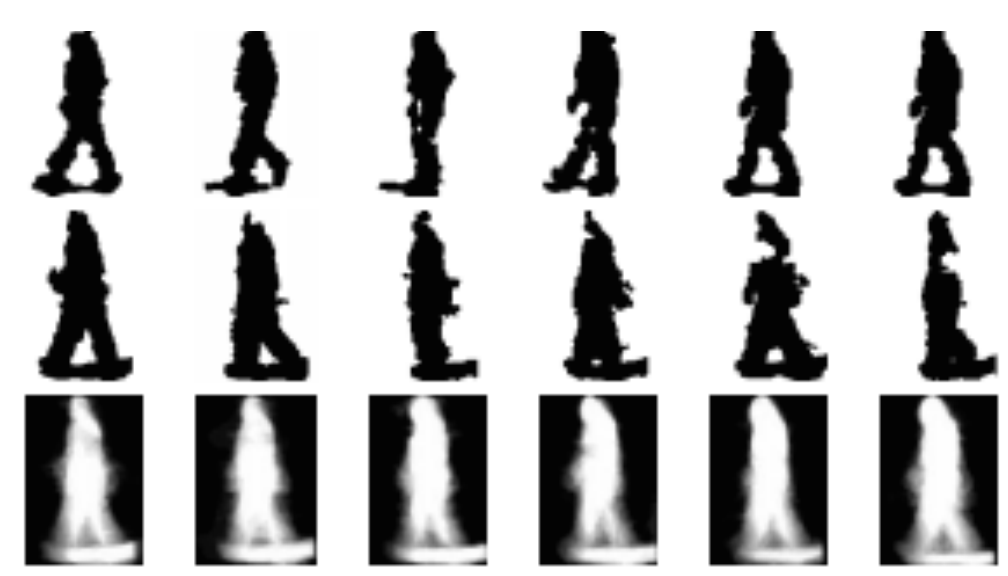}
\caption{Example of average silhouette illustrated in \citep{liu2004}.}
\label{fig:average}
\end {figure}
\end{itemize}

After we have briefly introduced the main feature families, in the following we make a non exhaustive state of the art of the main works which have been introduced in model-free gait recognition context.

\subsubsection{Model-free gait state of the art}

There exists a considerable amount of work in the context of model-free approach gait recognition. In the beginning, researchers were more focused on features based on silhouette and contour. 
 
 \citep{kale2002} introduced a method that directly incorporates the structural and transitional knowledge about the identity of the person performing the activity. They used the width of the outer contour of the binarized silhouette of a walking person as features. Hidden Markov Model (HMM) is used for classification. \citep {collins2002}  have presented a simple method  based on matching 2D silhouettes extracted from key frames across a gait cycle sequence (information such as body height, width, stride length and amount of arm swing is implicitly captured). These key frames are compared to training frames using the correlation and subject classification is performed by nearest-neighbor among correlation scores. \citep{wang2003b} introduced a method based on statistical shape analysis. They represented a gait sequence by the so called "eigenshape" signature based on Procrustes analysis \citep{kent1992new}, which implicitly captures the structural shape cue of the walking subject. The similarity between signatures is measured by Procrustes mean shape distance \citep{kent1992new} and the classification is performed based on nearest-neighbor. \citep {lee2007} suggested a novel Shape Variation-Based Frieze Pattern (SVB frieze pattern) gait signature which captures horizontal and vertical motion of the walking subject over time. It is calculated by projecting pixel values of the difference between key frames along horizontal or vertical axes. For recognition they have defined a cost function for matching. \citep {hayfron2003} suggested a contour representation by analyzing the symmetry of human motion. The symmetry operator, essentially forms an accumulator of points, which are measures of the symmetry between image points to give a signature. Discrete Fourier transform of the signature and nearest-neighbor were used for classification.

Some works tried to find good and suitable feature representation spaces for the extracted contour and silhouette features based on supervised and unsupervised representation learning techniques \cite{rida2018comprehensive}. \citep{wang2003} proposed a method able to implicitly capture the structural and transitional characteristics of gait. In this method, the 2D silhouette images are mapped into a 1D normalized distance signal by contour unwrapping with respect to the silhouette centroid (the shape changes of these silhouettes over time are transformed into a sequence of 1D distance signals to approximate temporal changes of gait pattern). Principal  Component Analysis (PCA) is applied to vectorized 1D distance signals to reduce the dimensionality and the similarity between two sequences is performed by Spatial-Temporal Correlation (STC) and Normalized Euclidean Distance (NED). The classification process is carried out via nearest-neighbor. \citep{benabdelkader2004} introduced a technique capable to capture 3D information (XYT) of the patterns. This is done by computing image Self Similarity Plot (SSP) defined as the correlation of all pairs of images in the sequence. Normalized SSPs containing an equal number of walking cycles and starting at the same body pose were used as features. Principal Component Analysis (PCA) and Linear Discriminant Analysis (LDA) combined with nearest-neighbor was used for classification. \citep {kobayashi2004} presented a novel method called Cubic Higher-order Local Auto-Correlation (CHLAC), which is an improved and extended version of Higher-order Local Auto-Correlation (HLAC) \citep{otsu1988}. CHLAC was proposed to extract spatial correlation in local regions. Linear Discriminant Analysis (LDA) combined with nearest-neighbor were used for classification. \citep{lu20077} proposed a gait recognition method based on human silhouettes characterized with three kinds of gait representations including Fourier and Wavelet descriptor. Independent Component Analysis (ICA) and Genetic Fuzzy Support Vector Machine (GFSVM) classifier were chosen for recognition. 
 
Recent trends seem to favor Gait Energy Image (GEI) representation suggested by \citep {han2006}. It is a spatio-temporal representation of the gait obtained by averaging the silhouettes over a gait cycle (see Section \ref{gei}). It is an effective representation, which makes a good compromise between the computational cost and the recognition performance. For the recognition step, they have used Canonical Discriminant Analysis (CDA) which corresponds to PCA followed by LDA combined with nearest-neighbor. The efficiency of the PCA+LDA strategy has been demonstrated in face recognition \citep{belhumeur1997eigenfaces}, in which PCA aims to retain the most representative information and suppress noise for object representation, while LDA aims to pursue a set of features that can best distinguish different objects. Furthermore, in the GEI based recognition, the dimensionality of the feature space is usually much larger than the size of the training set, this is known as the Under Sample Problem (USP). LDA often fails when faced the USP and one solution is to reduce the dimensionality of the feature space using PCA \citep {tao2007}.

In the literature, a considerable amount of works combined GEI features with different feature representation techniques to find suitable feature spaces. \citep{hofmann2012} extracted discriminative information from GEI based on Histogram of Oriented Gradient (HOG). CDA combined with nearest-neighbor were applied for classification. \citep{martin2014}, formulated the gait recognition problem as a bipartite ranking problem for more generalization of unseen gait scenarios. \citep{xing2016} have proposed a novel scheme which is called Complete Canonical Correlation Analysis (C3A) to overcome the shortcomings of Canonical Correlation Analysis (CCA) when dealing with high dimensional data. \citep {yu2006} applied a Template Matching (TM) on GEIs without any dimensionality reduction, and classification was out carried based on nearest-neighbor.

Motivated by the problem caused by the vectorization of the feature vectors when using conventional dimensionality reduction techniques which leads to under sample problem and the specialized structure of the extracted features (in form of second-order or even higher order tensor), tensor-based dimension reduction methods have been introduced. \citep{xu2006} used two supervised and unsupervised subspace learning methods: Coupled Subspaces Analysis (CSA) \citep{xu2004} and Discriminant Analysis with Tensor Representation (DATER) \citep{yan2005} to extract discriminative information from GEIs. \citep {tao2007} used Gabor filters to extract information from GEI templates. Motivated also by under sample problem, they developed a General Tensor Discriminant Analysis (GTDA) instead of conventional PCA as a preprocessing step for LDA. Inspired also by recent advances in matrix and tensor-based dimensionality reduction, \citep {xu2007} presented an extension of Marginal Fisher analysis (MFA) introduced by \citep{yan2005} to address the problem of gait recognition. \citep{chen2010} proposed a Tensor-based Riemannian Manifold distance-Approximating Projection (TRIMAP) framework to preserve the local manifold structure of the high-dimensional Gabor feature extracted from GEIs.  \citep{guan2015} introduced a classifier ensemble method based on the Random Subspace Method (RSM) and Majority Voting (MV). The random subspaces are constructed based on 2D Principal Component Analysis (2DPCA) and further enhanced with 2D Linear Discriminant Analysis (2DLDA). Table \ref{tab:summgei} summarizes the different features, transformations and classifiers for GEI-based gait recognition methods.


\begin{table}[h]
  \centering
  \caption {Overview of GEI-based methods (features, transformations and classifiers).}
  \noindent
\resizebox{0.8\linewidth}{!}{
  \begin{tabular}{llll}
    \multicolumn{4}{c}{}  \\
        \toprule
         \toprule
    Method & Features & Transformation  & Classification  \\
    \midrule  
     \midrule  
    \tabitem  \citep {han2006} & GEI & PCA+LDA  & nearest-neighbor \\
    \tabitem  \citep{hofmann2012} & GEI+HOG & PCA+LDA  & nearest-neighbor \\
     \tabitem \citep{martin2014} & GEI & transfer learning (RankSVM) & SVM   \\
      \tabitem \citep{xing2016} & GEI &  C3A & nearest-neighbor \\
       \tabitem \citep {yu2006} & GEI & - & nearest-neighbor \\
        \tabitem \citep{xu2006} & GEI & CSA+DATER & nearest-neighbor \\
        \tabitem \citep {tao2007} & GEI+Gabor  & GTDA+LDA & nearest-neighbor \\
        \tabitem  \citep {xu2007} & GEI & MFA &nearest-neighbor \\
         \tabitem \citep{chen2010} & GEI+Gabor  & TRIMAP & nearest-neighbor \\
          \tabitem  \citep{guan2015} & GEI   & RSM (2DPCA+2DLDA) & nearest-neighbor \\
    \bottomrule
     \bottomrule
  \end{tabular}}
  \label{tab:summgei}
\end{table}

Despite its good performances, GEI and like all features in model-free gait recognition suffers from various intra-class variations caused by different conditions such as the presence of shadows, clothing variations and carrying conditions which drastically influence the recognition performances. Silhouettes segmentation to calculate GEI and view angle variations represent further causes of the recognition errors \citep {han2006,yu2006,matovski2012,rida2014improved}. To overcome the limitations of GEI representation, several approaches have been proposed. They can be broadly organized in two groups: the first group tried to improve GEI by applying different feature selection techniques while the second introduced novel feature representations based on the gaps of GEI.

In the former,  \citep {bashir2008} suggested filter selection method which selects GEI pixels based on their intensity value. The idea is to keep the pixels with intensity value greater than a threshold and discard the remaining ones. In other terms, this method aims to select the dynamic pixels since it has been found that they are more discriminative and less sensitive to intra-class variations compared to the static ones \citep{han2006}. Remaining in the same idea of capturing dynamic information of the walking subject, \citep{bashir2010} introduced a  feature selection method named Gait Entropy Image (GEnI). It computes entropy for each pixel from GEI to distinguish static and dynamic pixels: 

\begin{equation}
\begin{split}
\begin{aligned}
\mathbf {GEnI}(x,y)=\sum_{k=1}^{K} p_{k}(x,y) \log_{2}(p_{k}(x,y))
\end{aligned}
\end{split}
\end{equation}
where $p_{k}(x, y)$ is the probability that the pixel $(x,y)$ takes the $k^{th}$ value in an entire gait cycle. The GEnI represents in this case a measure of feature significance or importance since the dynamic pixels (with high entropy value) are less sensitive to different intra-class variations. Pixels with greater entropy value than a threshold are kept when others are discarded. \citep {dupuis2013} introduced an embedded feature selection method based on Random Forest (RF) feature ranking algorithm in order to select features maximizing the recognition accuracy. To avoid the overfitting of the selected features to a specific training dataset, they divided the initial dataset into training, validation and testing datasets. Random Forest feature rank was applied to GEIs on validation dataset and the features were ranked based on their importance. Optimal feature subset was selected based on forward and backward selection algorithms. \citep {rida2015} learned a mask based on the pixel variations. The mask takes the value $1$ for the selected features and $0$ otherwise. The role of the mask is to select GEI features with low variations over time. In all previously introduced methods, CDA of the selected GEI pixels combined with nearest-neighbor were applied for recognition. Recently, \citep{rida2016} introduced a wrapper feature selection technique based on Modified Phase-Only Correlation (MPOC) matching algorithm to select the discriminative human body-part. The classification was carried out based on nearest-neighbor. 

In the introduced features to cope against the gaps of GEI, \citep{bashir2009} suggested a gait representation by a weighted sum of the optical flow corresponding to each direction of human motion. Because of the lack o robustness of GEI towards the appearance changes and ability of the Shannon Entropy to encode the randomness of pixel values in the silhouette images over a complete cycle, \citep {jeevan2013} introduced a novel temporal feature representation as an extension of GEI representation named Gait Pal and Pal Entropy Image (GPPE). It is calculated based on Pal and Pal Entropy \citep{pal1991entropy}:

\begin{equation}
\begin{split}
\begin{aligned}
\mathbf {GPPE}(x,y)=\sum_{k=1}^{K} p_{k}(x,y) e^{(1-p_{k}(x,y))}
\end{aligned}
\end{split}
\end{equation}
where $p_{k}(x, y)$ is the probability that the pixel $(x,y)$ takes the $k^{th}$ value. PCA followed by SVM has been used for recognition. \citep {kusakunniran2014a, kusakunniran2014b} proposed a new method for gait recognition which constructs new gait features directly from a raw video. The proposed gait features are extracted in the spatio-temporal domain. The Space-Time Interest Points (STIPs) are detected from a raw gait video sequence. They represent significant movements of human body along both spatial and temporal directions. Then, HOG and Histogram of Optical Flow (HOF) are used to describe each detected STIP. Finally, a gait feature is constructed by applying Bag of Words (BoW) on a set of HOG/HOF-based STIP descriptors from each gait sequence. Nearest-neighbor and SVM has been respectively used for classification. \citep{hu2013} used Local Binary Pattern (LBP) of optical flow as features and the classification is carried out based on Hidden Markov Model (HMM). \citep {rokanujjaman2015} used frequency domain-based gait entropy features (EnDFT) calculated by applying Discrete Fourier Transform (DFT) to GEIn. To further improve the accuracy of the proposed method, a wrapper feature selection technique has been applied. PCA combined with nearest-neighbor were used for classification.

Finally, in recent years, researchers started to have an increasing interest for gait recognition in view angle variations. \citep {choudhury2015}  introduced a two-phase View-Invariant Multiscale Gait Recognition method (VI-MGR) which is robust to variation in clothing and presence of carried items. In phase 1, VI-MGR uses the entropy of the limb region of the gait energy image (GEI) combined with 2DPCA and nearest-neighbor to determine the matching training view of the query testing GEI. In phase 2, the query subject is compared with the matching view of the training subjects using multiscale shape analysis and  ensemble classifier.

In the following we propose a novel method capable to address the problem of intra-class variations caused by carrying conditions, clothing and view-angle variations. The method represents our major contribution for gait based recognition.

\section{Body-part segmentation for improved gait recognition}
\subsection{Introduction}

Among the available feature representations we choose GEI which is an effective representation making good compromise between the computational cost and the recognition performance  \citep{bashir2010,dupuis2013,rida2016robust}. However it has also been shown that the GEI suffers from intra-class variations caused by different conditions which affect the recognition accuracy. One possible solution to tackle this problem is to focus only on dynamic parts of GEI which has been proven to be less sensitive to intra-class variations  \citep {bashir2009,dupuis2013,rida2017improved}.

\begin{figure}[!h]
\centering
\includegraphics [width= 8 cm] {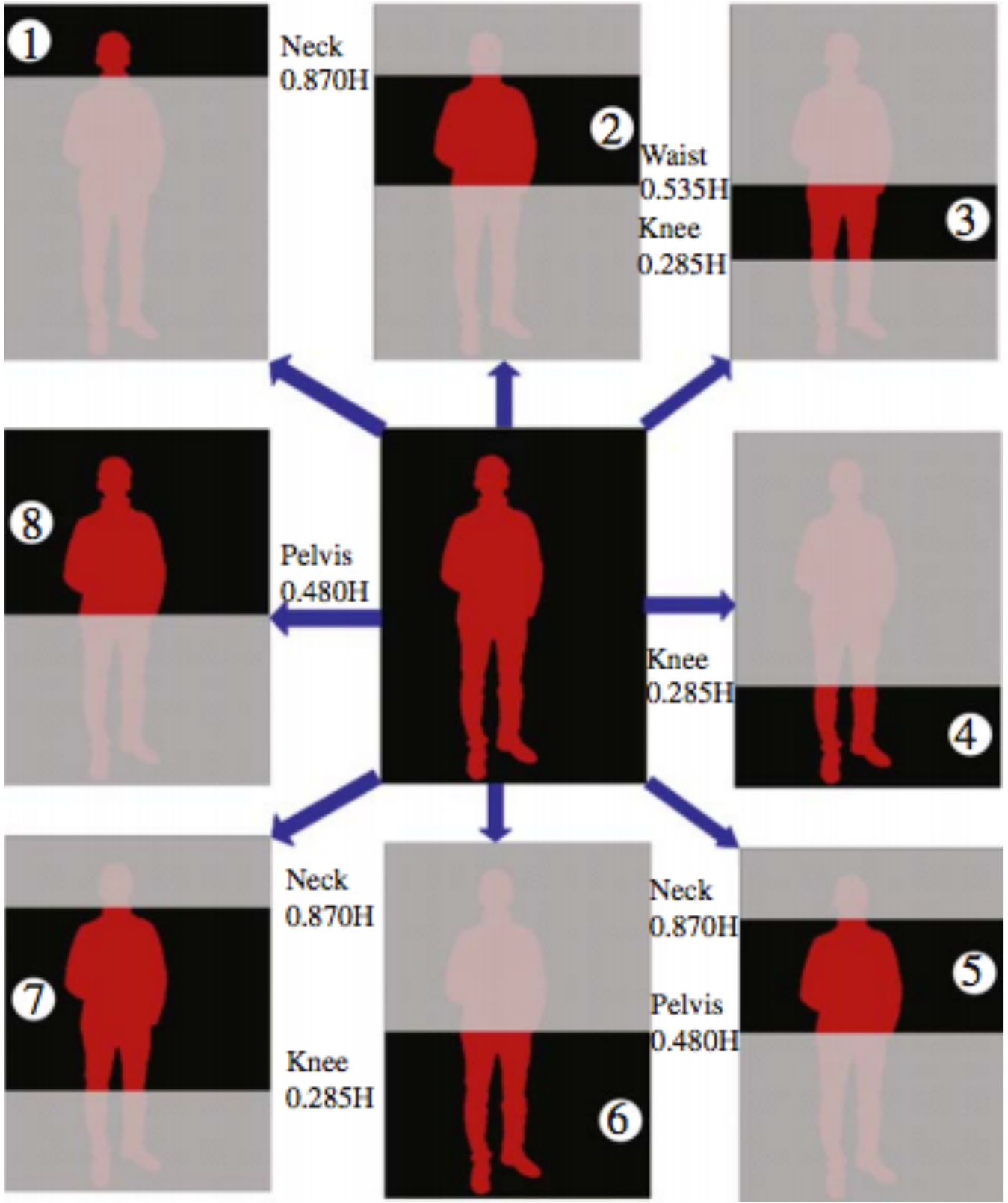}
\caption{Estimation of the body-parts based on predefined anatomical knowledge in \citep{hossain2010}.}
\label{fig:atlab}
\end {figure}

\begin{figure}[!h]
\centering
\includegraphics [width= 10 cm] {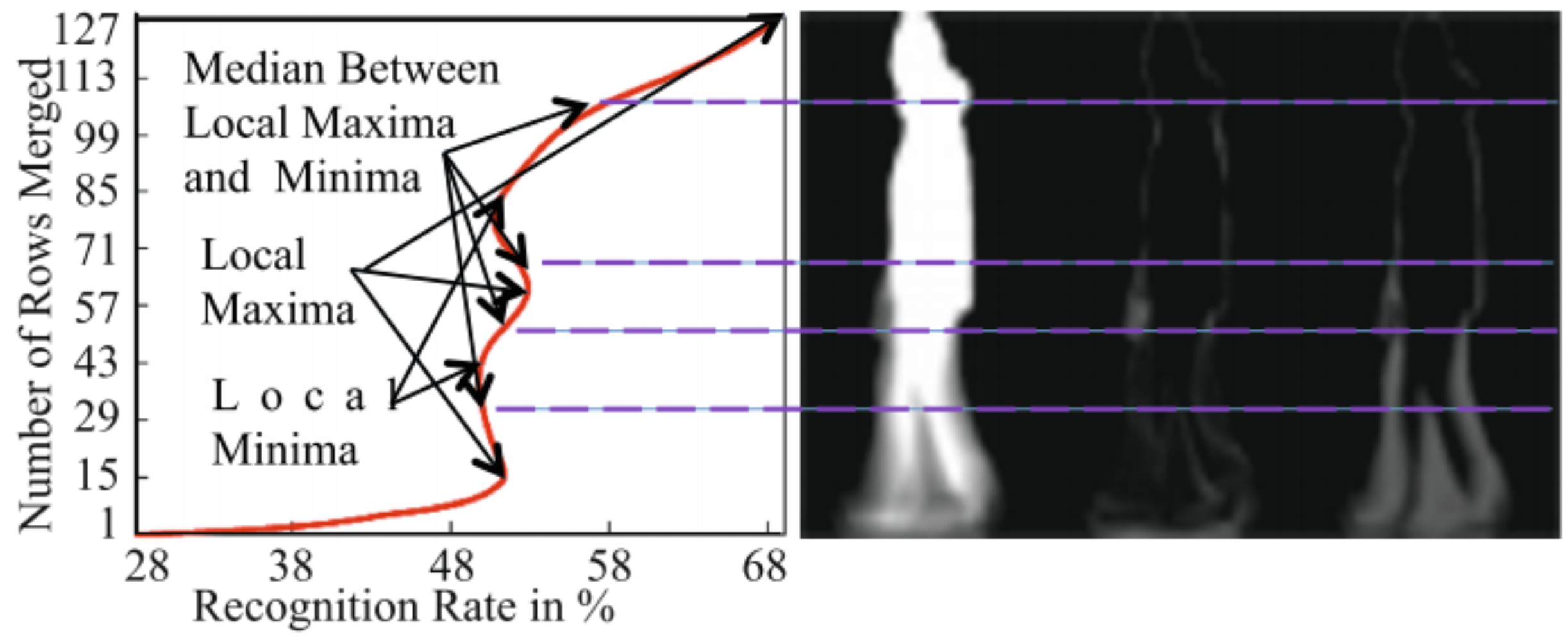}
\caption{Estimation of the body-parts based on recognition accuracy in \citep {rokanujjaman2015}.}
\label{fig:rokanujjama}
\end {figure}

In our work we propose to automatically select the dynamic body-parts contrary to the existing methods in the literature which tried to select the body-parts based on predefined anatomical properties of the human body \cite{rida2016human}. For instance in \citep{hossain2010} for a body height $H$, the human body is segmented according to the vertical position of the neck $(0.87 H)$, waist $(0.535 H)$, pelvis $(0.48 H)$, and knee $(0.285 H)$ as is shown in Figure \ref{fig:atlab}. In some other works, the human body-parts were estimated empirically,  such as in \citep {bashir2008, rokanujjaman2015} where they defined each row of the GEI as a new feature unit and tried different combinations of the new feature units which maximize the recognition accuracy as is shown in Figure \ref{fig:rokanujjama}.

\begin{figure}[!h]
\centering
\begin{tikzpicture}  [thick,scale=1, every node/.style={scale=0.9}]
\tikzstyle{box} = [rectangle,draw,thick,align=center,minimum height=10mm];
\tikzstyle{arrow} = [->,thick];
\node[box] (vid) {Videos};
\node [below=7mm of vid.south,anchor=south] (labvid) {$\{N \times M \times T\}^{P}$}; 
\node[box,right=11mm of vid.east,anchor=west] (gei) {GEIs};
\node [below=7mm of gei.south,anchor=south] (labgei) {$\{N \times M \}^{P}$}; 
\node[box,right=7mm of gei.east,anchor=west] (mbv) {Motion Based \\ Vectors};
\node [below=7mm of mbv.south,anchor=south] (labmbv) {$\{N \times 1 \}^{P}$};
\node[box,right=7mm of mbv.east,anchor=west,minimum height=30mm] (data) {$\quad\ldots\quad$};

	\path (data.north west)--(data.north east) coordinate[pos=0.1] (lu1);
	\path (data.south west)--(data.south east) coordinate[pos=0.1] (ld1);
	\draw (lu1)--(ld1);
	\path (data.north west)--(data.north east) coordinate[pos=0.2] (lu2);
	\path (data.south west)--(data.south east) coordinate[pos=0.2] (ld2);
	\draw (lu2)--(ld2);
	\path (data.north west)--(data.north east) coordinate[pos=0.9] (lu3);
	\path (data.south west)--(data.south east) coordinate[pos=0.9] (ld3);
	\draw (lu3)--(ld3);	
	\node[above=-1mm of data.north west,anchor=south] {$1$};
	\node[above=-1mm of data.north east,anchor=south] {$P$};
	\node[left=-1mm of data.north west,anchor=east] {$1$};
	\node[left=-1mm of data.south west,anchor=east] {$N$};

\node[box,right=7mm of data.east,anchor=west,minimum height=30mm] (gfl) {Group\\Fused\\Lasso};
\node[right=7mm of gfl.east,anchor=west,inner sep=0] (pict) {\includegraphics[height=30mm]{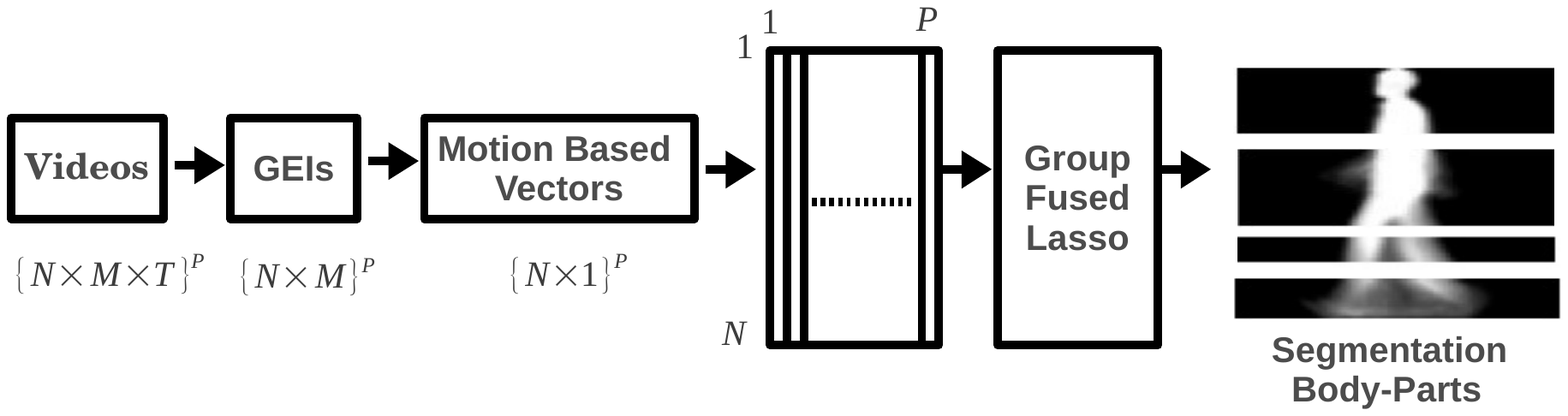}};
\draw[arrow] (vid)--(gei);
\draw[arrow] (gei)--(mbv);
\draw[arrow] (mbv)--(data);
\draw[arrow] (data)--(gfl);
\draw[arrow] (gfl)--(pict);

\end{tikzpicture}
\caption{Processing flow of body segmentation into parts based on group fused Lasso of motion.}
\label{fig:feature_selection_contribution}
\end {figure}

\citep{foster2003} used horizontal and vertical masks to capture both horizontal and vertical motion of the walking subject. They have found that the gait of an individual is characterized much more by the horizontal than the vertical motion. Furthermore, they pointed out that the horizontal motion is more reliable to represent the characteristic of gait. Therefore, instead of estimating the motion of each pixel \citep{bashir2010}, we propose to estimate the horizontal motion by taking the Shannon entropy of each row from the GEI. The resulting column vector is named as motion based vector. Group Fused Lasso is applied to the motion based vectors to segment the human body into parts with coherent motion value across the subjects. The body segmentation processing flow is shown in Figure \ref{fig:feature_selection_contribution}.

\begin{figure}[!h]
\centering
\begin{tikzpicture} [thick,scale=1, every node/.style={scale=0.9}]
\tikzstyle{box} = [rectangle,draw,thick,align=center,minimum height=10mm];
\tikzstyle{arrow} = [->,thick];
\node[box] (training) {Video};
\node [below=7mm of training.south,anchor=south] (lab1) {$N \times M \times T$}; 
\node[box,right=8mm of training.east,anchor=west] (tf) {GEI};
\node [below=7mm of tf.south,anchor=south] (lab1) {$N \times M$}; 
\node[box,right=8mm of tf.east,anchor=west] (dicolearn) {Body-Part\\ Selection};
\node [below=7mm of dicolearn.south,anchor=south] (lab1) {$N^{'} \times M$}; 
\node[box,right=8mm of dicolearn.east,anchor=west] (dico) {Feature\\ Representation};
\node[box,right=8mm of dico.east,anchor=west] (dicoo) {Classification};
\draw[arrow] (training)--(tf);
\draw[arrow] (tf)--(dicolearn);
\draw[arrow] (dicolearn)--(dico);
\draw[arrow] (dico)--(dicoo);
\node [right=8mm of dicoo.east,anchor=west] (lab) {Label}; 
\draw[arrow] (dicoo)--(lab);
\end{tikzpicture}
\caption{Classification of testing samples.} 
\label{fig:clas}
\end{figure}
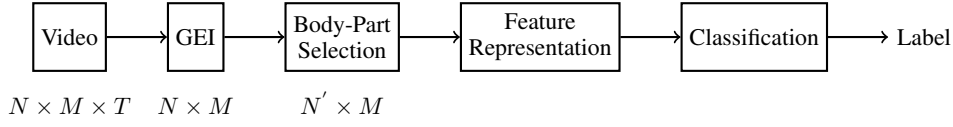

Given the segmentation process, our overall gait recognition system is described in Figure  \ref{fig:overall} and Figure \ref{fig:clas} depicting the representation learning based on the selected body-part of training data and the classification of testing samples respectively.

In the next subsections we introduce the notion of Gait Energy Image, body segmentation based on group fused Lasso of motion as well as feature representation and classification. Intensive experiments under carrying conditions, clothing and view-angle variations using CASIA gait database will be reported in comparison with state-of-the-art methods.

\begin{figure}[!h]
\centering
\begin{tikzpicture}  [thick,scale=1, every node/.style={scale=0.9}]
\tikzstyle{box} = [rectangle,draw,thick,align=center,minimum height=10mm];
\tikzstyle{arrow} = [->,thick];
\node[box] (training) {Videos};
\node [below=7mm of training.south,anchor=south] (lab1) {$\{N \times M \times T\}^{s}$}; 
\node[box,right=10mm of training.east,anchor=west] (tf) {GEIs};
\node [below=7mm of tf.south,anchor=south] (lab1) {$\{N \times M \}^{s}$}; 
\node[box,right=10mm of tf.east,anchor=west] (dicolearn) {Body-Part \\ Selection};
\node [below=7mm of dicolearn.south,anchor=south] (lab1) {$\{N^{'} \times M \}^{s}$}; 
\node[box,right=10mm of dicolearn.east,anchor=west] (dico) {Learning\\ Representation};
\draw[arrow] (training)--(tf);
\draw[arrow] (tf)--(dicolearn);
\draw[arrow] (dicolearn)--(dico);
\end{tikzpicture}
\caption{Representation learning based on the selected body-part of the training data.} %
\label{fig:overall}
\end{figure}
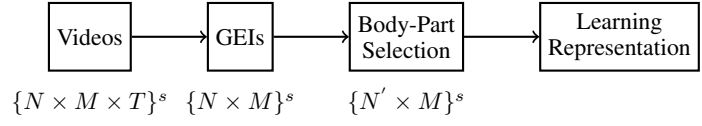


\subsection{Gait Energy Image}
\label{gei}

\begin{figure}[!h]
\centering
\subfigure[Normal Walk] {\label{fig:a}\includegraphics[width=30mm]{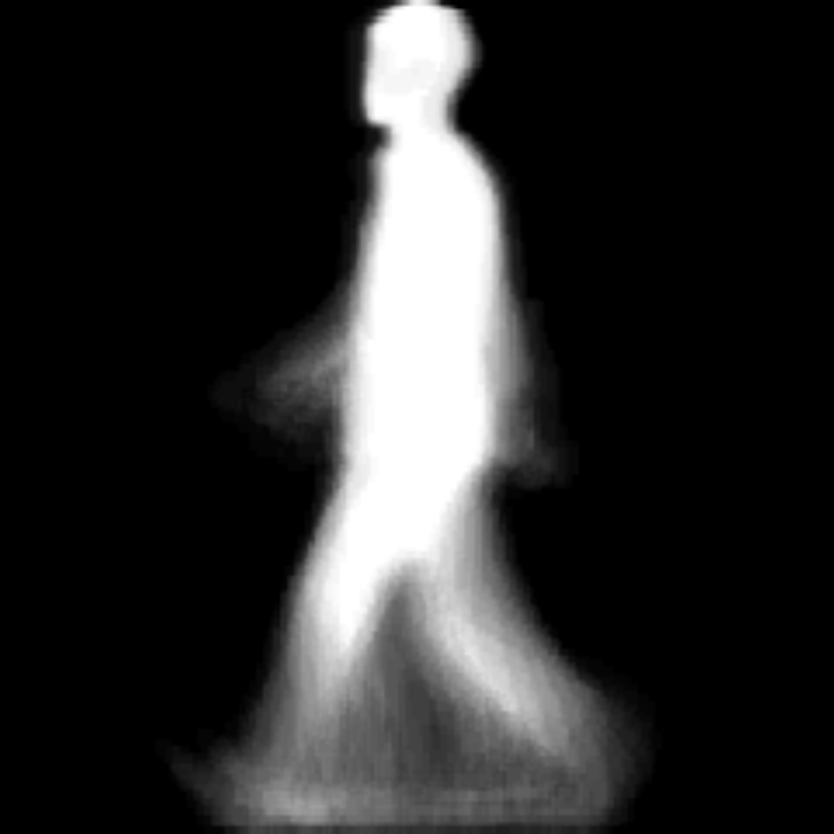}}
\subfigure[ Carrying Bag ]{\label{fig:b}\includegraphics[width=30mm]{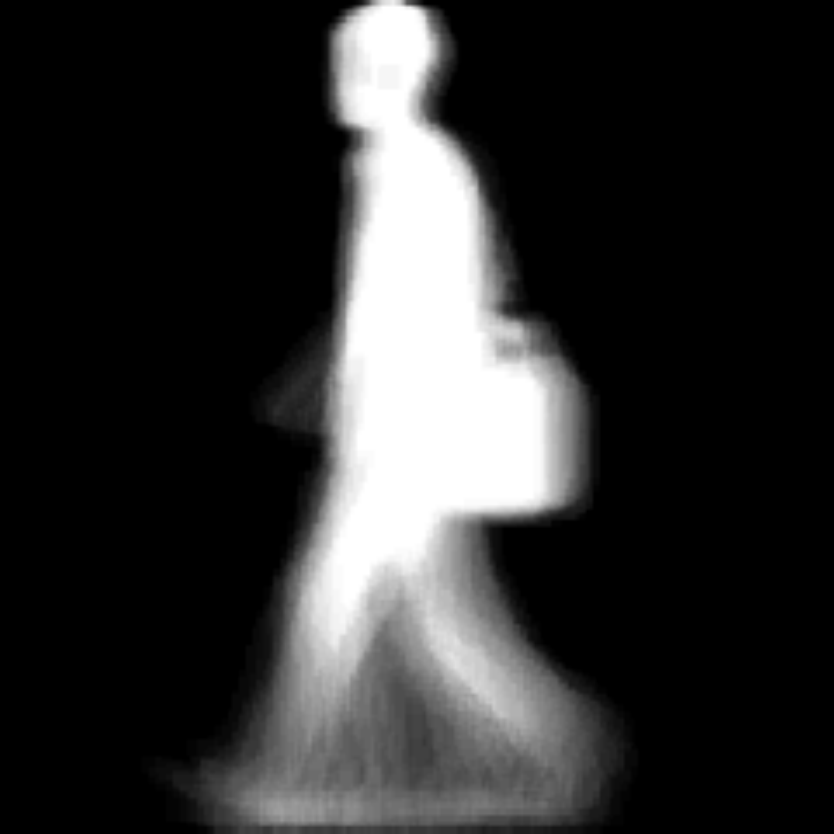}}
\subfigure[Wearing Coat ]{\label{fig:b}\includegraphics[width=30mm]{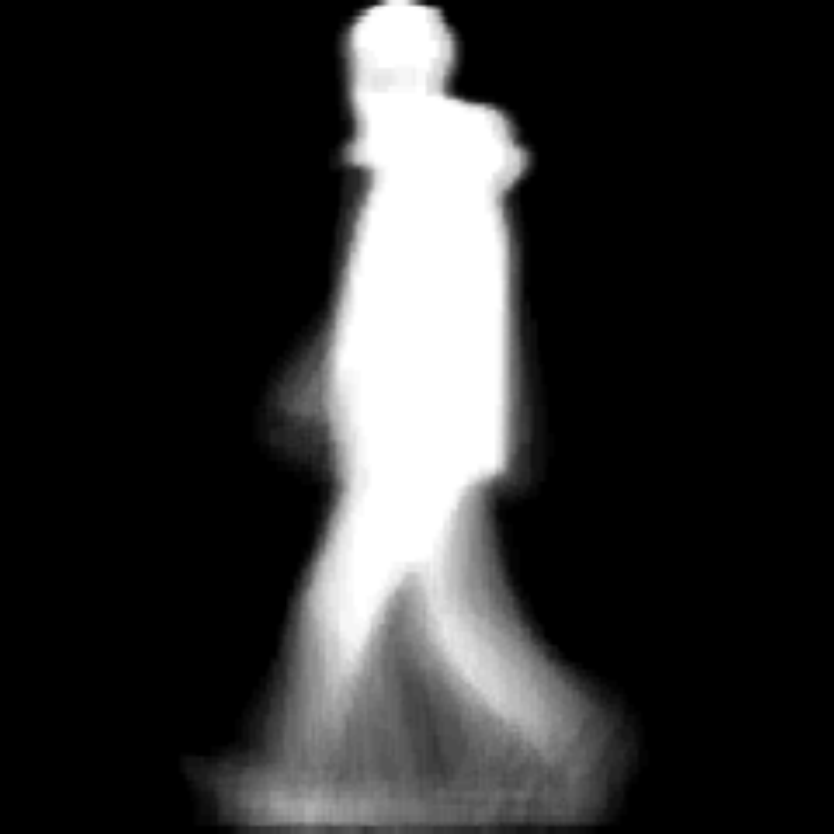}}
        \caption{Gait energy image of an individual under different conditions.}
\label{fig:gaitenergyimage}
\end{figure}

GEI is a spatio-temporal representation of gait pattern. It is a single grayscale image (see Figure \ref{fig:gaitenergyimage}) obtained by averaging the silhouettes extracted over a complete gait cycle \citep {han2006} as follows:

\begin{equation}
\begin{split}
\begin{aligned}
\mathbf {G}=\frac{255}{T}\sum_{t=1}^{T} \mathbf {B}(t)
\end{aligned}
\end{split}
\end{equation}
Here $\mathbf {G}=\{g_{i,j}\}$ is GEI, $1\leq i\leq N$ and $1\leq j\leq M$ are the spatial coordinates, $T$ is the number of the frames of a complete gait cycle, $\mathbf {B}(t)$ is the silhouette image of frame $t$. 

GEI has two main regions, the static and dynamic areas. These two areas contain different types of information. Dynamic areas are considered as being invariant to individual's appearance and most informative. Static parts despite being useful for identification they should be discarded because are greatly influenced by clothing variance \citep{bashir2010}. Static parts are localized in the top of GEI while the dynamic parts are localized in the bottom part of GEI (see Figure \ref{fig:gaitenergyimage}).

\subsection{Motion based vector}
\label{motionbasedvector}
\citep{bashir2010} tried to distinguish between the static and dynamic areas of the human body by calculating the motion of each pixel of the GEI (the motion is estimated based on Shannon entropy). As we have mentioned previously, during the walking process humans are much more characterized by horizontal than vertical motion. For the latter an horizontal motion vector is proposed that is more reliable and better characterizes the gait than the pixel-wise motion. 

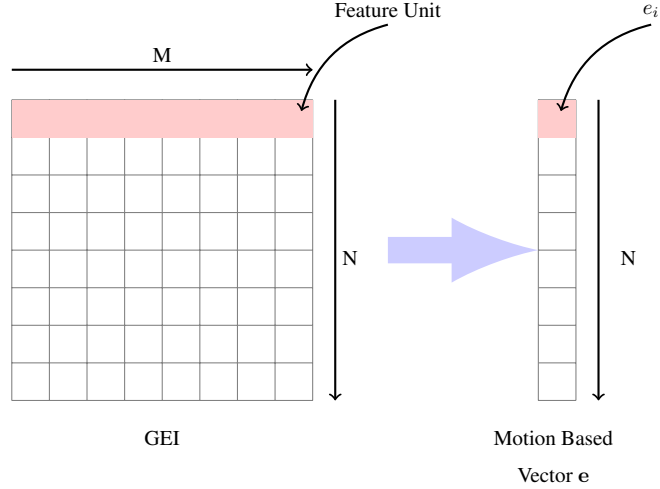
\begin{figure} [h]
\centering

\begin{tikzpicture}[thick,scale=1, every node/.style={scale=0.8}]
\draw[step=0.5cm,gray, thin] (-2,-2) grid (2,2);
\draw[thick,->] (2.3,2) -- (2.3,-2);
\draw[thick,->] (-2,2.4) -- (2,2.4); 
\draw [above] (0,2.4) node{M};
\draw [above] (2.5,-0.30) node{N};

\draw[step=0.5cm,gray, thin] (5,-2) grid (5.5,2);
\draw[step=,gray, thin] (5,-2) grid (5,2);

\draw [below] (0,-2.3) node{GEI};
 \fill[red!20](-2,1.5) rectangle (2,2);
 \fill[red!20](5,1.5) rectangle (5.5,2);
 \draw[thick,->] (5.8, 2) -- (5.8,-2);
 \draw [above] (6.2,-0.30) node{N};
\draw[bend right,->]  (3,3) to node [auto] {} (1.85,1.85) ;
\draw [above] (3,3) node{Feature Unit};
\draw [below] (5.2,-2.3) node{Motion Based};
\draw [below] (5.2,-2.8) node{Vector $\mathbf{e}$};

\coordinate (a) at (6.5,0);
\coordinate (b) at (8.5,0);

\draw[bend right,->]  (6.5,3) to node [auto] {} (5.30,1.85) ;
\draw [above] (6.5,3) node{$e_{i}$};

\coordinate (a) at (3,0);
\coordinate (b) at (5,0);
\draw[->, >=latex, blue!20!white, line width= 10 pt]   (a) to node[black]{} (b) ;

\end{tikzpicture}
\caption{Illustration of the motion based vector.} \label{fig:tikz1}
\end{figure}

For each GEI, a motion based vector $\mathbf{e} \in \mathbb{R}^{N}$ shown in Figure \ref{fig:tikz1} is generated by computing the Shannon entropy of each row of GEI which is considered as a new feature unit. The resulting vector is named motion based vector. The entry $i$ of the motion based vector $\mathbf{e}$ is given by:

 \begin{equation}
e_{i}=-\sum_{k=0}^{255}p_{k}^{i} \log_{2} p_{k}^{i}
    \end{equation}
where $p_{k}^{i}$ is the probability that the pixel value $k$ occurs in the $i^{th}$ row of image $\mathbf {G}$, which is estimated by:

    
 \begin{equation}
 \displaystyle p_{k}^{i}=\frac{\# (g_{i,j}=k)}{M} \qquad  \displaystyle  \forall j \in 1, \cdots, M  \hspace{3 mm}   \forall i \in 1, \cdots, N  \\  
\label{eq}
\end{equation} \\
where $\# (g_{i,j}=k)$ counts the number of pixels containing the value $k$.        

\subsection {Group fused lasso for body-part segmentation}
\label{lasso}
\begin{figure}[!h]
\centering
\includegraphics [width= 8 cm] {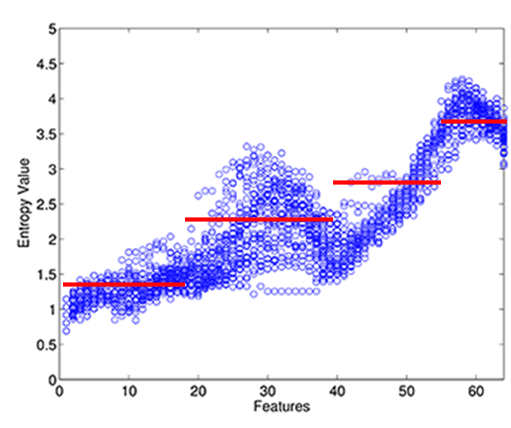}
\caption{Example of shared change points across motion based vectors. Blue dots correspond to the motion-based vectors and red lines stand for the piecewise approximation.}
\label{fig:uns}
\end {figure}

Let $P$ motion based vectors $\{\mathbf{e}_{k}\}_{k=1}^{P}$ of $P$ GEIs stored in $N \times P$ matrix $\mathbf {E}$. The aim is to detect the shared change-point locations across all motion based vectors $ \{\mathbf{e}_{k}\}_{k=1}^{P}$ (see Figure \ref{fig:uns}) by approximating matrix $\mathbf{E} \in \mathbb{R}^{N \times P}$ by a matrix $\mathbf{V} \in \mathbb{R}^{N \times P}$ of piecewise-constant vectors that share change points. This can be achieved by resolving the following convex optimization problem:

\begin{equation}
\begin{split}
\begin{aligned}
\displaystyle \min_{\mathbf {V}\in \mathbb {R}^{N \times P}} \norme{\mathbf{E} - \mathbf{V}}_{F}^{2}+ \lambda \hspace{0.3mm} \sum_{i=1}^{N-1} \norme{\mathbf{v}_{i+1,\LargerCdot}-\mathbf{v}_{i,\LargerCdot}}_{1}
\end{aligned}
\end{split}
\label{glasso}
\end{equation}
where $\mathbf{v}_{i,\LargerCdot}$ is the $i$-th row of $\mathbf{V}$ and $\lambda > 0$ a regularization parameter. Intuitively, increasing $\lambda$ enforces many increments $\mathbf{v}_{i+1}-\mathbf{v}_{i}$ to converge towards zero. This implies that the position of non-zeros increments will be same for all vectors $\mathbf{e}_{k}$. Therefore, the solution of (\ref{glasso}) provides an approximation of $\mathbf{E}$ by a matrix $\mathbf{V} $ of piecewise-constant vectors with shared change-points. The problem (\ref{glasso}) is reformulated as a group Lasso regression problem as follows: 
\begin{equation}
\begin{split}
\begin{aligned}
\displaystyle \min_{\mathbf {\beta}\in \mathbb{R}^{(N-1) \times P}}  \norme{\mathbf{\overbar{E}} - \mathbf{\overbar{X}}\pmb{\beta}}^{2}_{F}+ \lambda \hspace{0.3mm} \sum_{i=1}^{N-1} \norme{\pmb {\beta_{i,\LargerCdot}}}_{1}
\end{aligned}
\end{split}
\label{gglasso}
\end{equation}
where $\overbar{\mathbf {X}}$ and $\overbar{\mathbf {E}}$ are obtained by centering each column from $\mathbf {X}$ and $\mathbf{E}$ knowing that:

\begin{equation}
     \left\{
    \begin{array}{l}
   \mathbf{X} \in \mathbb{R}^{N \times (N-1)}; \hspace{8 mm}    x_{i,j} = \begin{cases} 1 \hspace{3 mm} \mbox{for} \hspace{3 mm} i>j  \\ 0 \hspace{3 mm} \mbox{otherwise} \end{cases}
        \\
        \\
      \pmb {\beta_{i,\LargerCdot}}= \mathbf{v}_{i+1,\LargerCdot}-\mathbf{v}_{i,\LargerCdot}
    \end{array}\right.
    \end{equation}
For more details about the reformulation we refer the reader to the appendix and \citep{bleakley2011}. The problem (\ref{gglasso}) can be solved based on the group LARS described in \citep{yuan2006} which approximates the solution path with a piecewise-affine set of solutions and iteratively finds change-points. Note that the segmentation borders are located on non null values of $\pmb{\beta}$.

\subsection{Feature representation and classification}
\label {cda}
The body-part segmentation provides the relevant features that characterize best dynamic information in the GEI images. The selected parts of the GEI are further used for adequate feature representation followed by the classification scheme \cite{rida:tel-01515364}.
Feature representation is carried out based on Canonical Discriminant Analysis (CDA) which was initially introduced in gait recognition by \citep{huang1999}. CDA corresponds to Principal Component Analysis (PCA) followed by Linear Discriminant Analysis (LDA). The efficiency of the PCA+LDA strategy has been demonstrated in several applications such as face recognition  \citep{belhumeur1997eigenfaces}, in which PCA aims to retain the most representative information and suppress noise \citep{jiang2009,jiang20111}, while LDA aims to determine features which maximize the distance between classes and preserve the distance inside the classes. Furthermore, in the GEI based recognition, the dimensionality of the feature space is usually much larger than the size of the training set. Hence applying CDA help avoiding the overfitting phenomenon.

In our work CDA is applied to the GEI features of the robust human body of the training dataset. As suggested by \citep{han2006} we retain $2c$ eigenvectors after applying PCA, where $c$ corresponds to the number of classes. The classification is carried out by a nearest-neighbor classifier and the performance of our method is measured by the Correct Classification Rate (CCR) which is the ratio of the number of correctly classified samples over the total number of samples.

\subsection{Experiments}
In this section, we evaluate our proposed gait recognition methodology. We introduce first the dataset for this sake and hence the different experiments performed on it as well as the obtained results.

\subsubsection{Dataset}
\begin{figure}[!h]
\centering
\includegraphics [width= 13.3 cm] {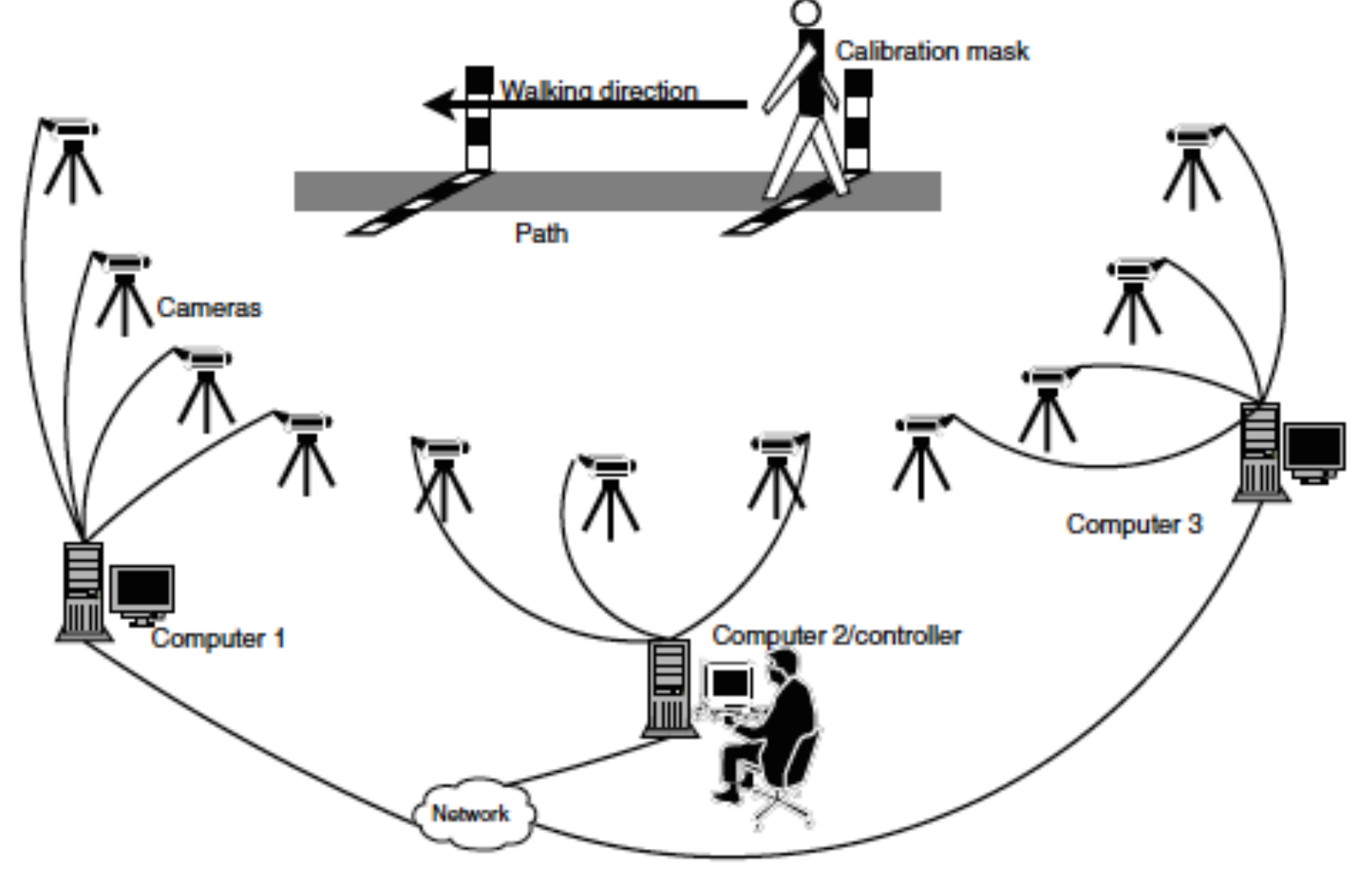}
\caption{Set-up for gait data collection in CASIA  \citep{yu2006}.}
\label{fig:setup}
\end {figure}

The proposed method is tested on CASIA dataset B \footnote{\url{http://www.cbsr.ia.ac.cn/english/Gait\%20Databases.asp}} \citep{yu2006}  to evaluate its ability to handle the carrying, clothing and view angle variations. CASIA dataset B is a large multiview gait database created in January 2005 containing 124 subjects captured from 11 different view angles using 11 USB cameras around the left hand side of the walking subject starting from $0^{\circ}$ to $180^{\circ}$ (see Figure \ref {fig:setup}).

\begin{figure}[!h]
\centering
\includegraphics [width= 13 cm] {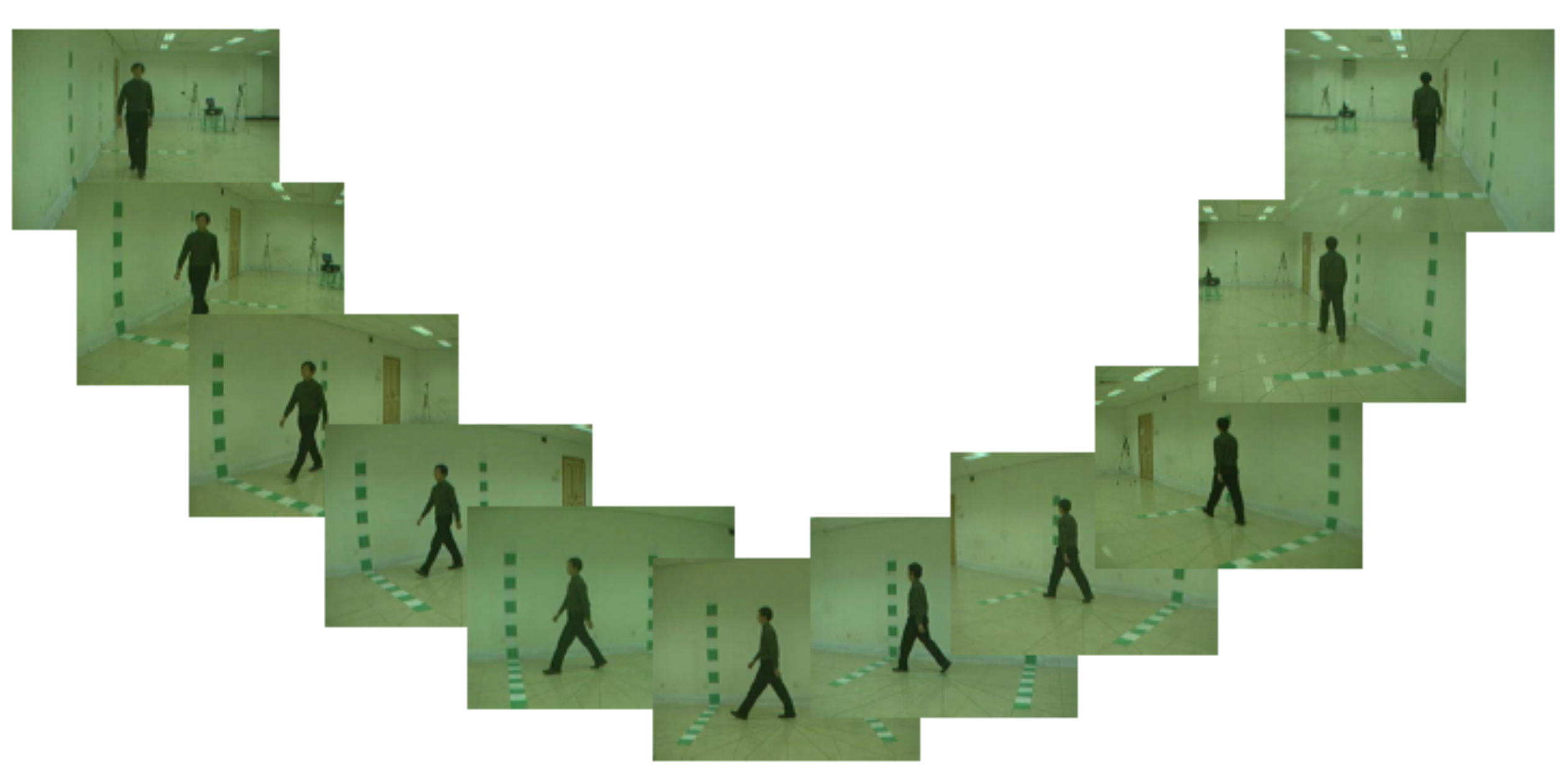}
\caption{Normal walking conditions under different view angles from $0^{\circ}$ to $180^{\circ}$  \citep{yu2006}.}
\label{fig:nl}
\end {figure}

\begin{figure}[!h]
\centering
\includegraphics [width= 15 cm] {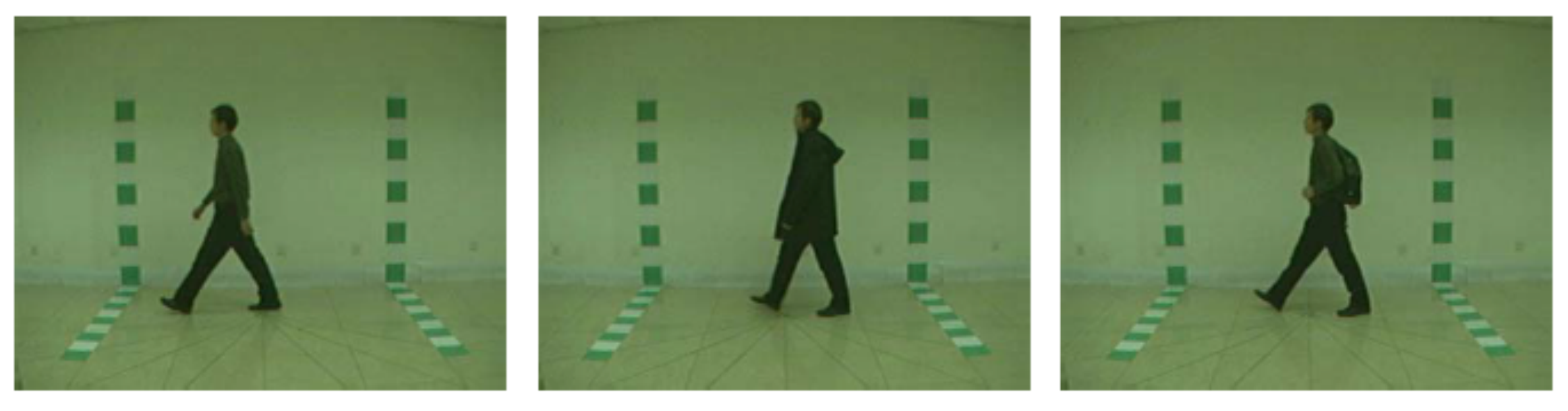}
\caption{Normal, clothing and carrying conditions under $90^{\circ}$ angle \citep{yu2006}.}
\label{fig:bgcl}
\end {figure}

Each subject is recorded six times under normal conditions (NL), twice under carrying bag conditions (CB) and twice under clothing variation conditions (CL) (see Figures \ref {fig:nl} and \ref {fig:bgcl}). 
The first four sequences of (NL) are used for training. The two remaining sequences of (NL) as well as (CB) and (CL) are used for testing normal, carrying and clothing conditions, respectively. For each sequence, GEI of size $64\times64$ is computed.

The selected robust human body-part should not be overspecialized for a specific training dataset \citep{dupuis2013}. As consequence, human body-parts are estimated on a validation dataset independent from training and testing datasets. To create our body-part selection dataset, we have randomly selected 24 GEIs for each variant (normal, carrying, clothing), hence our validation dataset contains in total $72$ GEIs. Table \ref{casia} summarizes the content of CASIA database under each view angle from $0^{\circ}$ to $180^{\circ}$.

Table \ref{partition90} and Table \ref{partitionrest} represent the data partition of the carried out experiments under $90^{\circ}$ and the other remaining view angles respectively. Contrary to $90^{\circ}$, the remaining view angles do not contain a validation set, the body parts selected for experiments under $90^{\circ}$ are kept for the other angles experiments.

\begin{table}[!h]
\centering
\caption{CASIA database content under each view angle from $0^{\circ}$ to $180^{\circ}$.}
\begin{tabular}{|*{6}{c|}}
\hline
\multicolumn{2}{|c|}{Normal} & \multicolumn{2}{c|}{Carrying conditions} & \multicolumn{2}{c|}{Clothing variation}\\ \hline
\multicolumn{1}{|c}{\# Subjects} & \multicolumn{1}{|c|}{ \# GEIs} &
\multicolumn{1}{|c}{\# Subjects} & \multicolumn{1}{|c|}{ \# GEIs} &
\multicolumn{1}{|c}{\# Subjects} & \multicolumn{1}{|c|}{ \# GEIs}  \\ \hline 
124 & 744 & 124 & 248 &124 & 248\\ \hline
\end{tabular}
\label{casia}
\end{table}

\begin{table*}[!h]
\centering
\caption{Data partition of carried out experiments under $90^{\circ}$ view.}
 \noindent
\resizebox{\linewidth}{!}{
\begin{tabular}{|*{10}{c|}}
\hline
\multicolumn{2}{|c|}{Validation set} & \multicolumn{2}{c|}{Training set} & \multicolumn{2}{c|}{Test set normal} & \multicolumn{2}{c|}{Test set carrying} & \multicolumn{2}{c|}{Test set clothing} \\ \hline
\multicolumn{1}{|c}{ \# Subjects} & \multicolumn{1}{|c|}{ \# GEIs} & 
\multicolumn{1}{|c}{\# Subjects} & \multicolumn{1}{|c|}{ \# GEIs} &
\multicolumn{1}{|c}{\# Subjects} & \multicolumn{1}{|c|}{ \# GEIs} &
\multicolumn{1}{|c}{\# Subjects} & \multicolumn{1}{|c|}{ \# GEIs} &
\multicolumn{1}{|c}{\# Subjects} & \multicolumn{1}{|c|}{ \# GEIs}  \\ \hline 
24 & 72 & 124 & 472 &124 & 248 & 124 &224 &124 &224\\ \hline
\multicolumn{2}{|c|}{24 NL, 24 CB, 24 CL} & \multicolumn{2}{c|}{472 NL} & \multicolumn{2}{c|}{248 NL} & \multicolumn{2}{c|}{224 CB} & \multicolumn{2}{c|}{224 CL} \\ \hline
\end{tabular}}
\label{partition90}
\end{table*}

\begin{table*}[!h]
\centering
\caption{Data partition of carried out experiments under view angles from $0^{\circ}$ to $72^{\circ}$ and from $108^{\circ}$ to $180^{\circ}$. }
 \noindent
\resizebox{\linewidth}{!}{
\begin{tabular}{|*{8}{c|}}
\hline
 \multicolumn{2}{|c|}{Training set} & \multicolumn{2}{c|}{Test set normal} & \multicolumn{2}{c|}{Test set carrying} & \multicolumn{2}{c|}{Test set clothing} \\ \hline
\multicolumn{1}{|c}{\# Subjects} & \multicolumn{1}{|c|}{ \# GEIs} &
\multicolumn{1}{|c}{\# Subjects} & \multicolumn{1}{|c|}{ \# GEIs} &
\multicolumn{1}{|c}{\# Subjects} & \multicolumn{1}{|c|}{ \# GEIs} &
\multicolumn{1}{|c}{\# Subjects} & \multicolumn{1}{|c|}{ \# GEIs}  \\ \hline 
 124 & 496 &124 & 248 & 124 &248 &124 &248\\ \hline
 \multicolumn{2}{|c|}{496 NL} & \multicolumn{2}{c|}{248 NL} & \multicolumn{2}{c|}{248 CB} & \multicolumn{2}{c|}{248 CL} \\ \hline
\end{tabular}}
\label{partitionrest}
\end{table*}

To sum up validation set serves for body-part selection. The retained parts are then exploited for feature representation (PCA followed by LDA) on the basis of the training set which is used as reference data for a nearest neighbor classifier. Reported performances are calculated over test set.

\subsubsection{Selected robust human body-part}

As we have already mentioned, the segmentation of the body into parts (regions of interest) and the selection of the robust part should not be overspecialized for a specific training dataset. As consequence we perform it on the validation dataset. To evaluate the robustness of our body segmentation method, we perform a without-replacement bagging of size $P=45$ GEIs from the validation dataset containing $72$ GEIs. The operation was repeated $L=5$ times, resulting in $5$ subsets of size $45$ GEIs.

\begin{figure}[!t]
\centering
\subfigure[Experiment 1]{\includegraphics[width=46mm]{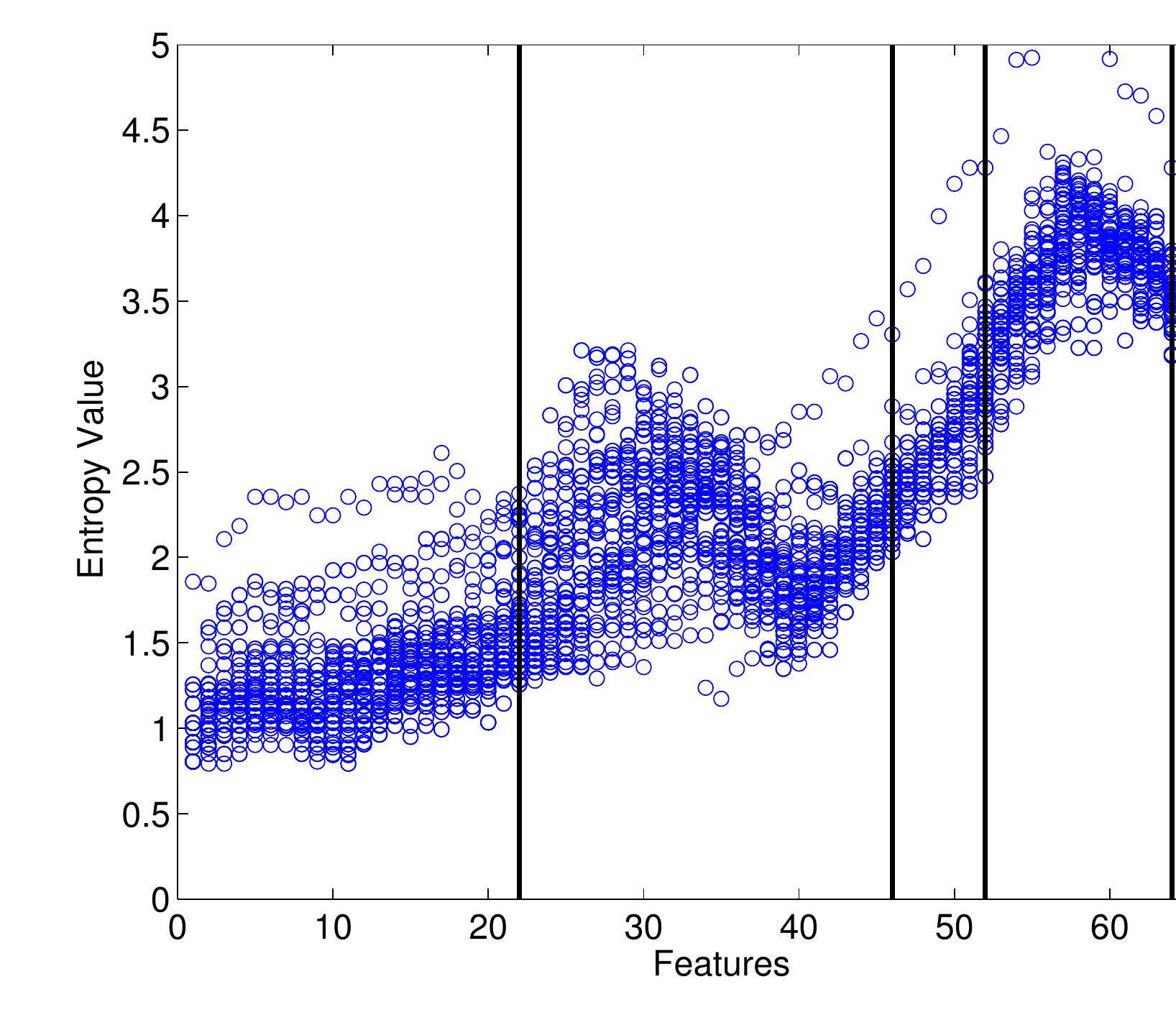}}
\subfigure[Experiment 2]{\includegraphics[width=46mm]{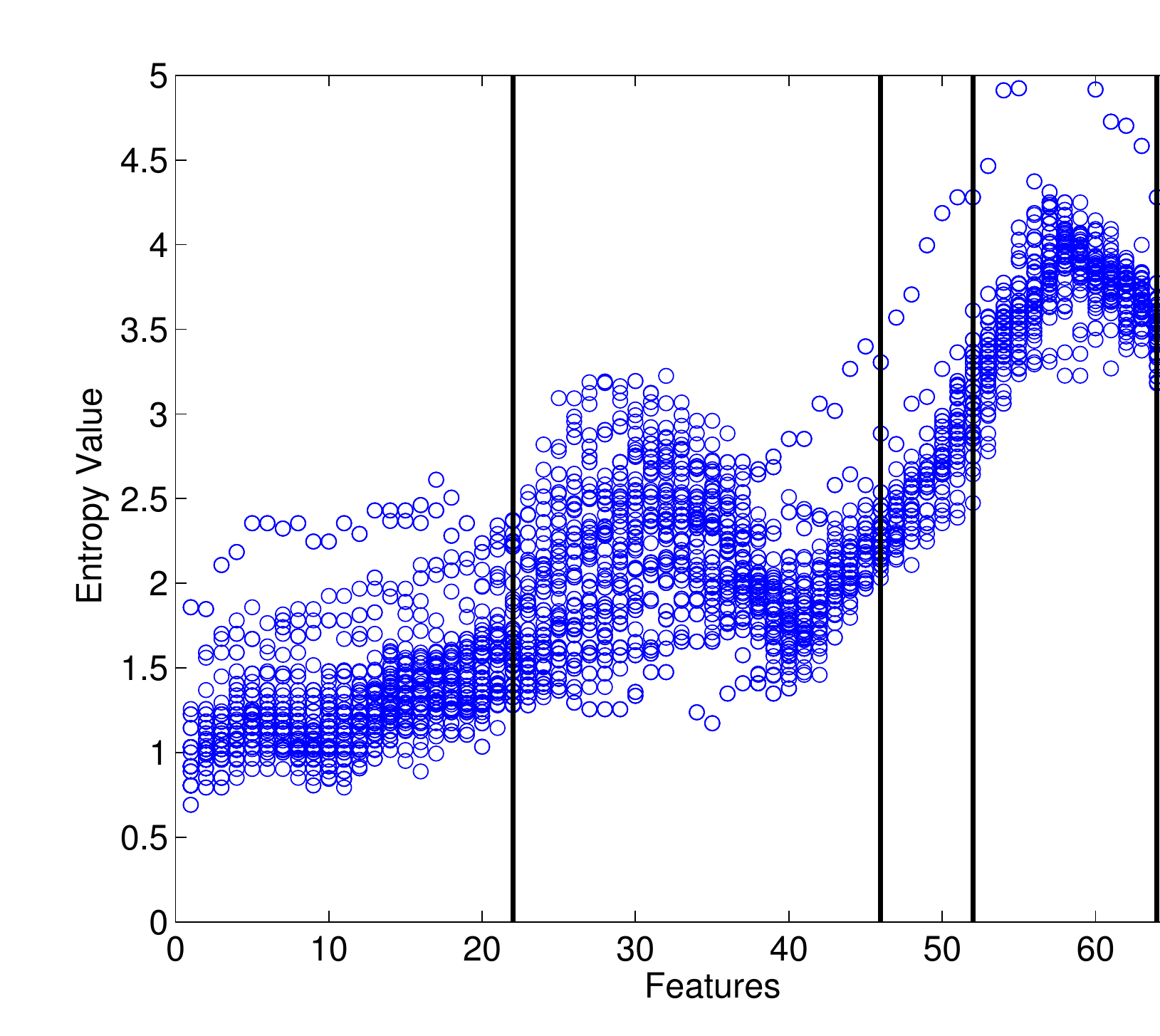}}
\subfigure[Experiment 3]{\includegraphics[width=46mm]{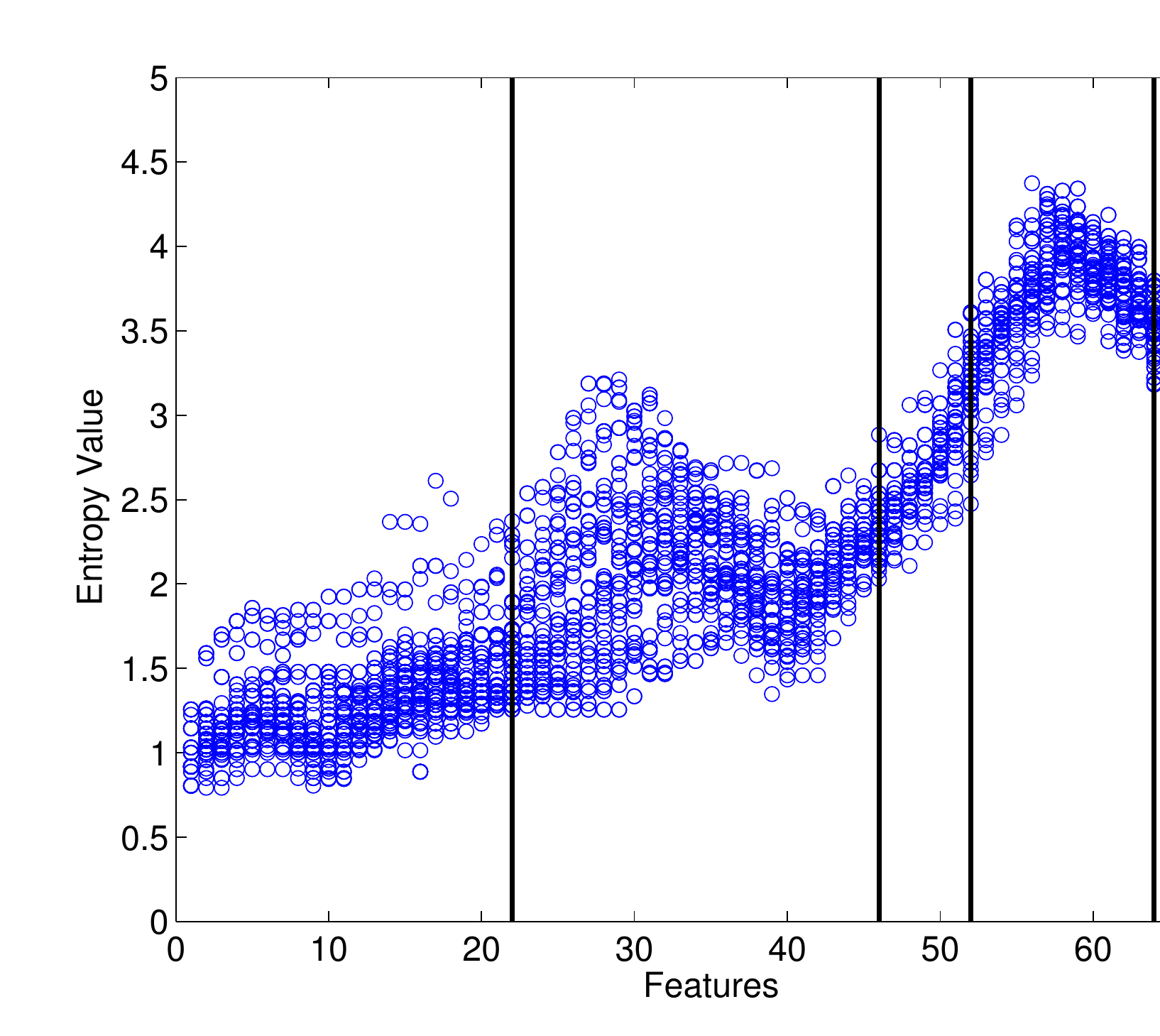}}
\subfigure[Experiment 4]{\includegraphics[width=46mm]{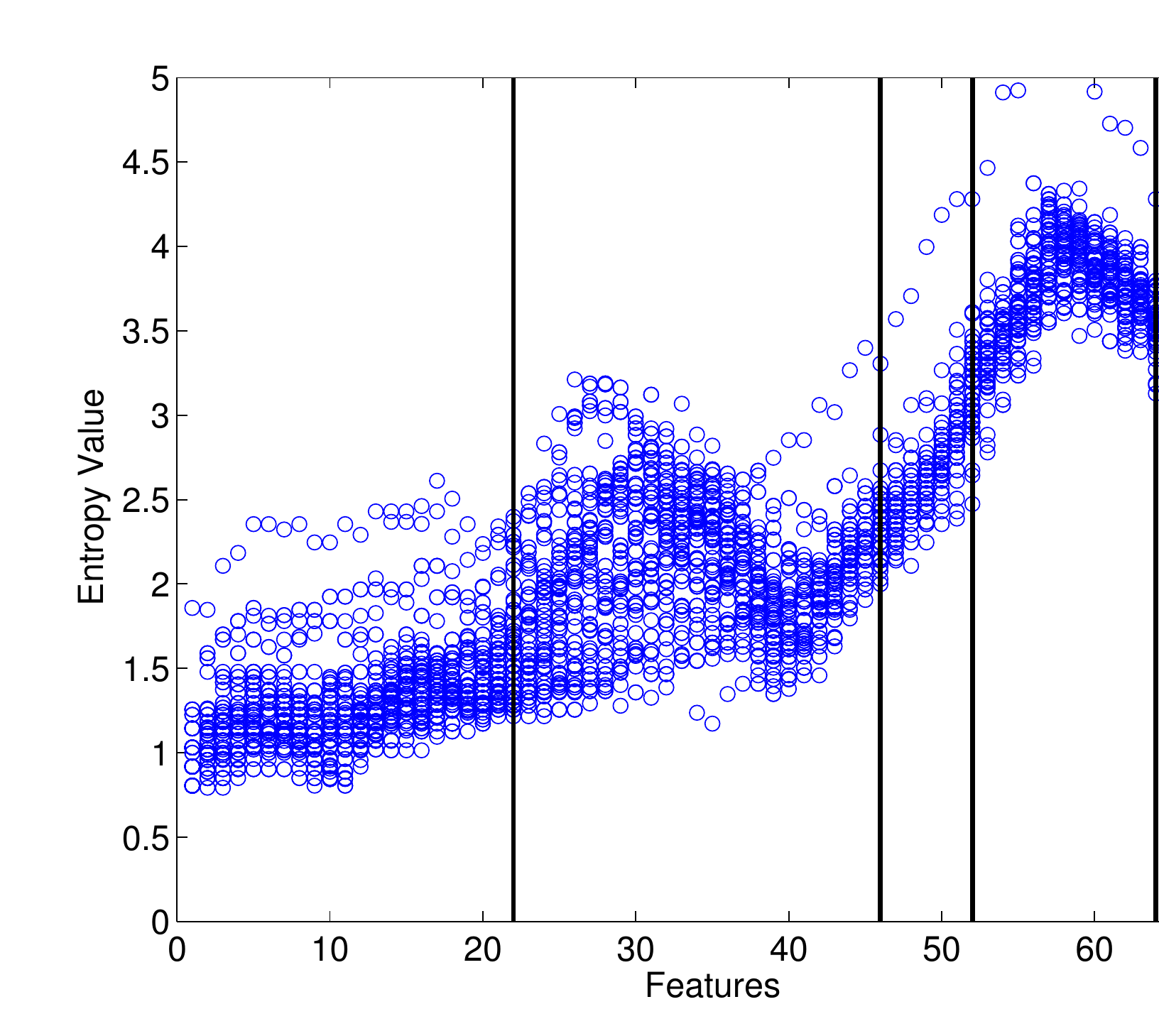}}
\subfigure[Experiment 5]{\includegraphics[width=46mm]{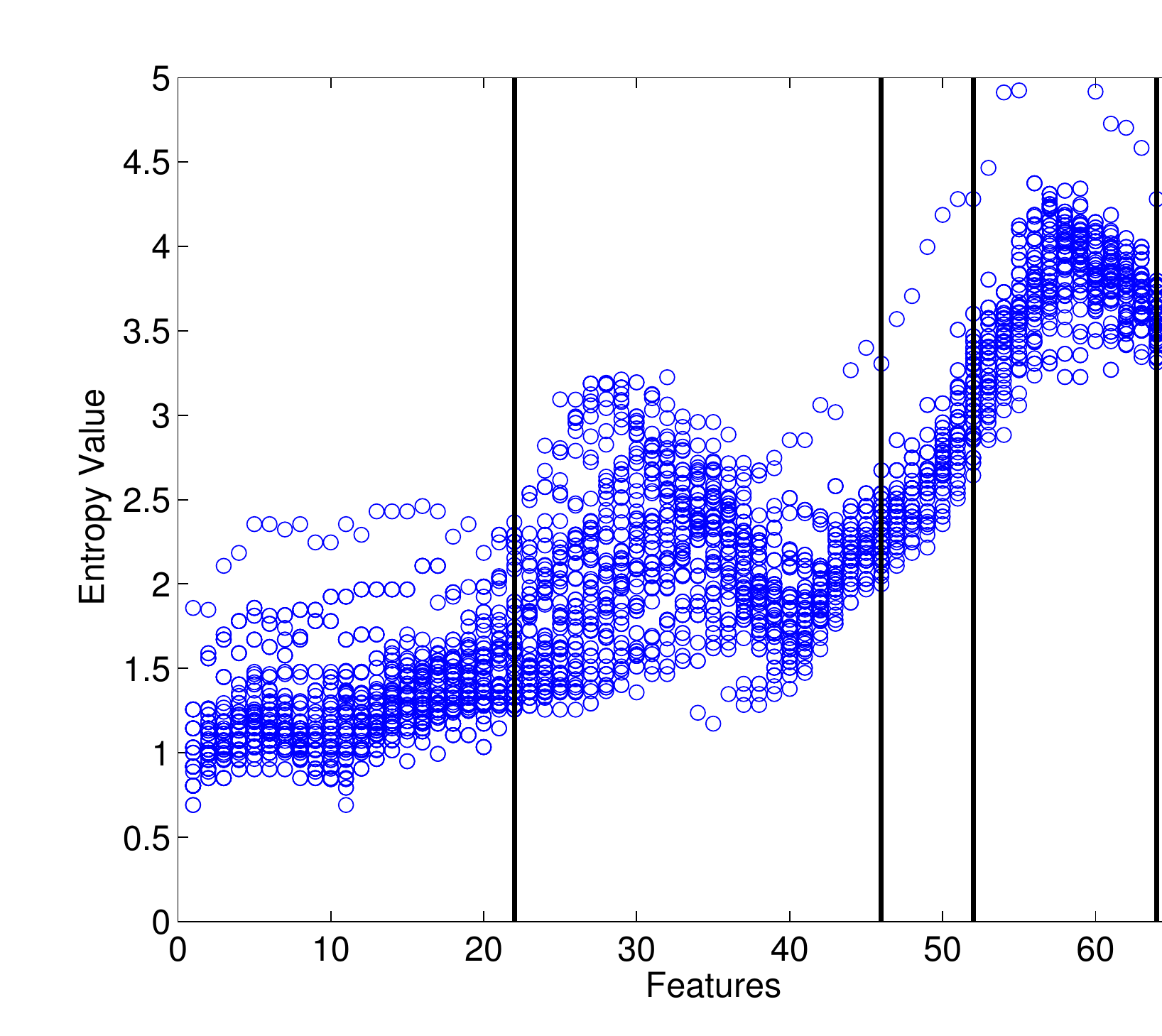}}
 \caption{Values of motion based vectors in selection datasets and parts of shared motion value separated by group fused Lasso.}
 \label{parts}
 \end{figure}

For each subset, motion based vectors are calculated and the body-parts are segmented based using group fused Lasso. Figure \ref{parts} shows the entropy value (y-axis) of all GEIs against feature index (x-axis) for the $5$ subsets. The vertical lines represent the limits of human body-parts learned by the group fused Lasso on each subset. 

\begin{figure}[!htbp]
\centering
\subfigure[Part1] {\label{fig:a}\includegraphics[width=43mm]{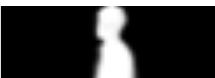}}
\subfigure[Part 2]{\label{fig:b}\includegraphics[width=43mm]{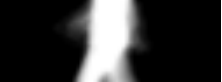}}\\
\subfigure[Part 3]{\label{fig:d}\includegraphics[width=43mm]{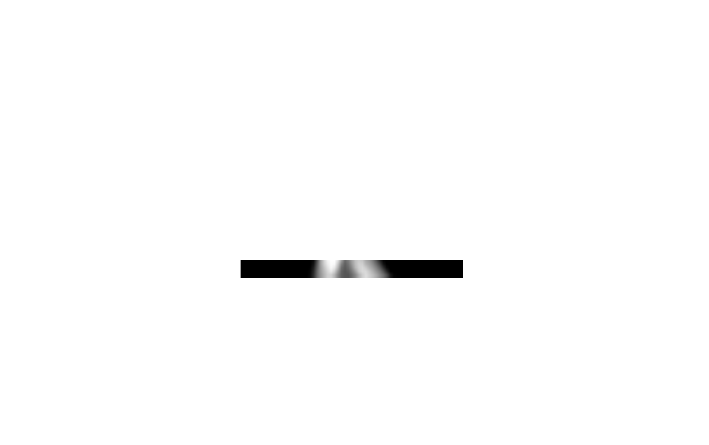}}
\subfigure[part 4]{\label{fig:c}\includegraphics[width=43mm]{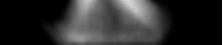}}
        \caption{Human body parts of GEI separated by group fused Lasso.}
\label{fig:geiparts}
\end{figure}

We can see in Figure \ref{parts} that the proposed method is stable and divides the body into similar parts for the $5$ subsets. It can be also seen that the group fused Lasso divides the horizontal motion of human body into $4$ parts. The corresponding parts of GEI are shown in Figure \ref {fig:geiparts}.

It has been shown that dynamic body-parts contain discriminative information to differentiate people and are robust to intra-class variations \citep{bashir2010,dupuis2013}. Based on the latter assumption, we select the body-parts with the highest motion which are supposed to cope against the intra-class variations problem. This can be seen as a filter feature selection approach since the estimated  parts by group fused Lasso are ranked according to their scores. The scores are calculated based on predefined criterion corresponding in our case to the mean entropy value of each part. The parts with the highest mean motion values are selected for classification.

From Figure \ref{parts} we can see that the parts formed by feature units (rows of GEI) from $46$ to $64$ have the highest mean motion value. They correspond to the GEI parts shown in Figures \ref{fig:d} and \ref {fig:c}. In the following we will perform experiments under different conditions using those selected parts.

\subsubsection{Effect of clothing and carrying conditions}

\begin{figure}[!h]
\centering
\subfigure[Normal Walk] {\label{fig:a}\includegraphics[width=34mm]{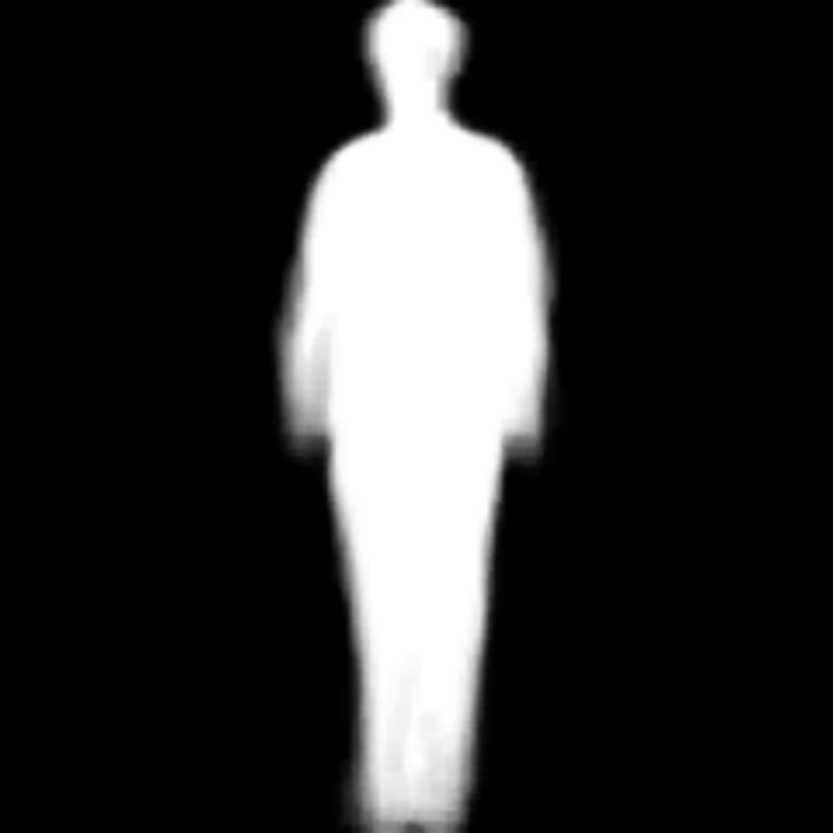}}
\subfigure[ Carrying Bag ]{\label{fig:b}\includegraphics[width=34mm]{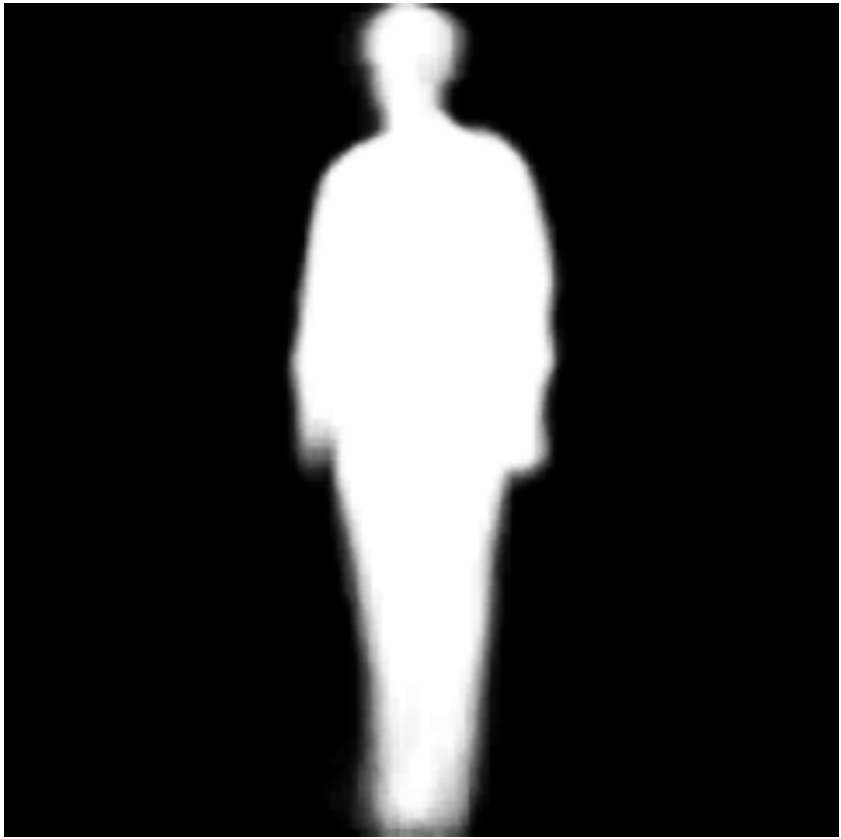}}
\subfigure[Wearing Coat ]{\label{fig:b}\includegraphics[width=34mm]{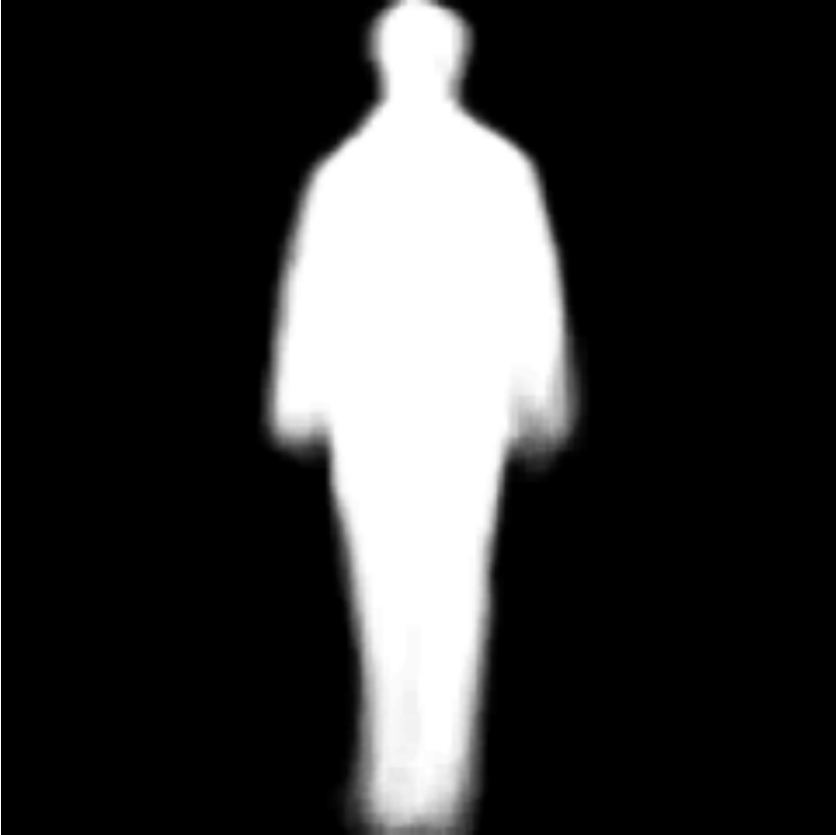}}
        \caption{Gait energy image of an individual under different conditions in frontal view.}
\label{fig:geifrontal}
\end{figure}

\begin{figure}[!h]
\centering
\subfigure[Normal Walk] {\label{fig:a}\includegraphics[width=34mm]{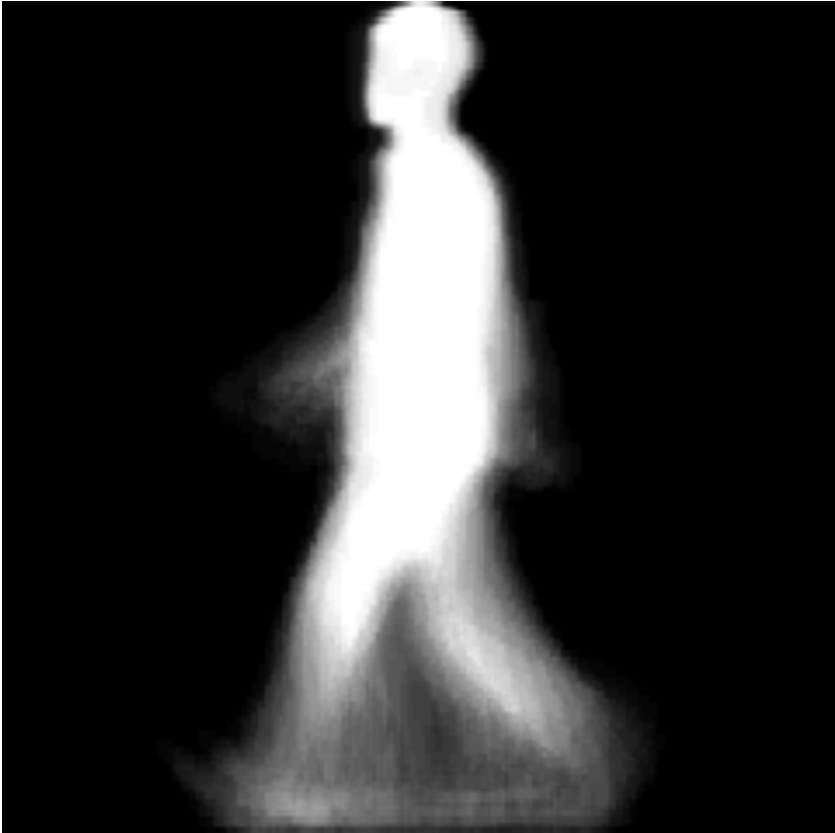}}
\subfigure[ Carrying Bag ]{\label{fig:b}\includegraphics[width=34mm]{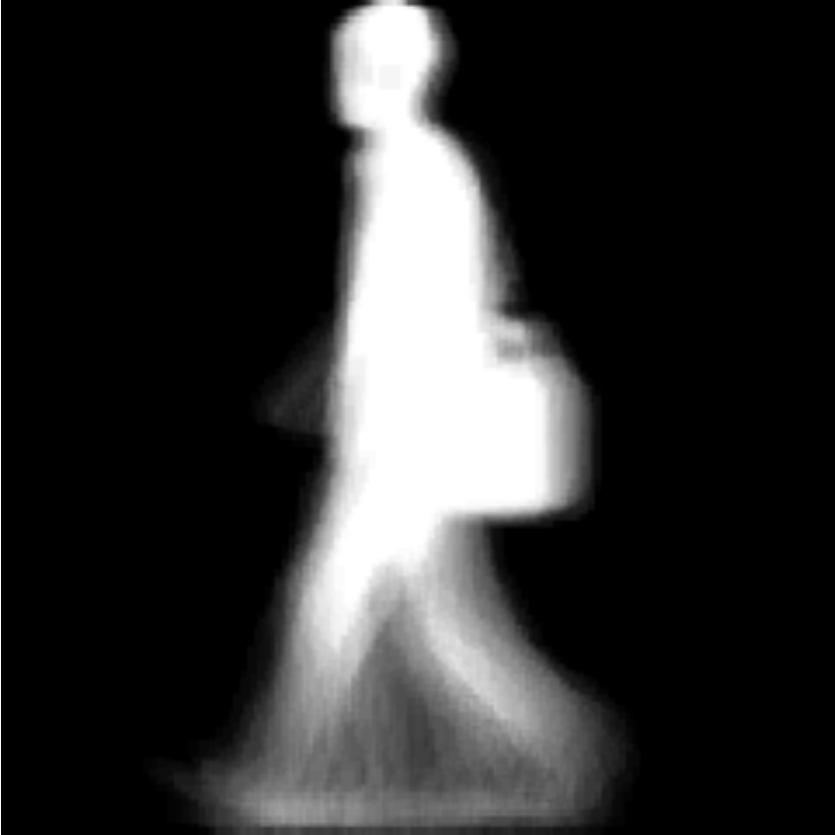}}
\subfigure[Wearing Coat ]{\label{fig:b}\includegraphics[width=34mm]{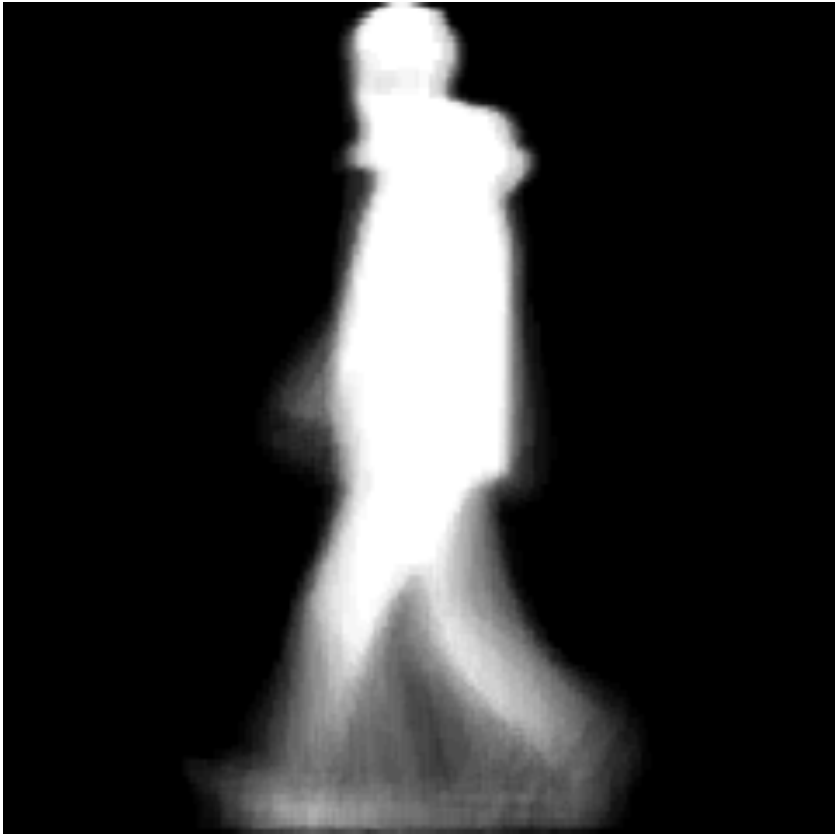}}
        \caption{Gait energy image of an individual under different conditions in side view.}
\label{fig:geiside}
\end{figure}

In this section, we focus on the effect of the carrying conditions and clothing variations so we carried out our experiments under $90^{\circ}$ view angle. This is motivated by the fact that side view is more affected by the clothing and carrying conditions than frontal view (see Figure \ref{fig:geifrontal} and  \ref{fig:geiside}). Furthermore gait information is more significant and reliable in the side view \citep{bashir2010}.
 
Table \ref {tab:ones} compares, Correct Correction Rate (CCR) under normal, carrying and clothing conditions, the mean and standard deviation of the performances under the three conditions of our proposed method, against the reported by other methods under $90^{\circ}$ view angle using similar experimental protocol. It shows that the CCR of our method is marginally lower in the normal and carrying conditions and significantly higher in the clothing variations than all other methods.

\begin{table}[!htbp] \centering
 \caption{Comparison of performances under different conditions (in percent), mean and standard deviation of the performances using $90^{\circ}$ view. Part-selection and without part-selection correspond to our method using the selected GEI part with group fused Lasso and whole GEI respectively. The best and second best results are highlighted by bold and star respectively. }
 \noindent
\resizebox{0.9\linewidth}{!}{
{\renewcommand{\arraystretch}{1.5}
\begin{tabularx}{14 cm}{ l c c c c l}
\hline
\hline
\textbf{Method}             & \centering \textbf{Normal}   & \centering \textbf{Carrying} & \centering \textbf{Clothing} & \centering  \textbf {Mean}  &  \textbf {Std}\\
\hline
\hline
GEI+TM \citep{yu2006}        & \centering 97.60 & \centering 32.70 & \centering 52.00 &  \centering 60.77 & 33.33 
\\
\hline
GEI+CDA \citep{han2006}  & \centering $99.60^{*}$ & \centering 57.20 & \centering 23.80 &  \centering 60.20 & 37.99\\
\hline
GEI+Filter+CDA \citep {bashir2008}  & \centering 99.40 & \centering 79.90 & \centering 31.30 & \centering 70.20 & 35.07 \\
\hline
GEnI+CDA \citep{bashir2010} & \centering \textbf{100.00}& \centering 78.30 & \centering 44.00 & \centering 74.10 &   28.24 \\
\hline
GEI+RF+CDA \citep {dupuis2013}   & \centering 98.80 & \centering 73.80 & \centering 63.70 &  \centering 78.77 &18.07 \\
\hline
GEI+Filter+CDA \citep {rida2015} & \centering 95.97 &\centering 63.39 & \centering  $72.77^{*}$ & \centering 77.38& $16.77^{*}$ \\
\hline
GEI+Wrapper+CDA \citep{rida2016}     & \centering 93.60 & \centering 81.70 & \centering 68.80 &  \centering $81.37^{*}$  &12.40 \\
\hline
\hline
Optical flow+CDA \citep{bashir2009}   & \centering 97.50 & \centering $83.60^{*}$ & \centering 48.80 & \centering76.63 &  25.09 \\
\hline
Optical flow+LBP+HMM \citep {hu2013}  & \centering 94.00 &\centering 45.20 & \centering  42.90 & \centering 60.70& 28.86  \\ 
\hline
GPPE+PCA+SVM \citep{jeevan2013} & \centering 93.36 &\centering 56.12 & \centering  22.44 & \centering  57.31 & 35.47 \\
\hline
STIPs+HOG/HOF+NN \citep {kusakunniran2014a} & \centering 95.40 &\centering 60.90 & \centering  52.00 & \centering 69.43& 22.92 \\
\hline
STIPs+HOG/HOF+SVM \citep {kusakunniran2014b} & \centering 94.50 &\centering 60.90 & \centering  58.50 & \centering  71.30 & 20.13  \\
\hline
EnDFT+PCA+NN \citep{rokanujjaman2015}  & \centering 97.61 &\centering \textbf{83.87} & \centering  51.61 & \centering  77.70 & 23.61\\
\hline
\hline
Proposed method without part-selection  & \centering \textbf{100.00} &\centering 55.80 & \centering {25.45} & \centering 60.42 & 37.49\\
\hline
Proposed method with part-selection    & \centering 98.39 &\centering 75.89 & \centering \textbf {91.96}& \centering \textbf {88.75} & \textbf{11.59}\\
\hline
\hline
\end{tabularx}}}
\label {tab:ones}
\end{table}
It is common in real life that people have different clothes depending on days (warm or cool days)  and seasons (summer or winter). Unfortunately, the intra-class variation of the static features (low motion) is mainly caused by the clothing variation that greatly affects the recognition accuracy adversely. It has been demonstrated by \citep{matovski2012} that clothing is the factor that drastically affects the performance of gait recognition. Thus, alleviating the problems caused by the clothing variation has significant meaning for gait recognition.  

The proposed method alleviates the clothing variation problem very well as it significantly outperforms all other approaches as shown in Table \ref {tab:ones}. In the normal and carrying conditions, different persons have different clothing conditions but all samples of a same person always have the same clothing condition in the dataset. Thus, the clothes in the normal and carrying conditions in fact undesirably contribute to differentiate persons. Therefore, these recognition rates could be misleading as they do not well reflect the real gait recognition performance. Note also that in the carrying conditions, some walking subjects carry handbags which influence the selected body-part leading to lower performances. 

Nevertheless, the proposed method performs the best among all approaches on the whole test dataset that contains one-third samples with cloth variation and two-third samples without the cloth variation and offers the best performance compromise between different conditions. This can be seen in the mean and standard deviation of our method which outperforms the mean and standard  deviation of the other methods.

\subsubsection{Effect of view-angle variations}

In this section we focus on the effect of the view-angle variations. In real life subjects are often captured under different view angles. To simulate these conditions we perform experiments in the so called "cross-view gait recognition". It corresponds to recognizing walking subjects where training and testing data are recorded from two different view angles.

Different view angle combinations (from $0^{\circ}$ to $180^{\circ}$) between training and testing data are used to estimate the recognition performances based on CDA. Tables \ref{tab:nl1} to \ref{tab:cl1} summarize the performances of the body-part cross-view gait recognition under normal, carrying conditions and clothing variations respectively when Tables \ref{tab:nl2} to  \ref{tab:cl2} show the same performances of whole-body (GEI without segmentation based group fused Lasso) under the same conditions.

The results demonstrate that our body-part method significantly outperforms the whole-body one under cloth variations however it has marginally lower performances in normal conditions due the undesirable contribution of clothing in recognition which was already pointed out previously. From the same results it can be seen that both the whole-body and body-part give good performances when the training view angle is similar to the testing one, however the performances significantly decrease when the difference between the training view angle and the testing one increases. This makes us conclude that there is an invert relationship between the view angle difference between training and testing data and the performance.

Based on the obtained results, we can clearly understand that conventional methods without pose estimation fail to give good recognition performances in case of the large intra-class variations caused by view angle variations between the training and testing data. Unfortunately, the latter is frequently encountered in real life gait recognition applications. This clearly show the mandatory to introduce new methods capable to address these issues.

Starting from the observation that the view-angle similarity between the training and testing data impacts performances, we introduce in the following section a novel method named "gait recognition without prior knowledge of the view angle" capable to reduce the intra-class variations. Our method is based on two main steps, the first one aims to estimate the view-angle of the testing samples when the second one compares them to training samples with similar view-angle. Based on this approach, the intra-class variations caused by view-angle variations are considerably reduced which leads to an improvement in the recognition performances. The method is described in next section. \\ \\

\begin{table}[htbp!]\centering
\caption{Cross-view body-part recognition under normal conditions(\%). Bold values correspond to CCR when training angle is similar to testing angle.}
\noindent
\resizebox{\linewidth}{!}{
\ra{1.3}
\setlength{\tabcolsep}{10pt}
\renewcommand{\arraystretch}{1.5}
\begin{tabular}{@{} ccccccccccccc}\toprule
& \multicolumn{4}{l}{ \hspace {7 mm} Testing angle normal conditions ($\degree$) } & \multicolumn{4}{c}{}
& \multicolumn{4}{r}{}\\
\cmidrule{3-13} 
\parbox[t]{6mm}{\multirow{13}{*}{\rotatebox[origin=c]{90}{Training angle normal conditions ($\degree$)  }}}
& $$ & $0$ & $18$ & $36$ & $54$ & $72$ & $90$ & $108$ & $126$ & $144$ & $162$ & $180$\\ \midrule

   & 0 & \textbf {98.37} & 5.24 & 1.61 & 1.21 & 0.40  & 0.81 & 0.81 & 1.61   & 0.81  & 0.81   & 9.27 \\

 & 18 & 6.10 & \textbf {98.79} & 17.74 & 1.61 & 0.81  & 0.81 & 1.21 & 1.61   & 4.44  & 2.42   & 2.82\\

  & 36 & 3.66 & 23.79 & \textbf {95.97} & 32.66 & 5.65 & 0.81 & 1.21 & 0.81  & 0.40  & 3.63   & 2.42\\

 & 54 & 2.03 & 5.24 & 33.87 & \textbf {96.77} & 11.69  & 4.84 & 1.61 & 1.21   & 0.40  & 1.61   & 2.02\\

 &72 & 1.22 & 2.02 & 3.23 & 10.08 & \textbf {98.39}  & 82.26 & 20.16 & 1.21   & 0.81  & 1.61   & 2.02\\

 &90& 1.22 & 1.21 & 2.82 & 7.66 & 67.74 & \textbf {98.39} & 48.79 & 4.84  &  3.23 & 1.61  & 1.21   \\

  &108 & 2.03 & 2.82 & 4.44 & 4.44 & 23.79 & 67.34 & \textbf {97.18} & 30.24  &  4.84   &  3.63 & 1.61   \\

   &126& 0.81 & 2.42 & 2.42 & 4.03 & 5.65 & 7.26 & 29.03 &  \textbf {95.56}  & 38.31   & 3.63   & 1.61 \\

    &144& 0.81 & 2.02 & 1.21 & 2.42 & 5.24 & 4.44 & 6.05 &  47.18  &  \textbf {97.18}   &  2.02  &  0.81 \\

     &162& 3.66 & 3.23 & 0.81 &  0.81 & 0.81 & 0.81 & 0.81 & 0.81  & 1.21   & \textbf {97.98}  & 6.85   \\

      &180& 10.57 & 2.42 & 1.61 & 0.40 & 0 & 0.40 & 0.81 & 1.61    & 2.42   & 3.63   &  \textbf {97.58}\\
\bottomrule
\end{tabular}}
\label{tab:nl1}
\end{table} 
\vspace{2cm}
\begin{table}[htbp!]\centering
\caption{Cross-view body-part recognition under carrying conditions (\%). Bold values correspond to CCR when training angle is similar to testing angle.}
\noindent
\resizebox{\linewidth}{!}{
\ra{1.3}
\setlength{\tabcolsep}{10pt}
\renewcommand{\arraystretch}{1.5}
\begin{tabular}{@{} ccccccccccccc}\toprule
& \multicolumn{4}{l}{ \hspace {7 mm} Testing angle carrying conditions ($\degree$) } & \multicolumn{4}{c}{}
& \multicolumn{4}{r}{}\\
\cmidrule{3-13} 
\parbox[t]{6mm}{\multirow{13}{*}{\rotatebox[origin=c]{90}{Training angle normal conditions ($\degree$)  }}}
& $$ & $0$ & $18$ & $36$ & $54$ & $72$ & $90$ & $108$ & $126$ & $144$ & $162$ & $180$\\ \midrule

   & 0 & \textbf {72.36} & 2.02 & 0.81 & 0.81 & 0.40  & 0 & 0.40 & 2.02   & 1.62  & 2.04   & 8.50 \\

 & 18 & 5.28 & \textbf {73.79} & 9.68 & 2.03 & 2.02  & 1.79 & 1.61 & 2.02   & 1.62  & 3.67   & 2.02\\

  & 36 & 4.07 & 16.94 & \textbf {77.02} & 27.64 & 4.44 & 1.34 & 2.02 & 0.81  & 0  & 5.31   & 1.62\\

 & 54 & 1.63 & 6.45 & 25.40 & \textbf {75.61} & 10.48  & 3.57 & 1.21 & 1.21   & 0.81  & 2.04   & 2.02\\

 &72 & 1.63 & 1.61 & 1.61 & 10.16 & \textbf {75.00}  & 56.70 & 15.32 & 2.02   & 0.81  & 2.04   & 2.83\\

 &90& 0.81 & 1.61 & 2.42 & 5.69 & 45.16 & \textbf {75.89} & 25.00 & 4.86  &  2.43 & 0.82  & 1.21   \\

  &108 & 0.81 & 0.81 & 4.03 & 3.66 & 14.92 & 53.57 & \textbf {75.00} & 22.27  &  6.88   &  3.27 & 2.43   \\

   &126& 1.22 & 1.21 & 2.42 & 2.44 & 6.85 & 6.25 & 29.84 &  \textbf {76.52}  & 28.34   & 2.04   & 1.21 \\

    &144& 1.22 & 0.81 & 1.61 & 2.03 & 4.84 & 4.46 & 5.24 &  33.60  &  \textbf {77.33}   &  0  &  0.81 \\

     &162& 2.85 & 1.21 & 1.21 &  1.22 & 1.21 & 1.34 & 0.81 & 0.81  & 0.40   & \textbf {74.69}  & 3.24   \\

      &180& 9.76 & 2.42 & 0.81 & 0.81 & 0.40 & 0.89 & 0.81 & 2.02    & 1.62   & 4.08   &  \textbf {75.71}\\
\bottomrule
\end{tabular}}
\label{tab:bg1}
\end{table}

\begin{table}[htbp!]\centering
\caption{Cross-view body-part recognition under clothing variations (\%). Bold values correspond to CCR when training angle is similar to testing angle.}
\noindent
\resizebox{\linewidth}{!}{
\ra{1.3}
\setlength{\tabcolsep}{10pt}
\renewcommand{\arraystretch}{1.5}
\begin{tabular}{@{} ccccccccccccc}\toprule
& \multicolumn{4}{l}{ \hspace {7 mm} Testing angle clothing conditions ($\degree$) } & \multicolumn{4}{c}{}
& \multicolumn{4}{r}{}\\
\cmidrule{3-13} 
\parbox[t]{6mm}{\multirow{13}{*}{\rotatebox[origin=c]{90}{Training angle normal conditions ($\degree$)  }}}
& $$ & $0$ & $18$ & $36$ & $54$ & $72$ & $90$ & $108$ & $126$ & $144$ & $162$ & $180$\\ \midrule

   & 0 & \textbf {80.89} & 4.03 & 2.42 & 1.62 & 0.81  & 0.89 & 0.81 & 2.43   & 2.02  & 0.82   & 9.27 \\

 & 18 & 5.28 & \textbf {83.06} & 12.90 & 2.02 & 0.81  & 0.89 & 0.81 & 1.62   & 2.83  & 2.04   & 3.23\\

  & 36 & 2.44 & 19.35 & \textbf {85.08} & 29.55 & 6.85 & 2.68 & 1.61 & 1.62  & 0.40  & 2.45   & 1.21\\

 & 54 & 1.63 & 5.65 & 30.24 & \textbf {87.04} & 10.08  & 4.02 & 1.21 & 0.81   & 0  & 0.82   & 0.81\\

 &72 & 1.22 & 1.61 & 2.42 & 12.96 & \textbf {91.13}  & 62.95 & 18.55 & 0.40   & 0  & 0.82   & 0.81\\

 &90& 0.41 & 1.61 & 3.23 & 6.07 & 60.48 & \textbf {91.96} & 40.32 & 4.05  &  2.43 & 1.63  & 1.61   \\

  &108 & 1.63 & 3.23 & 1.61 & 3.64 & 18.95& 56.25 & \textbf {88.71} & 31.58  &  4.45   &  3.67 & 1.61   \\

   &126& 1.22 & 1.61 & 1.61 & 4.05 & 4.44 & 4.91 & 22.18 &  \textbf {87.04}  & 40.08   & 3.67   & 1.61 \\

    &144& 2.03 & 1.21 & 1.61 & 2.02 & 5.65 & 1.79 & 4.03 &  27.13  &  \textbf {90.28}   &  2.86  &  1.61 \\

     &162& 3.25 & 2.82 & 2.02 &  1.62 & 1.21 & 1.34 & 1.21 & 1.62  & 1.21   & \textbf {86.94}  & 6.85   \\

      &180& 9.35 & 2.02 & 2.02 & 0.81 & 0.81 & 0.89 & 0.81 & 1.62    & 0.81   & 2.86   &  \textbf {84.27}\\
\bottomrule
\end{tabular}}
\label{tab:cl1}
\end{table}


\begin{table}[htbp!]\centering
\caption{Cross-view whole-body recognition normal (\%). Bold values correspond to CCR when training angle is similar to testing angle.}
\noindent
\resizebox{\linewidth}{!}{
\ra{1.3}
\setlength{\tabcolsep}{10pt}
\renewcommand{\arraystretch}{1.5}
\begin{tabular}{@{} ccccccccccccc}\toprule
& \multicolumn{4}{l}{ \hspace {7 mm} Testing angle normal conditions ($\degree$) } & \multicolumn{4}{c}{}
& \multicolumn{4}{r}{}\\
\cmidrule{3-13} 
\parbox[t]{6mm}{\multirow{13}{*}{\rotatebox[origin=c]{90}{Training angle normal conditions ($\degree$)  }}}
& $$ & $0$ & $18$ & $36$ & $54$ & $72$ & $90$ & $108$ & $126$ & $144$ & $162$ & $180$\\ \midrule

   & 0 & \textbf {100} & 70.16 & 14.92 & 5.24 & 2.42  & 2.02 & 0.81 & 0.81   & 4.44  & 15.32   & 40.32 \\

 & 18 & 82.11& \textbf {100} & 92.74 & 16.13 & 3.63  & 1.21 & 2.42 & 4.84   & 15.32  & 21.77   & 31.85\\

  & 36 & 38.21 & 94.76 & \textbf {99.19} & 85.89 & 30.24 & 15.73 & 12.50 & 22.58  & 20.97  & 21.77   & 9.27\\

 & 54 & 9.76 & 27.82 & 92.34 & \textbf {99.19} & 70.97  & 35.48 & 21.77 & 27.42   & 23.79  & 6.05   & 6.45\\

 &72 & 6.10 & 4.03 & 16.13 & 63.31 & \textbf {99.19}  & 98.79 & 74.19 & 14.92   & 4.84  & 5.24   & 4.44\\

 &90& 2.03 & 2.02 & 6.45 & 17.34 & 98.79 & \textbf {100} & 97.18 & 22.98  &  6.05 & 2.82  & 2.42   \\

  &108 & 2.44 & 0.81 & 8.06 & 33.06 & 79.84 & 97.98 & \textbf {99.60} & 91.53  &  22.58   &  3.63 & 2.42   \\

   &126& 6.50 & 4.84 & 12.10 & 31.45 & 47.58 & 50.81 & 90.73 &  \textbf {98.39}  & 94.76   & 15.32   & 6.45 \\

    &144& 13.01 & 15.73 & 27.02 & 19.35 & 8.87 & 6.45 & 31.45 &  95.16  &  \textbf {99.19}   &  34.68  &  11.29 \\

     &162& 20.73 & 25.00 & 15.32 &  6.05 & 0.81 & 0.81 & 1.21 & 2.42  & 6.05   & \textbf {99.60}  & 70.56   \\

      &180& 52.44 & 18.55 & 12.10 & 4.84 & 3.23 & 1.61 & 0.81 & 2.42    & 9.27   & 77.42   &  \textbf {100}\\
\bottomrule
\end{tabular}}
\label{tab:nl2}
\end{table}

\begin{table}[htbp!]\centering
\caption{Cross-view whole-body recognition carrying conditions (\%). Bold values correspond to CCR when training angle is similar to testing angle.}
\noindent
\resizebox{\linewidth}{!}{
\ra{1.3}
\setlength{\tabcolsep}{10pt}
\renewcommand{\arraystretch}{1.5}
\begin{tabular}{@{} ccccccccccccc}\toprule
& \multicolumn{4}{l}{ \hspace {7 mm} Testing angle carrying conditions ($\degree$) } & \multicolumn{4}{c}{}
& \multicolumn{4}{r}{}\\
\cmidrule{3-13} 
\parbox[t]{6mm}{\multirow{13}{*}{\rotatebox[origin=c]{90}{Training angle normal conditions ($\degree$)  }}}
& $$ & $0$ & $18$ & $36$ & $54$ & $72$ & $90$ & $108$ & $126$ & $144$ & $162$ & $180$\\ \midrule

   & 0 & \textbf {83.74} & 45.56 & 14.92 & 6.50 & 4.44  & 2.23 & 1.61 & 2.02   & 2.83  & 6.53   & 21.46 \\

 & 18 & 54.07 & \textbf {79.44} & 54.03 & 11.79 & 4.44  & 0.45 & 1.21 & 4.45   & 5.67  & 10.20   & 10.53\\

  & 36 & 27.64 & 55.24 & \textbf {74.60} & 46.34 & 16.13 & 6.70 & 3.63 & 7.69  & 6.48  & 8.98   & 5.26\\

 & 54 & 4.88 & 14.52 & 48.79 & \textbf {69.11} & 37.90  & 23.21 & 10.08 & 11.74   & 9.31  & 8.98   & 5.67\\

 &72 & 5.69 & 4.44 & 7.66 & 24.80 & \textbf {59.68}  & 47.77 & 23.79 & 8.91   & 4.86  & 3.67   & 5.26\\

 &90& 2.03 & 2.42 & 3.63 & 11.79 & 47.98 & \textbf {55.80} & 39.92 & 9.72  &  4.05 & 2.86  & 2.43   \\

  &108 & 2.44 & 0.81 & 4.44 & 15.45 & 40.73 & 50.89 & \textbf {59.27} & 35.22  & 12.55   &  4.08 & 2.83   \\

   &126& 4.07 & 3.23 & 9.68 & 20.73 & 27.02 & 28.57 & 38.31 &  \textbf {62.35}  & 43.32   & 8.57   & 4.45 \\

    &144& 5.69 & 8.87 & 15.32 & 11.38 & 5.24 & 5.36 & 8.47 &  48.58  &  \textbf {70.45}   & 17.96  &  8.10 \\

     &162& 10.98 & 13.71 & 5.24 &  2.44 & 1.61 & 1.79 & 1.61 & 2.43  & 4.05   & \textbf {67.35}  & 31.17   \\

      &180& 29.27 & 13.71 & 6.05 & 3.66 & 2.42 & 0.45 & 2.02 & 2.02    & 6.48   & 34.29   &  \textbf {76.11}\\
\bottomrule
\end{tabular}}
\label{tab:bg2}
\end{table}

\begin{table}[htbp!]\centering
\caption{Cross-view body-part recognition clothing variations (\%). Bold values correspond to CCR when training angle is similar to testing angle.}
\noindent
\resizebox{\linewidth}{!}{
\ra{1.3}
\setlength{\tabcolsep}{10pt}
\renewcommand{\arraystretch}{1.5}
\begin{tabular}{@{} ccccccccccccc}\toprule
& \multicolumn{4}{l}{ \hspace {7 mm} Testing angle clothing conditions ($\degree$) } & \multicolumn{4}{c}{}
& \multicolumn{4}{r}{}\\
\cmidrule{3-13} 
\parbox[t]{6mm}{\multirow{13}{*}{\rotatebox[origin=c]{90}{Training angle normal conditions ($\degree$)  }}}
& $$ & $0$ & $18$ & $36$ & $54$ & $72$ & $90$ & $108$ & $126$ & $144$ & $162$ & $180$\\ \midrule

   & 0 & \textbf {28.05} & 14.52 & 5.65 & 2.02 & 1.21  & 0.45 & 1.21 & 1.62   & 3.64  & 6.94   & 7.66 \\

 & 18 & 11.38 & \textbf {25.81} & 21.37 & 6.48 & 4.03  & 3.57 & 2.82 & 4.05   & 6.07  & 6.94   & 5.65\\

  & 36 & 8.94 & 18.95 & \textbf {31.05} & 23.48 & 8.87 & 6.70 & 4.44 & 6.88  & 5.26  & 7.76   & 2.42\\

 & 54 & 1.22 & 7.66 & 20.97 & \textbf {28.34} & 16.53  & 7.59 & 6.85 & 6.88   & 4.45  & 2.45   & 0.40\\

 &72 & 0.81 & 1.61 & 2.42 & 9.31 & \textbf {29.44}  & 22.32 & 12.50 & 4.86   & 1.62  & 1.63   & 2.02\\

 &90& 2.85 & 1.61 & 2.02 & 7.29 & 16.53 & \textbf {25.45} & 14.92 & 5.67  &  1.62 & 2.04  & 0   \\

  &108 & 0.81 & 1.61 & 3.23 & 5.26 & 13.71 & 17.86 & \textbf {24.60} & 12.96  &  5.26   &  1.63 & 0.40   \\

   &126& 1.22 & 2.02 & 3.23 & 5.26 & 10.48 & 11.61 & 23.39 &  \textbf {31.58}  & 19.43   & 1.22   & 1.21 \\

    &144& 5.28 & 5.65 & 7.26 & 8.50 & 6.45 & 3.13 & 6.05 &  25.91  &  \textbf {37.25}   &  4.08  &  3.23 \\

     &162& 5.28 & 6.45 & 7.26 &  5.67 & 1.21 & 1.34 & 0.81 & 2.02  & 4.45   & \textbf {31.02}  & 12.10   \\

      &180& 10.16 & 7.66 & 5.24 & 1.21 & 1.61 & 1.79 & 2.02 & 2.83    & 4.45   & 12.24   &  \textbf {30.65}\\
\bottomrule
\end{tabular}}
\label{tab:cl2}
\end{table}

 \subsubsection {Gait recognition without prior knowledge of the view angle} 

\begin{figure}[!htbp]
\centering
\includegraphics [width= 12 cm] {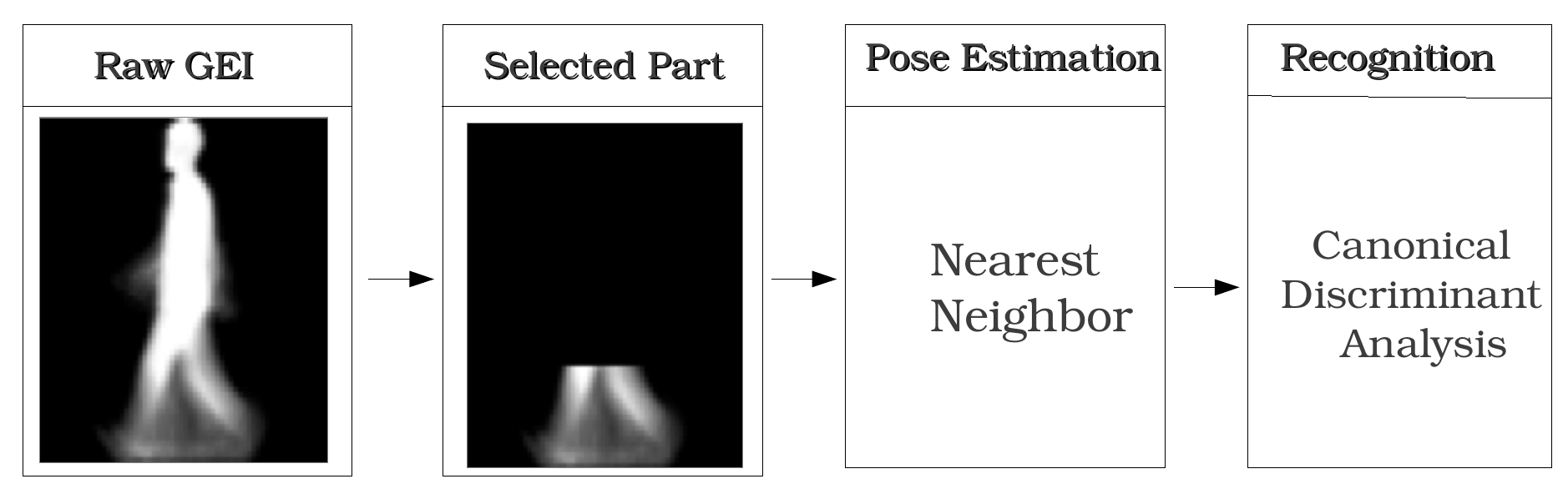}
\caption{Framework of view angle variation without prior knowledge of the view angle.}
\label{fig:pose}
\end {figure}

The framework in Figure \ref{fig:pose} is designed to recognize individuals without a prior knowledge of the viewpoint. Towards this end, the first step consists on estimating the pose of the query test sample using the selected human body part .i.e. row $46$ to $64$ (it has been explained above how the body part is selected using the group fused Lasso of motion) and nearest-neighbor classifier to find the group of training samples which have the pose similar to that of the query subject. The next step consists on identifying the query subject among the group of training samples with the same pose using Canonical Discriminant Analysis (CDA).

The results of pose estimation are shown in Table \ref{tab:pose}, it can be seen that the  selected body-part is very discriminative and we are able to estimate the pose of the query subjects of the test dataset with an error less than 3 \% for all view angles from $0^{\circ}$ to $180^{\circ}$. 

\begin{figure}[!h]
\centering
\subfigure[Normal conditions and angle variations ] {\label{fig:aa}\includegraphics[width=71mm]{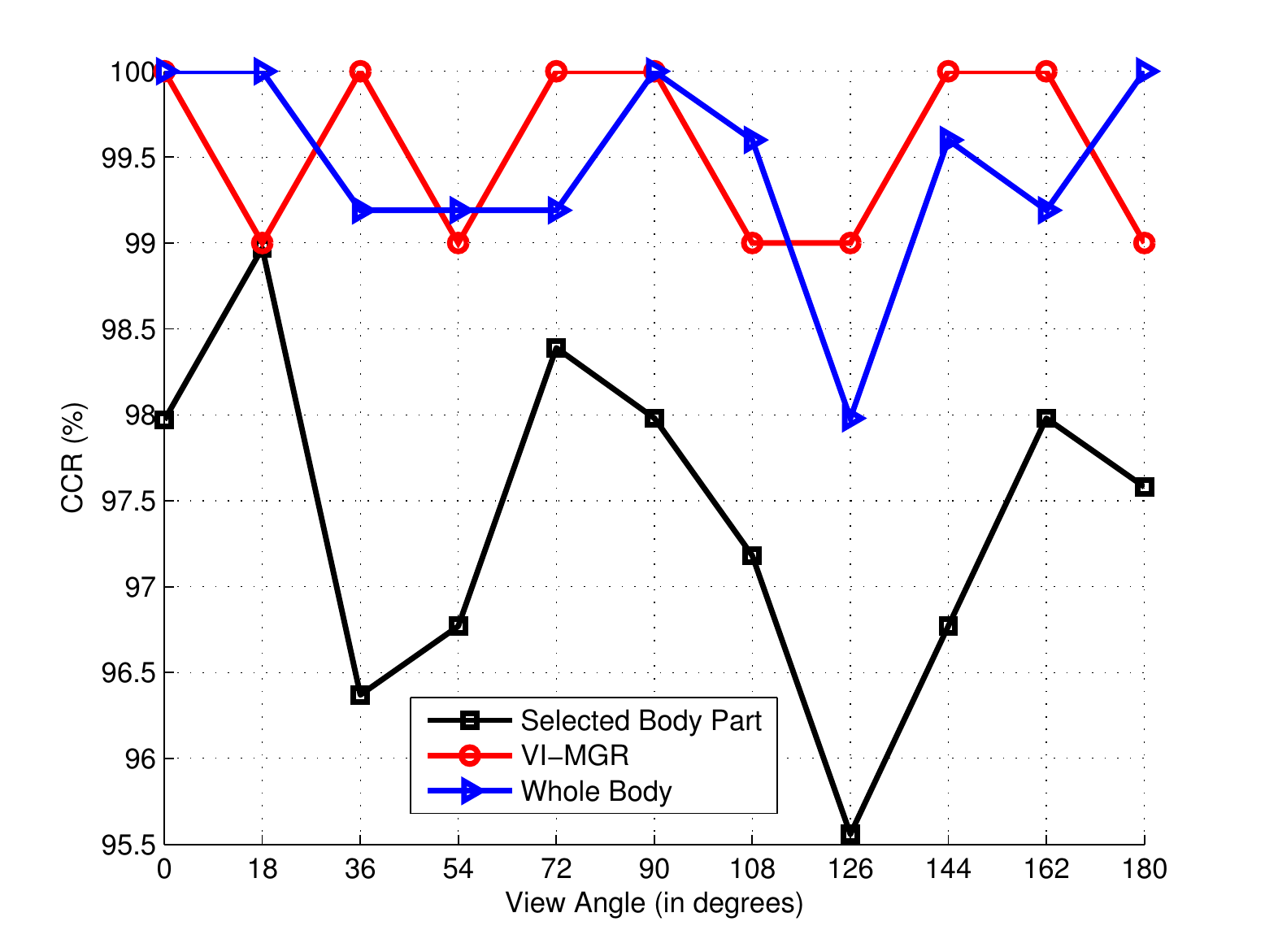}}
\subfigure[Carrying conditions and angle variations]{\label{fig:bb}\includegraphics[width=71mm]{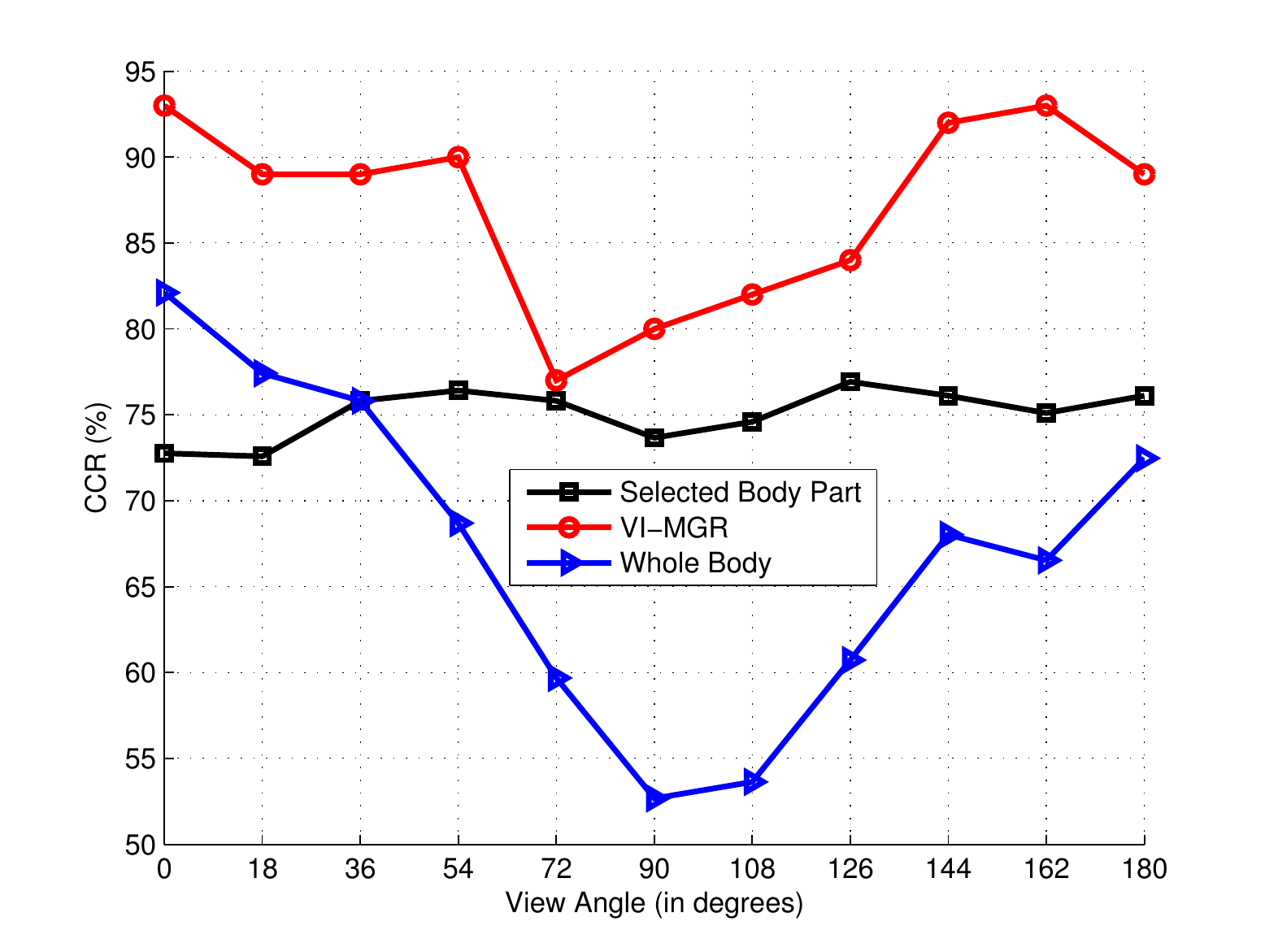}}\\
\subfigure[Cloth and angle variations]{\label{fig:dd}\includegraphics[width=71mm]{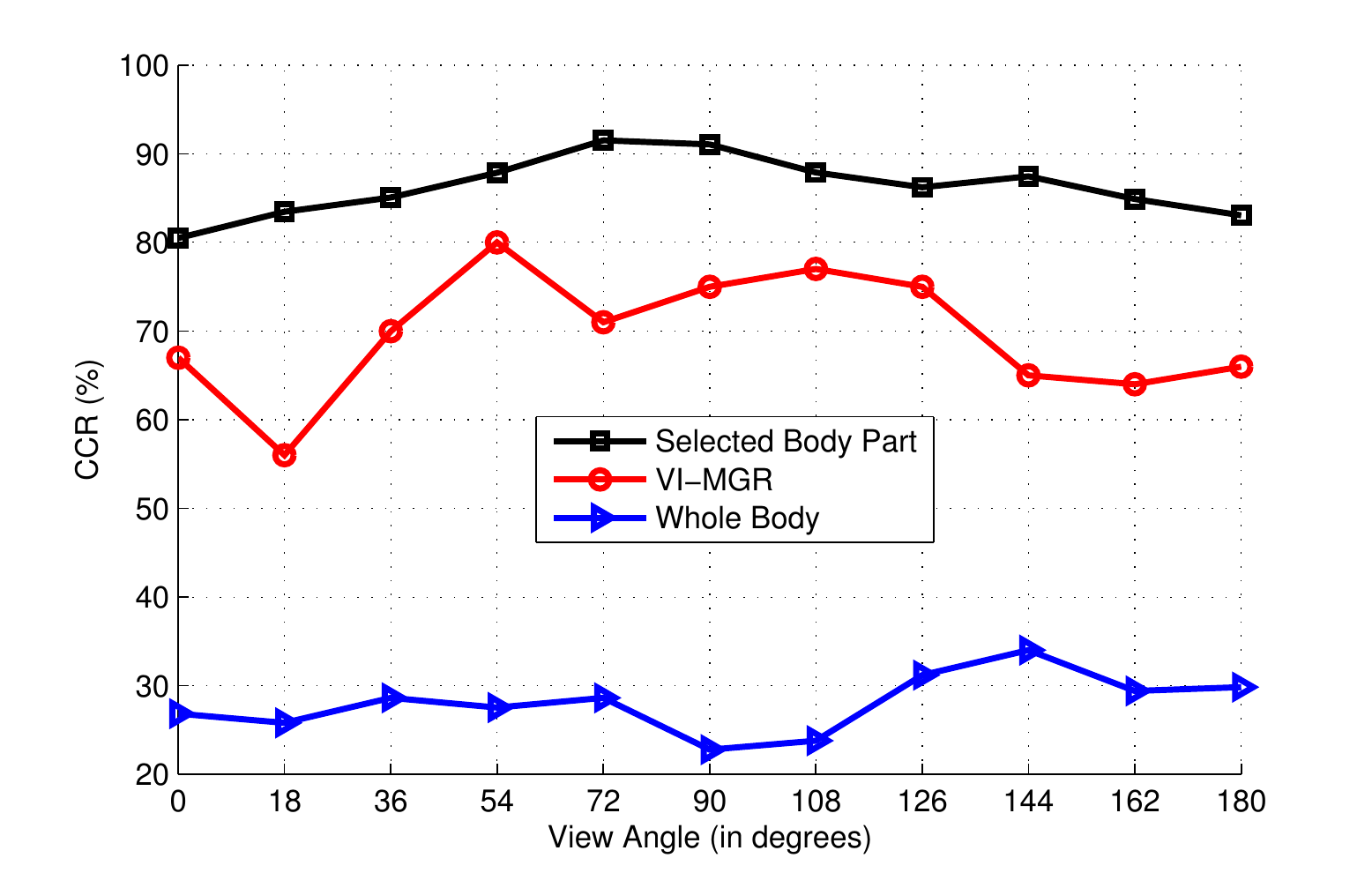}}
\subfigure[Mean under different conditions]{\label{fig:cc}\includegraphics[width=71mm]{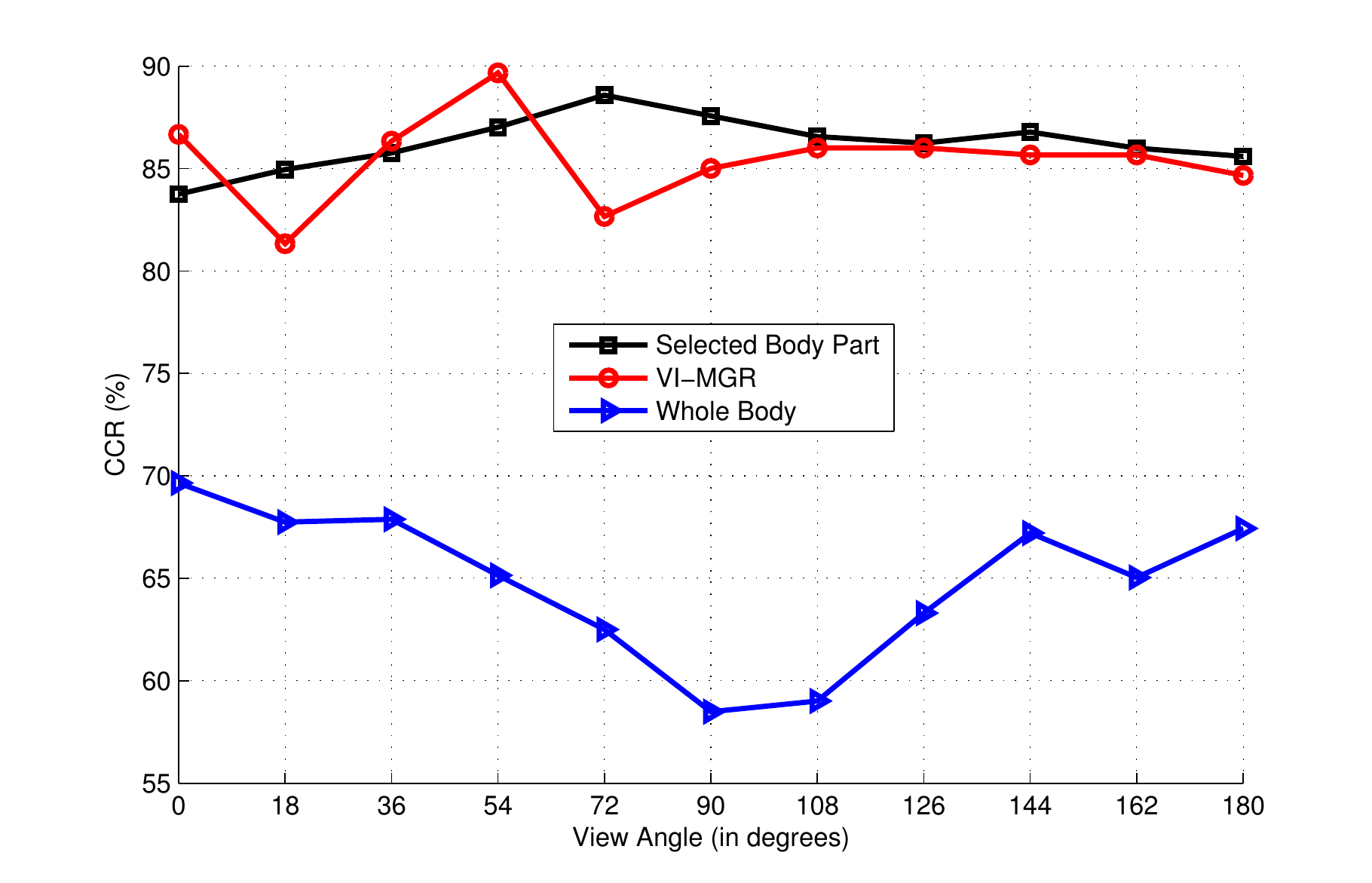}}
        \caption{Comparison of CCR under different conditions for body-part, whole-body and VI-MGR.}
\label{estimatedangleall}
\end{figure}

Figure \ref{estimatedangleall} shows the CCR under different conditions of our proposed body-part approach, the approach that uses the whole-body (without body segmentation) and the View-Invariant Multiscale Gait Recognition method (VI-MGR) \citep{choudhury2015} representing the most recent introduced method to deal with the problem view-angle variations based on the idea of estimating the pose. Results clearly show that our proposed body-part method significantly outperforms VI-MGR and the approach without the part selection for all 11 view angle variations in the case of the clothing variation (see Figure \ref{fig:dd}). On the whole test dataset that contains one-third samples with cloth variation and two-third samples without the cloth variation, the proposed approach outperforms the whole-body approach for all view angle variations and  outperforms VI-MGR  in 8 of the 11 view angle variations (see Figure \ref{fig:cc}). 

The previously encountered problems of the CCR for normal and carrying conditions are shown in Figure \ref{fig:aa} and Figure \ref{fig:bb}. Our approach takes in consideration only the dynamic part, when other approaches take both static and dynamic parts. The latter could be very discriminative and complementary to the dynamic information mostly when subjects keep the same clothes which is the case in normal condition experiments. In addition of this, in the carrying conditions, our selected body-part could be affected when the walking subjects carry handbag instead of backpack which influences the recognition performances.

\begin{table}[h!]\centering
\caption{Pose estimation-confusion matrix (\%). Bold values correspond to well-predicted angles.}
\noindent
\resizebox{\linewidth}{!}{
\ra{1.3}
\setlength{\tabcolsep}{10pt}
\renewcommand{\arraystretch}{1.5}
\begin{tabular}{@{} ccccccccccccc}\toprule
& \multicolumn{4}{l}{ \hspace {8 mm} Predicted angle ($\degree$) } & \multicolumn{4}{c}{}
& \multicolumn{4}{r}{}\\
\cmidrule{3-13}  
\parbox[t]{6mm}{\multirow{13}{*}{\rotatebox[origin=c]{90}{Real angle ($\degree$)  }}}
& $$ & $0$ & $18$ & $36$ & $54$ & $72$ & $90$ & $108$ & $126$ & $144$ & $162$ & $180$\\ \midrule

   & 0 & \textbf {98.78} & 0.27 & 0 & 0 & 0 & 0 & 0 & 0   & 0.40  & 0   & 0.54 \\

 & 18 & 0.40 & \textbf {97.58} & 1.34 & 0 & 0  & 0.13 & 0 & 0.13   & 0.26 & 0   & 0.13\\

  & 36 & 0.26 & 1.20 & \textbf {97.31} & 0.80 & 0 & 0 & 0 & 0  & 0.40  & 0   & 0\\

 & 54 & 0.13 & 0.13 & 0.8 & \textbf {98.65} & 0  & 0 & 0.13 & 0   & 0.13  & 0   & 0 \\

 &72 & 0 & 0.26 & 0.13 & 0 & \textbf {98.92}  & 0.13 & 0.40 & 0.13  & 0  & 0   & 0\\

 &90& 0 & 0.14 & 0 & 0.43 & 0.43 & \textbf {98.41} & 0.57 & 0  &  0 & 0 & 0   \\

  &108 & 0 & 0  & 0 & 0.13 & 0 & 1.34 & \textbf {97.71} & 0.53  &  0   &  0.26 & 0   \\

   &126& 0 & 0  & 0 & 0.13 & 0 & 0 & 0.40 &  \textbf {98.92}  & 0  & 0.26  & 0.26 \\

    &144& 0 & 0.13 & 0.13 & 0 & 0 & 0 & 0.13 &  0.26  &  \textbf {97.57}   &  1.48  &  0.26 \\

     &162& 0 & 0.27 & 0.13 &  0.13 & 0 & 0 & 0 & 0  & 1.62   & \textbf {97.83}  & 0   \\

      &180& 1.07  & 0.26 & 0 & 0 & 0 & 0 & 0 & 0.13    & 0   & 0   &  \textbf {98.51} \\
\bottomrule
\end{tabular}}
\label{tab:pose}
\end{table}

\section{Conclusion}

We have proposed a method that finds the discriminative human body-part that is also robust to the intra-class variations for improving the human gait recognition. The proposed method first generates a horizontal motion based vector from GEI and then applies the group fused Lasso on the horizontal motion based vectors of a feature selection dataset to automatically learn the discriminative human body-parts for gait recognition. The learned human body part is applied to the independent training and test datasets. The proposed method significantly improves the recognition accuracy in the case of large intra-class variation
such as the clothing variation. This is verified by the experiments, which show that the proposed methods not only significant outperforms other approaches in the case of clothing variations but also achieves the overall best performance among all approaches on the whole testing dataset that contains
normal, carrying, clothing and view angle variations.

The method was further improved to deal with the problem of intra-class variations caused by the view-angle variations between training and testing gait sequences based on a pose estimation technique able to compare the training and testing samples with similar pose. 

Some extensions to our approach for gait recognition are envisioned. For instance, a gain in performances can be expected by relying on more elaborate classification methods. Two aspects can be considered: learning of an adequate metric \citep{bellet2013survey} or investigating classifiers as SVM \cite{rida2014supervised}. Issues related to view-angle variations are reminiscent to domain adaptation \citep{gopalan2011domain, kulis2011you, sun2015return,rida2018efficient} where the statistics of testing samples differ from those of the training data used to learn the recognition system.  Indeed, because of the different acquisition angles the recorded gait images of a person lean on a manifold in an ambient high dimension-space inducing hence geometrical transformations of training and testing sets. Moreover the changing conditions (normal, clothing, carrying) affect more heavily the statistics of both sets. In that context, as an interesting perspective we plan to lift our body part-selection approach in domain adaptation techniques. Particularly, we intend to explore novel method such as optimal transport for domain adaptation based on a manifold regularization inspiring from the work in \citep{courty2016optimal}.


\begin{thebibliography}{10}

\bibitem{fei2017enhanced}
Lunke Fei, Shaohua Teng, Jigang Wu, and Imad Rida.
\newblock Enhanced minutiae extraction for high-resolution palmprint
  recognition.
\newblock {\em International Journal of Image and Graphics}, 17(04):1750020,
  2017.

\bibitem{rida2018ensemble}
Imad Rida, Somaya Al~Maadeed, Xudong Jiang, Fei Lunke, and Abdelaziz Bensrhair.
\newblock An ensemble learning method based on random subspace sampling for
  palmprint identification.
\newblock In {\em 2018 IEEE International conference on acoustics, speech and
  signal processing (ICASSP)}, pages 2047--2051. IEEE, 2018.

\bibitem{rida2018novel}
Imad Rida, Noor Al~Maadeed, and Somaya Al~Maadeed.
\newblock A novel efficient classwise sparse and collaborative representation
  for holistic palmprint recognition.
\newblock In {\em 2018 NASA/ESA Conference on Adaptive Hardware and Systems
  (AHS)}, pages 156--161. IEEE, 2018.

\bibitem{al2018palmprint}
Somaya Al~Maadeed, Xudong Jiang, Imad Rida, and Ahmed Bouridane.
\newblock Palmprint identification using sparse and dense hybrid
  representation.
\newblock {\em Multimedia Tools and Applications}, pages 1--15, 2018.

\bibitem{rida2018feature}
Imad Rida.
\newblock Feature extraction for temporal signal recognition: An overview.
\newblock {\em arXiv preprint arXiv:1812.01780}, 2018.

\bibitem{rida2018palmprint}
Imad Rida, Somaya Al-Maadeed, Arif Mahmood, Ahmed Bouridane, and Sambit Bakshi.
\newblock Palmprint identification using an ensemble of sparse representations.
\newblock {\em IEEE Access}, 6:3241--3248, 2018.

\bibitem{shariatmadari2018off}
Sima Shariatmadari, Somaya Al-maadeed, Younes Akbari, Imad Rida, and Sima
  Emadi.
\newblock Off-line persian signature verification using wavelet-based fractal
  dimension and one-class gaussian process.
\newblock In {\em 2018 NASA/ESA Conference on Adaptive Hardware and Systems
  (AHS)}, pages 168--173. IEEE, 2018.

\bibitem{micheletto2018multiple}
Marco Micheletto, Giulia Orr{\`u}, Imad Rida, Luca Ghiani, and Gian~Luca
  Marcialis.
\newblock A multiple classifiers-based approach to palmvein identification.
\newblock In {\em 2018 Eighth International Conference on Image Processing
  Theory, Tools and Applications (IPTA)}, pages 1--6. IEEE, 2018.

\bibitem{rida2019forensic}
Imad Rida, Sambit Bakshi, Xiaojun Chang, and Hugo Proenca.
\newblock Forensic shoe-print identification: A brief survey.
\newblock {\em arXiv preprint arXiv:1901.01431}, 2019.

\bibitem{boyd2005}
Jeffrey~E Boyd and James~J Little.
\newblock Biometric gait recognition.
\newblock In {\em Advanced Studies in Biometrics}, pages 19--42. Springer,
  2005.

\bibitem{johansson1973}
Gunnar Johansson.
\newblock Visual perception of biological motion and a model for its analysis.
\newblock {\em Perception \& psychophysics}, 14(2):201--211, 1973.

\bibitem{johansson1975}
Gunnar Johansson.
\newblock Visual motion perception.
\newblock {\em Scientific American}, 1975.

\bibitem{matovski2012}
Darko~S Matovski, Mark~S Nixon, Sasan Mahmoodi, and John~N Carter.
\newblock The effect of time on gait recognition performance.
\newblock {\em Information Forensics and Security, IEEE Transactions on},
  7(2):543--552, 2012.

\bibitem{yu2006}
Shiqi Yu, Daoliang Tan, and Tieniu Tan.
\newblock A framework for evaluating the effect of view angle, clothing and
  carrying condition on gait recognition.
\newblock In {\em Pattern Recognition, 2006. ICPR 2006. 18th International
  Conference on}, volume~4, pages 441--444. IEEE, 2006.

\bibitem{han2006}
Jinguang Han and Bir Bhanu.
\newblock Individual recognition using gait energy image.
\newblock {\em Pattern Analysis and Machine Intelligence, IEEE Transactions
  on}, 28(2):316--322, 2006.

\bibitem{sarkar2005}
Sudeep Sarkar, P~Jonathon Phillips, Zongyi Liu, Isidro~Robledo Vega, Patrick
  Grother, and Kevin~W Bowyer.
\newblock The humanid gait challenge problem: Data sets, performance, and
  analysis.
\newblock {\em Pattern Analysis and Machine Intelligence, IEEE Transactions
  on}, 27(2):162--177, 2005.

\bibitem{bashir2010}
Khalid Bashir, Tao Xiang, and Shaogang Gong.
\newblock Gait recognition without subject cooperation.
\newblock {\em Pattern Recognition Letters}, 31(13):2052--2060, 2010.

\bibitem{dupuis2013}
Yohan Dupuis, Xavier Savatier, and Pascal Vasseur.
\newblock Feature subset selection applied to model-free gait recognition.
\newblock {\em Image and vision computing}, 31(8):580--591, 2013.

\bibitem{cunado2003}
David Cunado, Mark~S Nixon, and John~N Carter.
\newblock Automatic extraction and description of human gait models for
  recognition purposes.
\newblock {\em Computer Vision and Image Understanding}, 90(1):1--41, 2003.

\bibitem{murray1964}
M~Pat Murray, A~Bernard Drought, and Ross~C Kory.
\newblock Walking patterns of normal men.
\newblock {\em J Bone Joint Surg Am}, 46(2):335--360, 1964.

\bibitem{murray1967}
M~Pat Murray.
\newblock Gait as a total pattern of movement: Including a bibliography on
  gait.
\newblock {\em American Journal of Physical Medicine \& Rehabilitation},
  46(1):290--333, 1967.

\bibitem{rida2018robust}
Imad Rida, Noor Almaadeed, and Somaya Almaadeed.
\newblock Robust gait recognition: a comprehensive survey.
\newblock {\em IET Biometrics}, 8(1):14--28, 2018.

\bibitem{rida2015unsupervised}
Imad Rida, Somaya Al~Maadeed, and Ahmed Bouridane.
\newblock Unsupervised feature selection method for improved human gait
  recognition.
\newblock In {\em 2015 23rd European Signal Processing Conference (EUSIPCO)},
  pages 1128--1132. IEEE, 2015.

\bibitem{benabdelkader2002}
Chiraz BenAbdelkader, Ross Cutler, and Larry Davis.
\newblock Stride and cadence as a biometric in automatic person identification
  and verification.
\newblock In {\em Automatic Face and Gesture Recognition, 2002. Proceedings.
  Fifth IEEE International Conference on}, pages 372--377. IEEE, 2002.

\bibitem{nixon2009model}
Mark Nixon et~al.
\newblock Model-based gait recognition.
\newblock 2009.

\bibitem{wagg2004}
David~K Wagg and Mark~S Nixon.
\newblock On automated model-based extraction and analysis of gait.
\newblock In {\em Automatic Face and Gesture Recognition, 2004. Proceedings.
  Sixth IEEE International Conference on}, pages 11--16. IEEE, 2004.

\bibitem{wang2004}
Liang Wang, Huazhong Ning, Tieniu Tan, and Weiming Hu.
\newblock Fusion of static and dynamic body biometrics for gait recognition.
\newblock {\em Circuits and Systems for Video Technology, IEEE Transactions
  on}, 14(2):149--158, 2004.

\bibitem{bobick2001}
Aaron~E Bobick and Amos~Y Johnson.
\newblock Gait recognition using static, activity-specific parameters.
\newblock In {\em Computer Vision and Pattern Recognition, 2001. CVPR 2001.
  Proceedings of the 2001 IEEE Computer Society Conference on}, volume~1, pages
  I--423. IEEE, 2001.

\bibitem{tanawongsuwan2001}
Rawesak Tanawongsuwan and Aaron Bobick.
\newblock Gait recognition from time-normalized joint-angle trajectories in the
  walking plane.
\newblock In {\em Computer Vision and Pattern Recognition, 2001. CVPR 2001.
  Proceedings of the 2001 IEEE Computer Society Conference on}, volume~2, pages
  II--726. IEEE, 2001.

\bibitem{boulgouris2007}
Nikolaos~V Boulgouris and Zhiwei~X Chi.
\newblock Human gait recognition based on matching of body components.
\newblock {\em Pattern Recognition}, 40(6):1763--1770, 2007.

\bibitem{zeng2014}
Wei Zeng, Cong Wang, and Yuanqing Li.
\newblock Model-based human gait recognition via deterministic learning.
\newblock {\em Cognitive Computation}, 6(2):218--229, 2014.

\bibitem{lee2002}
Lily Lee and W~Eric~L Grimson.
\newblock Gait analysis for recognition and classification.
\newblock In {\em Automatic Face and Gesture Recognition, 2002. Proceedings.
  Fifth IEEE International Conference on}, pages 148--155. IEEE, 2002.

\bibitem{zhang2004}
Jiayong Zhang, Robert Collins, and Yanxi Liu.
\newblock Representation and matching of articulated shapes.
\newblock In {\em Computer Vision and Pattern Recognition, 2004. CVPR 2004.
  Proceedings of the 2004 IEEE Computer Society Conference on}, volume~2, pages
  II--342. IEEE, 2004.

\bibitem{zhang2007}
Rong Zhang, Christian Vogler, and Dimitris Metaxas.
\newblock Human gait recognition at sagittal plane.
\newblock {\em Image and vision computing}, 25(3):321--330, 2007.

\bibitem{lu2007}
Haiping Lu, Konstantinos~N Plataniotis, and Anastasios~N Venetsanopoulos.
\newblock A full-body layered deformable model for automatic model-based gait
  recognition.
\newblock {\em EURASIP Journal on Advances in Signal Processing},
  2008(1):1--13, 2007.

\bibitem{ariyanto2012}
Gunawan Ariyanto and Mark~S Nixon.
\newblock Marionette mass-spring model for 3d gait biometrics.
\newblock In {\em Biometrics (ICB), 2012 5th IAPR International Conference on},
  pages 354--359. IEEE, 2012.

\bibitem{yoo2008}
Jang-Hee Yoo, Doosung Hwang, Ki-Young Moon, and Mark~S Nixon.
\newblock Automated human recognition by gait using neural network.
\newblock In {\em Image Processing Theory, Tools and Applications, 2008. IPTA
  2008. First Workshops on}, pages 1--6. IEEE, 2008.

\bibitem{tafazzoli2010}
Faezeh Tafazzoli and Reza Safabakhsh.
\newblock Model-based human gait recognition using leg and arm movements.
\newblock {\em Engineering applications of artificial intelligence},
  23(8):1237--1246, 2010.

\bibitem{hayfron2003}
James~B Hayfron-Acquah, Mark~S Nixon, and John~N Carter.
\newblock Automatic gait recognition by symmetry analysis.
\newblock {\em Pattern Recognition Letters}, 24(13):2175--2183, 2003.

\bibitem{bashir2009}
Khalid Bashir, Tao Xiang, Shaogang Gong, and Q~Mary.
\newblock Gait representation using flow fields.
\newblock In {\em BMVC}, pages 1--11, 2009.

\bibitem{benabdelkader2001}
Chiraz BenAbdelkader, Ross Cutler, Harsh Nanda, and Larry Davis.
\newblock Eigengait: Motion-based recognition of people using image
  self-similarity.
\newblock In {\em Audio-and Video-Based Biometric Person Authentication}, pages
  284--294. Springer, 2001.

\bibitem{tao2007}
Dacheng Tao, Xuelong Li, Xindong Wu, and Stephen~J Maybank.
\newblock General tensor discriminant analysis and gabor features for gait
  recognition.
\newblock {\em Pattern Analysis and Machine Intelligence, IEEE Transactions
  on}, 29(10):1700--1715, 2007.

\bibitem{liu2004}
Zongyi Liu and Sudeep Sarkar.
\newblock Simplest representation yet for gait recognition: Averaged
  silhouette.
\newblock In {\em Pattern Recognition, 2004. ICPR 2004. Proceedings of the 17th
  International Conference on}, volume~4, pages 211--214. IEEE, 2004.

\bibitem{kale2002}
Amit Kale, AN~Rajagopalan, Naresh Cuntoor, and Volker Kruger.
\newblock Gait-based recognition of humans using continuous hmms.
\newblock In {\em Automatic Face and Gesture Recognition, 2002. Proceedings.
  Fifth IEEE International Conference on}, pages 336--341. IEEE, 2002.

\bibitem{collins2002}
Robert~T Collins, Ralph Gross, and Jianbo Shi.
\newblock Silhouette-based human identification from body shape and gait.
\newblock In {\em Automatic Face and Gesture Recognition, 2002. Proceedings.
  Fifth IEEE International Conference on}, pages 366--371. IEEE, 2002.

\bibitem{wang2003b}
Liang Wang, Tieniu Tan, Weiming Hu, and Huazhong Ning.
\newblock Automatic gait recognition based on statistical shape analysis.
\newblock {\em IEEE transactions on image processing}, 12(9):1120--1131, 2003.

\bibitem{kent1992new}
John~T Kent.
\newblock New directions in shape analysis.
\newblock {\em The art of statistical science}, pages 115--127, 1992.

\bibitem{lee2007}
Seungkyu Lee, Yanxi Liu, and Robert Collins.
\newblock Shape variation-based frieze pattern for robust gait recognition.
\newblock In {\em Computer Vision and Pattern Recognition, 2007. CVPR'07. IEEE
  Conference on}, pages 1--8. IEEE, 2007.

\bibitem{rida2018comprehensive}
Imad Rida, Noor Al-Maadeed, Somaya Al-Maadeed, and Sambit Bakshi.
\newblock A comprehensive overview of feature representation for biometric
  recognition.
\newblock {\em Multimedia Tools and Applications}, pages 1--24, 2018.

\bibitem{wang2003}
Liang Wang, Tieniu Tan, Huazhong Ning, and Weiming Hu.
\newblock Silhouette analysis-based gait recognition for human identification.
\newblock {\em IEEE transactions on pattern analysis and machine intelligence},
  25(12):1505--1518, 2003.

\bibitem{benabdelkader2004}
Chiraz BenAbdelkader, Ross~G Cutler, and Larry~S Davis.
\newblock Gait recognition using image self-similarity.
\newblock {\em EURASIP Journal on Advances in Signal Processing},
  2004(4):1--14, 2004.

\bibitem{kobayashi2004}
Takumi Kobayashi and Nobuyuki Otsu.
\newblock Action and simultaneous multiple-person identification using cubic
  higher-order local auto-correlation.
\newblock In {\em Pattern Recognition, 2004. ICPR 2004. Proceedings of the 17th
  International Conference on}, volume~4, pages 741--744. IEEE, 2004.

\bibitem{otsu1988}
Nobuyuki Otsu and Takio Kurita.
\newblock A new scheme for practical flexible and intelligent vision systems.
\newblock In {\em MVA}, pages 431--435, 1988.

\bibitem{lu20077}
Jiwen Lu and Erhu Zhang.
\newblock Gait recognition for human identification based on ica and fuzzy svm
  through multiple views fusion.
\newblock {\em Pattern Recognition Letters}, 28(16):2401--2411, 2007.

\bibitem{belhumeur1997eigenfaces}
Peter~N. Belhumeur, Jo{\~a}o~P Hespanha, and David~J. Kriegman.
\newblock Eigenfaces vs. fisherfaces: Recognition using class specific linear
  projection.
\newblock {\em IEEE Transactions on pattern analysis and machine intelligence},
  19(7):711--720, 1997.

\bibitem{hofmann2012}
Martin Hofmann and Gerhard Rigoll.
\newblock Improved gait recognition using gradient histogram energy image.
\newblock In {\em 2012 19th IEEE International Conference on Image Processing},
  pages 1389--1392. IEEE, 2012.

\bibitem{martin2014}
Ra{\'u}l Mart{\'\i}n-F{\'e}lez and Tao Xiang.
\newblock Uncooperative gait recognition by learning to rank.
\newblock {\em Pattern Recognition}, 47(12):3793--3806, 2014.

\bibitem{xing2016}
Xianglei Xing, Kejun Wang, Tao Yan, and Zhuowen Lv.
\newblock Complete canonical correlation analysis with application to
  multi-view gait recognition.
\newblock {\em Pattern Recognition}, 50:107--117, 2016.

\bibitem{xu2006}
Dong Xu, Shuicheng Yan, Dacheng Tao, Lei Zhang, Xuelong Li, and Hong-Jiang
  Zhang.
\newblock Human gait recognition with matrix representation.
\newblock {\em IEEE Transactions on Circuits and Systems for Video Technology},
  16(7):896--903, 2006.

\bibitem{xu2004}
Dong Xu, Shuicheng Yan, Lei Zhang, Zhengkai Liu, and HongJiang Zhang.
\newblock Coupled subspaces analysis.
\newblock {\em Techn. Rep. No. MSR-TR-2004-106}, 2004.

\bibitem{yan2005}
Shuicheng Yan, Dong Xu, Benyu Zhang, and Hong-Jiang Zhang.
\newblock Graph embedding: A general framework for dimensionality reduction.
\newblock In {\em 2005 IEEE Computer Society Conference on Computer Vision and
  Pattern Recognition (CVPR'05)}, volume~2, pages 830--837. IEEE, 2005.

\bibitem{xu2007}
Dong Xu, Shuicheng Yan, Dacheng Tao, Stephen Lin, and Hong-Jiang Zhang.
\newblock Marginal fisher analysis and its variants for human gait recognition
  and content-based image retrieval.
\newblock {\em Image Processing, IEEE Transactions on}, 16(11):2811--2821,
  2007.

\bibitem{chen2010}
Changyou Chen, Junping Zhang, and Rudolf Fleischer.
\newblock Distance approximating dimension reduction of riemannian manifolds.
\newblock {\em IEEE Transactions on Systems, Man, and Cybernetics, Part B
  (Cybernetics)}, 40(1):208--217, 2010.

\bibitem{guan2015}
Yu~Guan, Chang-Tsun Li, and Fabio Roli.
\newblock On reducing the effect of covariate factors in gait recognition: a
  classifier ensemble method.
\newblock {\em IEEE transactions on pattern analysis and machine intelligence},
  37(7):1521--1528, 2015.

\bibitem{rida2014improved}
Imad Rida, Somaya Almaadeed, and Ahmed Bouridane.
\newblock Improved gait recognition based on gait energy images.
\newblock In {\em 2014 26th International Conference on Microelectronics
  (ICM)}, pages 40--43. IEEE, 2014.

\bibitem{bashir2008}
Khalid Bashir, Tao Xiang, and Shaogang Gong.
\newblock Feature selection on gait energy image for human identification.
\newblock In {\em Acoustics, Speech and Signal Processing, 2008. ICASSP 2008.
  IEEE International Conference on}, pages 985--988. IEEE, 2008.

\bibitem{rida2015}
Imad Rida, Ahmed Bouridane, Gian~Luca Marcialis, and Pierluigi Tuveri.
\newblock Improved human gait recognition.
\newblock In {\em Image Analysis and Processing?ICIAP 2015}, pages 119--129.
  Springer, 2015.

\bibitem{rida2016}
Imad Rida, Somaya Almaadeed, and Ahmed Bouridane.
\newblock Gait recognition based on modified phase-only correlation.
\newblock {\em Signal, Image and Video Processing}, 10(3):463--470, 2016.

\bibitem{jeevan2013}
Mahadevu Jeevan, Nikhil Jain, Madasu Hanmandlu, and Girija Chetty.
\newblock Gait recognition based on gait pal and pal entropy image.
\newblock In {\em Image Processing (ICIP), 2013 20th IEEE International
  Conference on}, pages 4195--4199. IEEE, 2013.

\bibitem{pal1991entropy}
Nikhil~R Pal and Sankar~K Pal.
\newblock Entropy: A new definition and its applications.
\newblock {\em IEEE transactions on systems, man, and cybernetics},
  21(5):1260--1270, 1991.

\bibitem{kusakunniran2014a}
Worapan Kusakunniran.
\newblock Recognizing gaits on spatio-temporal feature domain.
\newblock {\em Information Forensics and Security, IEEE Transactions on},
  9(9):1416--1423, 2014.

\bibitem{kusakunniran2014b}
Worapan Kusakunniran.
\newblock Attribute-based learning for gait recognition using spatio-temporal
  interest points.
\newblock {\em Image and Vision Computing}, 32(12):1117--1126, 2014.

\bibitem{hu2013}
Maodi Hu, Yunhong Wang, Zhaoxiang Zhang, De~Zhang, and James~J Little.
\newblock Incremental learning for video-based gait recognition with lbp flow.
\newblock {\em IEEE transactions on cybernetics}, 43(1):77--89, 2013.

\bibitem{rokanujjaman2015}
Md~Rokanujjaman, Md~Shariful Islam, Md~Altab Hossain, Md~Rezaul Islam, Yashushi
  Makihara, and Yasushi Yagi.
\newblock Effective part-based gait identification using frequency-domain gait
  entropy features.
\newblock {\em Multimedia Tools and Applications}, 74(9):3099--3120, 2015.

\bibitem{choudhury2015}
Sruti~Das Choudhury and Tardi Tjahjadi.
\newblock Robust view-invariant multiscale gait recognition.
\newblock {\em Pattern Recognition}, 48(3):798--811, 2015.

\bibitem{rida2016robust}
Imad Rida, Larbi Boubchir, Noor Al-Maadeed, Somaya Al-Maadeed, and Ahmed
  Bouridane.
\newblock Robust model-free gait recognition by statistical dependency feature
  selection and globality-locality preserving projections.
\newblock In {\em 2016 39th International Conference on Telecommunications and
  Signal Processing (TSP)}, pages 652--655. IEEE, 2016.

\bibitem{rida2017improved}
Imad Rida, Noor Al~Maadeed, Gian~Luca Marcialis, Ahmed Bouridane, Romain
  Herault, and Gilles Gasso.
\newblock Improved model-free gait recognition based on human body part.
\newblock In {\em Biometric Security and Privacy}, pages 141--161. Springer,
  2017.

\bibitem{rida2016human}
Imad Rida, Xudong Jiang, and Gian~Luca Marcialis.
\newblock Human body part selection by group lasso of motion for model-free
  gait recognition.
\newblock {\em IEEE Signal Processing Letters}, 23(1):154--158, 2016.

\bibitem{hossain2010}
Md~Altab Hossain, Yasushi Makihara, Junqiu Wang, and Yasushi Yagi.
\newblock Clothing-invariant gait identification using part-based clothing
  categorization and adaptive weight control.
\newblock {\em Pattern Recognition}, 43(6):2281--2291, 2010.

\bibitem{foster2003}
Jeff~P Foster, Mark~S Nixon, and Adam Pr{\"u}gel-Bennett.
\newblock Automatic gait recognition using area-based metrics.
\newblock {\em Pattern Recognition Letters}, 24(14):2489--2497, 2003.

\bibitem{bleakley2011}
Kevin Bleakley and Jean-Philippe Vert.
\newblock The group fused lasso for multiple change-point detection.
\newblock {\em arXiv preprint arXiv:1106.4199}, 2011.

\bibitem{yuan2006}
Ming Yuan and Yi~Lin.
\newblock Model selection and estimation in regression with grouped variables.
\newblock {\em Journal of the Royal Statistical Society: Series B (Statistical
  Methodology)}, 68(1):49--67, 2006.

\bibitem{rida:tel-01515364}
Imad Rida.
\newblock {\em {Temporal signals classification}}.
\newblock Theses, {Normandie Universit{\'e}}, February 2017.

\bibitem{huang1999}
Ping~S Huang, Chris~J Harris, and Mark~S Nixon.
\newblock Recognising humans by gait via parametric canonical space.
\newblock {\em Artificial Intelligence in Engineering}, 13(4):359--366, 1999.

\bibitem{jiang2009}
Xudong Jiang.
\newblock Asymmetric principal component and discriminant analyses for pattern
  classification.
\newblock {\em IEEE Transactions on Pattern Analysis and Machine Intelligence},
  31(5):931--937, 2009.

\bibitem{jiang20111}
Xudong Jiang.
\newblock Linear subspace learning-based dimensionality reduction.
\newblock {\em IEEE Signal Processing Magazine}, 28(2):16--26, 2011.

\bibitem{bellet2013survey}
Aur{\'e}lien Bellet, Amaury Habrard, and Marc Sebban.
\newblock A survey on metric learning for feature vectors and structured data.
\newblock {\em arXiv preprint arXiv:1306.6709}, 2013.

\bibitem{rida2014supervised}
Imad Rida, Romain Herault, and Gilles Gasso.
\newblock Supervised music chord recognition.
\newblock In {\em Machine Learning and Applications (ICMLA), 2014 13th
  International Conference on}, pages 336--341. IEEE, 2014.

\bibitem{gopalan2011domain}
Raghuraman Gopalan, Ruonan Li, and Rama Chellappa.
\newblock Domain adaptation for object recognition: An unsupervised approach.
\newblock In {\em 2011 international conference on computer vision}, pages
  999--1006. IEEE, 2011.

\bibitem{kulis2011you}
Brian Kulis, Kate Saenko, and Trevor Darrell.
\newblock What you saw is not what you get: Domain adaptation using asymmetric
  kernel transforms.
\newblock In {\em Computer Vision and Pattern Recognition (CVPR), 2011 IEEE
  Conference on}, pages 1785--1792. IEEE, 2011.

\bibitem{sun2015return}
Baochen Sun, Jiashi Feng, and Kate Saenko.
\newblock Return of frustratingly easy domain adaptation.
\newblock {\em arXiv preprint arXiv:1511.05547}, 2015.

\bibitem{rida2018efficient}
Imad Rida, Romain H{\'e}rault, and Gilles Gasso.
\newblock An efficient supervised dictionary learning method for audio signal
  recognition.
\newblock {\em arXiv preprint arXiv:1812.04748}, 2018.

\bibitem{courty2016optimal}
N.~Courty, R.~Flamary, D.~Tuia, and A.~Rakotomamonjy.
\newblock Optimal transport for domain adaptation.
\newblock {\em IEEE Transactions on Pattern Analysis and Machine Intelligence},
  PP(99):1--1, 2016.

\end{thebibliography}

\end{document}